\documentclass{article}

 \usepackage[preprint]{neurips_2026}

\usepackage[utf8]{inputenc} 
\usepackage[T1]{fontenc}    
\usepackage{hyperref}       
\usepackage{url}            
\usepackage{booktabs}       
\usepackage{amsfonts}       
\usepackage{nicefrac}       
\usepackage{microtype}      
\usepackage{xcolor}         
\usepackage{graphicx}
\usepackage{amsmath}
\usepackage{subfig}
\usepackage{placeins}
\title{A dimensional R2 regression metric}

\author{%
  Jaesung Yoo\\
  Neuroscience Center\\
  University of North Carolina at Chapel Hill\\
  Chapel Hill, NC, 27599\\
  \texttt{jsyoo61@unc.edu}\\
  \AND
  Stefan Lemke\\
  Neuroscience Center\\
  University of North Carolina at Chapel Hill\\
  Chapel Hill, NC, 27599\\
  \texttt{slemke@email.unc.edu}\\
  \And
  Jian Zhong Guo\\
  Neuroscience Center\\
  University of North Carolina at Chapel Hill\\
  Chapel Hill, NC, 27599\\
  \texttt{jayzhong@email.unc.edu}\\
  \And
  Kanaka Rajan\\
  Kempner Institute for the Study of Natural and Artificial Intelligence\\
  Harvard University\\
  Boston, MA, 02134\\
  \texttt{kanaka\_rajan@hms.harvard.edu} \\
  \And
  Adam Hantman\\
  Neuroscience Center\\
  University of North Carolina at Chapel Hill\\
  Chapel Hill, NC, 27599\\
  \texttt{adam\_hantman@med.unc.edu} \\
}

\begin{document}

\maketitle

\begin{abstract}
R2 score is the standard metric for evaluating regression tasks, offering a normalized magnitude-agnostic measure of accuracy that captures variance. However, R2 has three key limitations: it is limited to at most two dimensional inputs, it reduces the score to a single scalar that hides rich patterns of prediction accuracy, and it is sensitive to low-variance noise channels which can yield large, uninterpretable negative values. We introduce the Dimensional R2 score (Dim-R2), a simple extension of R2 that accepts data of arbitrary dimensionality, provides a multidimensional view of accuracy, and reduces sensitivity to noise. We demonstrate its advantages on both synthetic sinusoidal data and three multidimensional regression datasets. Dim-R2 offers an interpretable and flexible metric that highlights patterns in regression accuracy, guiding regression modeling.
\end{abstract}

\section{Introduction}
Evaluation metrics are key guides in modeling. They quantify how well the model's predictions match the target data and guide decisions such as model tuning and data cleaning \citep{doi:10.1126/science.aaa8415}. For regression tasks, the R2 score is considered the gold standard compared to metrics like mean absolute error (MAE) or mean squared error (MSE) \citep{chicco2021coefficient}. MAE shows a simple magnitude based error but cannot differentiate between biased predictions and lazy mean predictions. MSE penalizes variance and resolves the lazy mean prediction issue, but its value ranges per data domain making it hard to interpret. In contrast, R2 is a normalized, data domain-independent metric that reflects variance explained by the prediction, normalized by variance of the data (Eq. \ref{eq:r2_score}). Due to its ability to capture variance and its normalized score, R2 is used as the standard metric in regression evaluation \citep{chicco2021coefficient, sykes1993introduction, ash1999r2, sauerbrei2020cortical, scikit-learn}.

However, R2 has three limitations. First, R2 is defined for 1D data which is averaged across channels for 2D data, and it cannot be directly applied to higher-dimensional regression data \citep{heckel2024deep, ahmed2025comprehensive,10.1093/jrsssb/qkad073}. Second, the R2 reduces model performance to a single scalar or a 1D score, offering no insight into how accuracy varies across data dimensions (Fig. \ref{fig:Dim-R2 dimensional view}). A multidimensional view of regression accuracy could reveal structure that could help modelers target specific features of their data and model for improvement. Third, R2 is highly sensitive to low-variance noise channels in multi-channel (2D) regression tasks. It can yield large negative values when the true data has little variation, as is the case for noisy channels. When these R2 scores are averaged across channels for multi-channel data, the mean R2 can be largely negative which obscures the presence of high accuracy channels (Fig. \ref{fig:Dim-R2 resilience}).

To address these limitations, we introduce the Dimensional R2 score (Dim-R2), a simple extension of R2. Dim-R2 flattens selected dimensions into independent observations and computes R2 along the remaining dimensions, with explicit control over which dimensional variability to normalize against. This allows regression data of arbitrary dimensionality, overcoming the 2D limitation of R2. As Dim-R2 can flatten and keep any dimensions, it also enables a multidimensional view of prediction accuracy (Fig. \ref{fig:Dim-R2 dimensional view}). For example, in data shaped (Trials, Time, Channels) \citep{perich2020inferring,YOO2022107079}, Dim-R2 can highlight noisy trials, temporally localized features, or task-relevant channels by revealing predictability across each dimension. Finally, flattening dimensions into observations allows high-variance (informative) channels to outweigh low-variance (noisy) ones, yielding a more robust score than mean per-channel R2 scores. This highlights the presence of high accuracy channels (Fig. \ref{fig:Dim-R2 resilience}), allowing users to compare a noise-resilient score across experiments.

Here, we introduce Dim-R2, then present the dimensional view of regression accuracy using Dim-R2. Next, we evaluate the resilience of Dim-R2 by comparing it to Mean-R2. The dimensional-view is demonstrated using toy sinusoidal data and several multidimensional regression tasks where the data contains spatial features. The resilience feature is demonstrated with another toy sinusoidal dataset and a hyperparameter sweep on data-constrained recurrent neural network (DC-RNN) modeling \citep{perich2020inferring, PERICH2020146} trained to simulate mouse neural activity during a reach-to-grab task.

\section{Related works}
Regression tasks \citep{sykes1993introduction,krzywinski2015multiple, altman2015points} aim to estimate a continuous target variable $y$ from input data $x$, with predictions given by $\hat{y} = f(x)$ where $f(x)$ is any model. A regression metric should satisfy four criteria: (1) capture the variance structure of the target, (2) penalize bias between $y$ and $\hat{y}$,(3) be normalized to allow comparison across data scales, and (4) provide interpretable reference points (e.g., perfect prediction, mean prediction).

The R2 score satisfies all four criteria and remains the standard regression metric \citep{agriculture12101515,LIU2024100711,kogan2020assessing}. In contrast, many commonly used metrics \citep{hastie2015statistical} only satisfy a subset of the regression metric criteria (Table \ref{tab:regression-metric-comparison}). R2 measures the squared error between $y, \hat{y} \in \mathbb{R}^N$, the ground-truth target and predicted values, normalized by the variance of $y$ (Eq. \ref{eq:r2_score}). It is defined for a single channel data (1D) with $N$ observations. The value of R2 ranges from ($-\infty$, 1], where 1 indicates perfect prediction, 0 corresponds to predicting $\bar{y}$, the mean of $y$, and negative values indicate worse performance than predicting $\bar{y}$.

\begin{equation}
    R2=1-\frac{RSS}{TSS}=1-\frac{\sum^{N}_{i}{(y_{i}-\hat{y}_{i})^2}}{\sum^{N}_{i}{(y_{i}-\bar{y})^2}}, \qquad
    \label{eq:r2_score}
\end{equation}

where $i\in[1,N]$ is the observation index, and RSS, TSS refer to residual sum of squares and total sum of squares, respectively. When $y$ has low variance, TSS becomes small, amplifying modest prediction errors into large negative R2 values.

\begin{table}[t]
\centering
\caption{Comparison of common regression metrics (equations in Appendix \ref{appendix:common-regression-metrics}). Abbreviations:
R2 = R2 score;
MSE = mean squared error;
MAE = mean absolute error;
D2-AE = D2 absolute error;
EV = explained variance;
Corr = Pearson correlation. \textsuperscript{*}Not regression metrics but included for reference.
}
\label{tab:regression-metric-comparison}

\begin{tabular}{lcccc}
\toprule
\textbf{Metric} 
& \textbf{Captures Variance} 
& \textbf{Captures Bias} 
& \textbf{Normalized} 
& \textbf{Reference Baselines} \\
\midrule
R2                         & \checkmark & \checkmark & \checkmark & \checkmark \\
MSE        & \checkmark & \checkmark & $\times$ & $\times$ \\
MAE       & $\times$ & \checkmark & $\times$ & $\times$ \\
D2 AE               & $\times$ & \checkmark & \checkmark & \checkmark \\
EV\textsuperscript{*}        & \checkmark & $\times$ & \checkmark & \checkmark \\
Corr ($\rho$)\textsuperscript{*} & \checkmark & $\times$ & \checkmark & \checkmark \\
\bottomrule
\end{tabular}
\end{table}

\section{Materials and Methods}
\label{sec:materials_and_methods}

\subsection{Regression Metrics: Conventional and Dimensional R2}


For multi-channel 2D data $y, \hat{y} \in \mathbb{R}^{N \times C}$, R2 is measured per channel and averaged to yield the mean R2 (Eq. \ref{eq:mean-r2}). As a result, conventional R2 supports at most 2D input.

\begin{equation}
    Mean\ R2 = \frac{1}{C}\sum_{c}^{C}[1-\frac{\sum^{N}_{i}{(y_{i,c}-\hat{y}_{i,c})^2}}{\sum^{N}_{i}{(y_{i,c}-\overline{y}_{c})^2}}]
    \label{eq:mean-r2}
\end{equation}

where $C$, $c$ refers to the number of channels and channel index, respectively.

\begin{figure} [t]
    \centering
    \includegraphics[width=\textwidth]{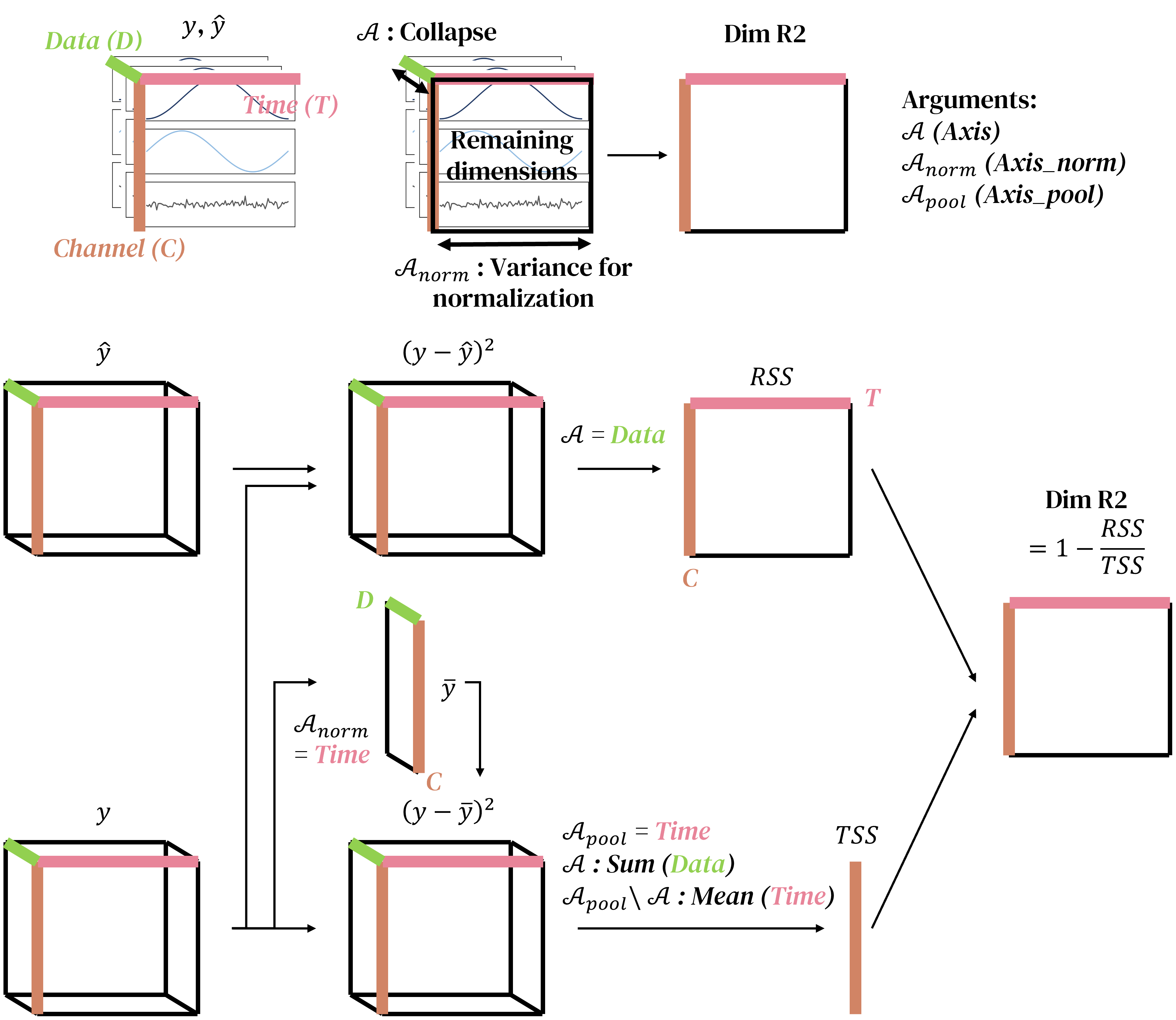}
    \caption{Schematic of Dim-R2 on 3D $y$ and $\hat{y}$ with $\mathcal{A}=Data$, $\mathcal{A}_{norm}=\mathcal{A}_{pool}=Time$.}
    \label{fig:Dim-R2 schematic}
\end{figure}

Dim-R2 follows the concept of Eq. \ref{eq:r2_score} but computes RSS and TSS in multidimensional form (Fig. \ref{fig:Dim-R2 schematic}, Eq. \ref{eq:RSS}-\ref{eq:dim-r2}). It takes one argument (\texttt{Axis}) and two optional arguments (\texttt{Axis\_norm}, \texttt{Axis\_pool}), noted as $\mathcal{A},\mathcal{A}_{norm},\mathcal{A}_{pool}$, respectively.  The $\mathcal{A}$ defines the dimensions to collapse which defines the output shape, $\mathcal{A}_{norm}$ defines the dimensions to measure variance for normalization, and $\mathcal{A}_{pool}$ averages the total sum of squares across additional dimensions for a more stable normalization. Let the full set of axes be $\mathcal{D}$.

\begin{equation}
    RSS = \sum_{k \in \mathcal{A}} (y_{k} - \hat{y}_{k} )^{2},
    \qquad
    RSS \in \mathbb{R}^{\mathcal{D} \setminus \mathcal{A} }
    \label{eq:RSS}
\end{equation}
\begin{equation}
    \bar{y} = \frac{1}{|\mathcal{A}_{norm}|}
    \sum_{k \in \mathcal{A}_{norm}} y_{k}, \qquad \bar{y} \in \mathbb{R}^{\mathcal{D} \setminus \mathcal{A}_{norm}}
    \label{eq:y_mean}
\end{equation}
\begin{equation}
    TSS = \frac{1}{|\mathcal{A}_{pool} \setminus \mathcal{A}|}
    \sum_{j \in \mathcal{A}_{pool} \setminus \mathcal{A}}
    \sum_{k \in \mathcal{A}}
    (y_{k,j} - \bar{y})^{2}, \qquad TSS \in \mathbb{R}^{\mathcal{D} \setminus (\mathcal{A}_{pool} \cup \mathcal{A})}
    \label{eq:TSS}
\end{equation}
\begin{equation}
    Dim\text{-}R2 = 1-\frac{RSS}{TSS}, \qquad Dim\text{-}R2 \in \mathbb{R}^{\mathcal{D}\setminus \mathcal{A} }
    \label{eq:dim-r2}
\end{equation}

where $k$ indexes the data in the specified dimensions, $|\mathcal{A}_{norm}|$ and $|\mathcal{A}_{pool} \setminus \mathcal{A}|$ denote the number of observations along the specified axes, used to compute the mean. The constraint $\mathcal{A}_{norm} \subseteq \mathcal{A}_{pool}$ must hold for proper variance normalization, and by default $\mathcal{A}_{pool}$ is set to $\mathcal{A}_{norm}$. By default, $\mathcal{A}_{norm}$ is set to $\mathcal{A}$, measuring variability across the collapsed dimensions.

When computing RSS and TSS, dimensions in $\mathcal{A}$ and $\mathcal{A}_{pool}$ are treated as independent observations. Dim-R2 is then computed by broadcasting RSS and TSS shapes, allowing Dim-R2 to accept data of arbitrary dimensionality. By selecting which dimensions to collapse ($\mathcal{A}$) and retain ($\mathcal{D} \setminus \mathcal{A}$), Dim-R2 provides a multidimensional view of accuracy across the retained dimensions. When $\mathcal{A}=\mathcal{D}$, Dim-R2 yields a single score where high-variance channels contribute more variance to a larger TSS, reducing the influence of low-variance noise channels compared to mean R2. Dim-R2 is a generalization of conventional R2 and reduces to the variance-weighted mean R2 for 2D data when $\mathcal{A}=\mathcal{D}$ and $\mathcal{A}_{norm}$ is set to the observation dimension (Appendix \ref{appendix:properties-of-dim-r2}). An illustrative example of Dim-R2 argument selection is provided in Fig. \ref{fig:Dim-R2 example data} and Table \ref{tab:dim-r2-example}. The implementation is in Appendix \ref{file:dim-R2}, written in Python using NumPy \citep{harris2020array} and following scikit-learn syntax \citep{scikit-learn}.


\subsection{Datasets}
\subsubsection{Synthetic Sinusoidal Dataset}
\label{datasets:synthetic}

\begin{figure}[t]
\centering
\includegraphics[width=0.8\textwidth]{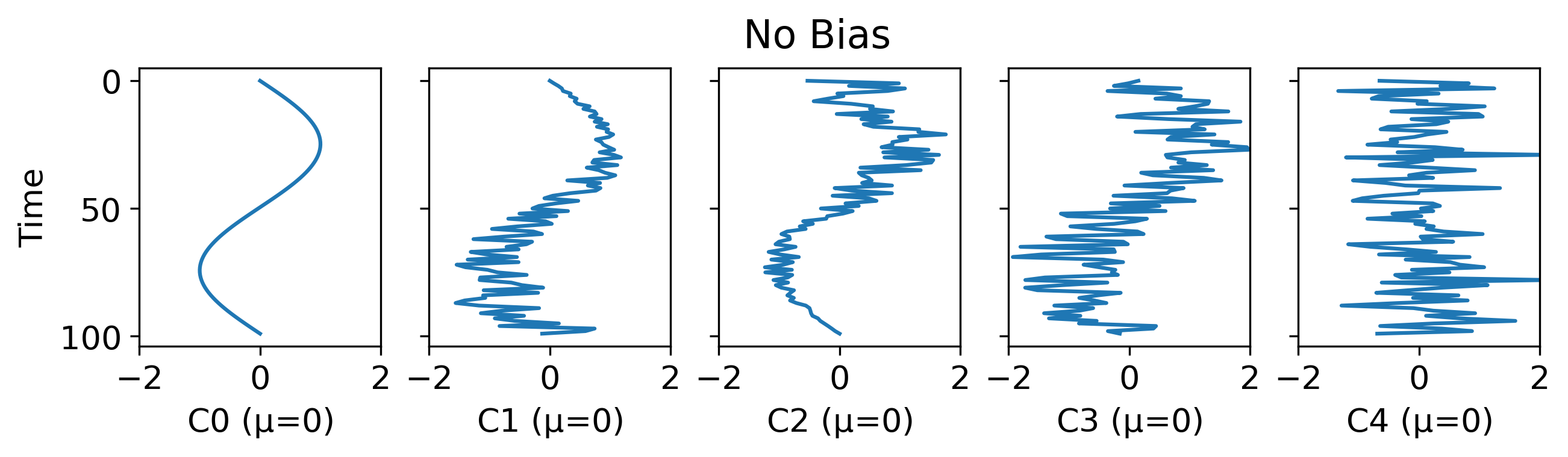}\\

\caption{Synthetic sinusoidal dataset $y$ and $\hat{y}$ with time-varying noise per channel. No Bias and Varying Channel Bias conditions differ in added bias, introducing variability across channel and data dimensions (Fig. \ref{fig:sample-time-varying-full}).}
\label{fig:sample-time-varying}
\vspace{-10pt}
\end{figure}

To illustrate the dimensional view of regression accuracy (Section \ref{results:dimensional-view}), we generated waveforms of shape (Data, Time, Channels)=(1000, 100, 5) (Fig. \ref{fig:sample-time-varying}). Channel C0 to C3 share the same sine wave while C4 is pure standard Gaussian noise ($\mu=0$,$\sigma=1$) adjusted to have the same variance as C0. Uniform random noise in $[-1,1]$ was added to $y$ and $\hat{y}$ with channel-specific temporal patterns: no noise (C0), linearly increasing scale from 0 to 1 over time (C1), linearly decreasing from 1 to 0 over time (C2), and constant noise (C3). To illustrate how Dim-R2 arguments affect the dimensional view, two conditions were used: no added bias with time-averaged values of 0 (No Bias), and a time bias varying from 0 to 4 across channels, randomly assigned across samples (Varying Channel Bias) (Fig. \ref{fig:sample-time-varying}, \ref{fig:sample-time-varying-full}). The No Bias condition has no variability across the data dimension and minimal variability across the channel dimension, while Varying Channel Bias condition has variability across both data and channel dimensions. To demonstrate Dim-R2's noise channel resilience compared to mean R2 (Section \ref{results:resilience}), multichannel sinusoidal data of shape (100 timesteps, 100 channels) was generated for $y$ and $\hat{y}$ across 100 random repetitions. A fixed ratio of channels was replaced by Gaussian noise of variance in \{0.01, 0.1, 1.0\} (Fig. \ref{fig:r2-sample}). 

\subsubsection{Application Datasets}
\label{datasets:application}

We applied Dim-R2 to multi-dimensional regression tasks ($y, \hat{y} \in \mathbb{R}^n$, $n\ge3$) to illustrate the dimensional view beyond synthetic data. These include neural activity prediction from DC-RNNs trained on mouse motor cortex Neuropixels recordings, and image reconstruction using Variational Autoencoder (VAE) \citep{NIPS2016_eb86d510, doersch2016tutorial, PinheiroCinelli2021} on MNIST (CC BY-SA 3.0) and CelebA (Non-commercial research only) datasets \citep{lecun2010mnist, liu2015faceattributes}. Full details are in Appendix \ref{appendix:dc-rnn} and \ref{appendix:vae}. To demonstrate how Dim-R2's noise resilience guides hyperparameter selection in noisy datasets, we conducted a hyperparameter sweep on the DC-RNN dataset (Appendix \ref{appendix:dc-rnn}).

\section{Results}
\subsection{Dim-R2 provides a dimensional view of regression accuracy}
\label{results:dimensional-view}

\begin{figure}[t]
\centering
\includegraphics[width=\textwidth]{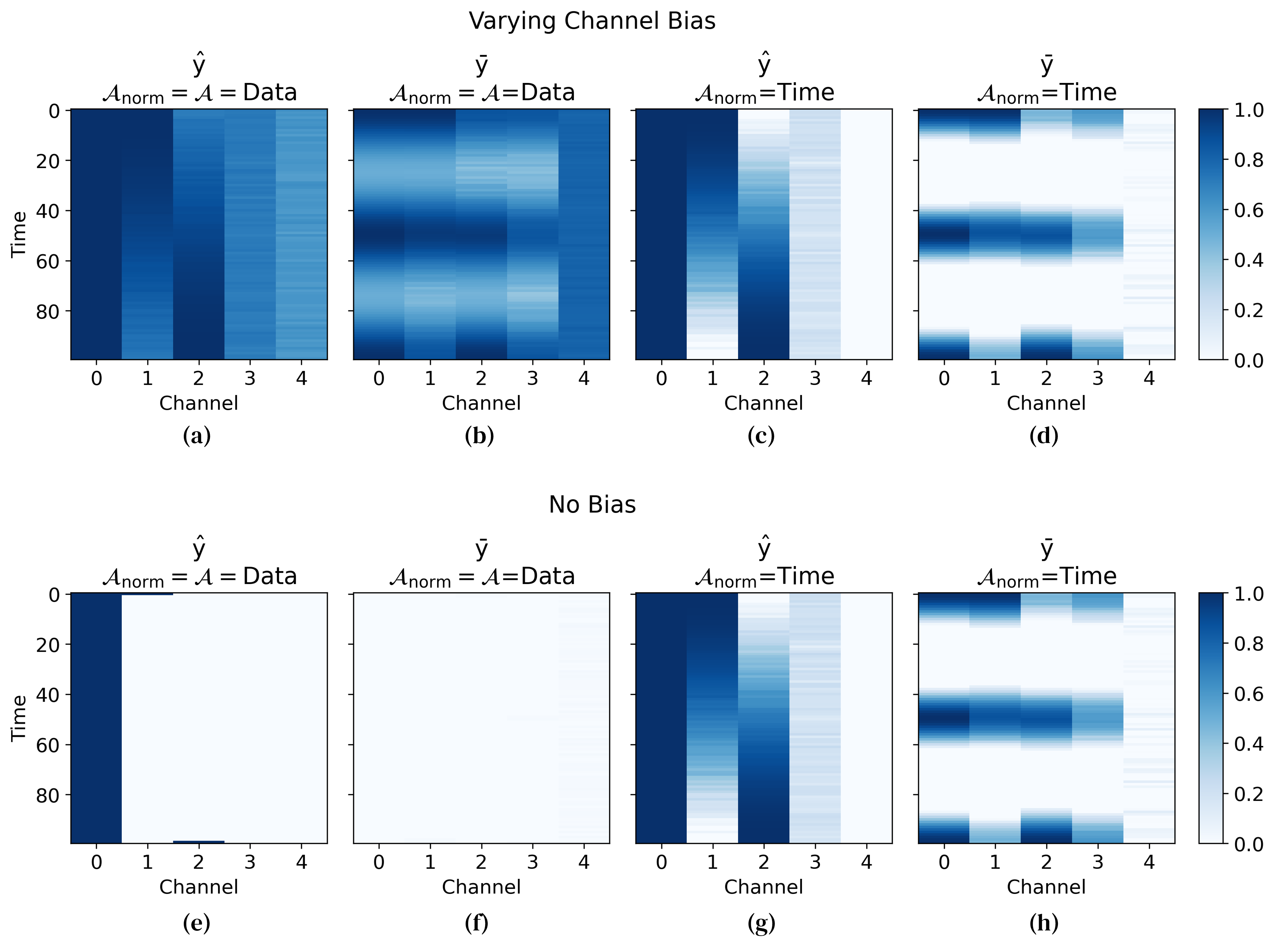}
\caption{Dimensional view of Dim-R2 on synthetic sinusoidal data with $\mathcal{A}$=Data, revealing scores across Time and Channel dimensions. Each heatmap shows Dim-R2 measured with $\hat{y}$ or $\bar{y}$ against $y$ under different $\mathcal{A}_{norm}$ settings. Top: Varying Channel Bias. Bottom: No Bias.}
\label{fig:dimensional-synthetic}
\vspace{-10pt}
\end{figure}

To demonstrate the dimensional view of regression accuracy, we applied Dim-R2 to the synthetic sinusoidal dataset with time-varying noise (Section \ref{datasets:synthetic}, Fig. \ref{fig:sample-time-varying},\ref{fig:dimensional-synthetic}). Mean R2 is not applicable as it is undefined for data beyond two dimensions. Capturing variability along each dimension is a meaningful regression goal. For Varying Channel Bias, $\mathcal{A}_{norm}$=Data measures accuracy relative to data variability, yielding high scores across all channels including noise channel C4 (Fig. \ref{fig:dimensional-synthetic}a). The time-averaged $\bar{y}$ also captures data variability when $\mathcal{A}_{norm}$=Data, yielding high scores rather than zero (Fig. \ref{fig:dimensional-synthetic}b). R2=0 for $\bar{y}$ requires $\mathcal{A}=\mathcal{A}_{norm}=\bar{y}$'s averaging dimension to all agree. Setting $\mathcal{A}_{norm}$=Time instead measures time-varying accuracy independent of data variability (Fig. \ref{fig:dimensional-synthetic}c). When the three conditions do not agree, $\bar{y}$ reveals how error varies across time rather than yielding R2=0 (Fig. \ref{fig:dimensional-synthetic}d). For No Bias, negligible data variability causes $\mathcal{A}_{norm}$=Data to amplify small errors, yielding largely negative values for both $\hat{y}$ and $\bar{y}$ (Fig. \ref{fig:dimensional-synthetic}e, f). Since time variability is identical between Varying Channel Bias and No Bias conditions, $\mathcal{A}_{norm}$=Time yields the same results (Fig. \ref{fig:dimensional-synthetic}g,h).


\begin{figure}[t]
\centering
\includegraphics[width=\textwidth]{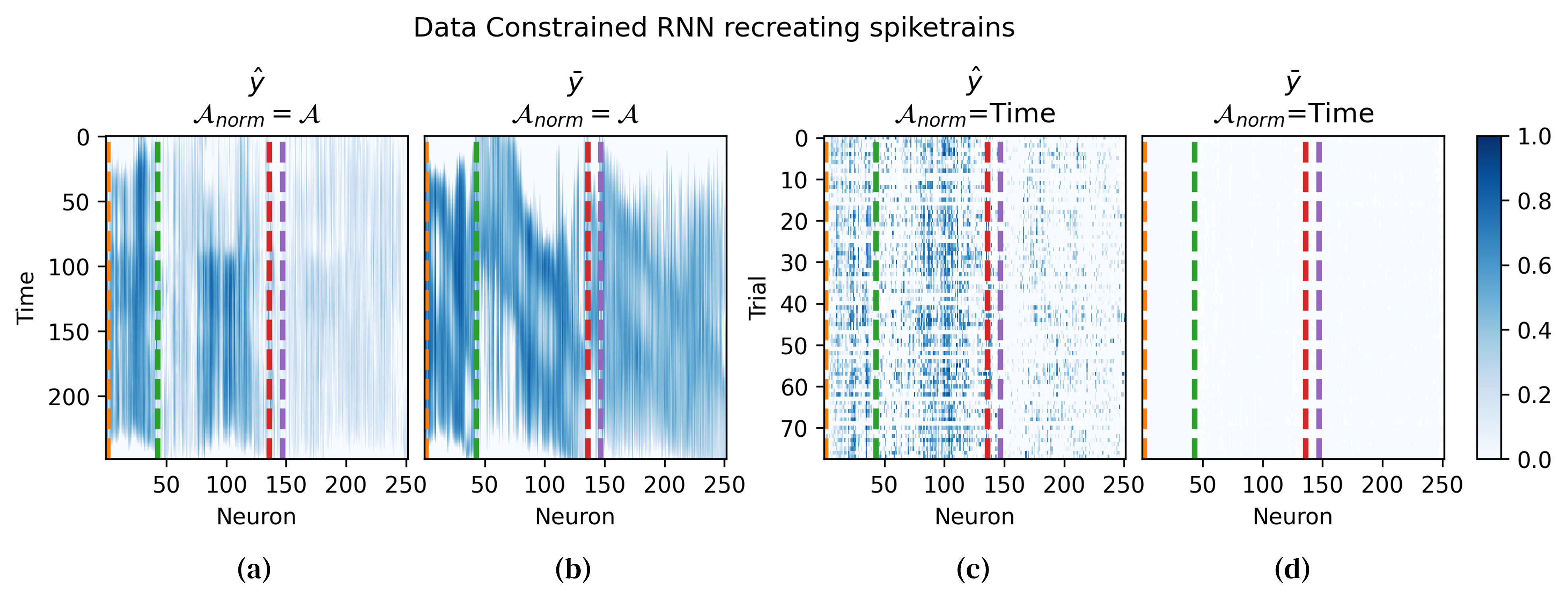}
\caption{Dimensional view of Dim-R2 on DC-RNN neural data. Each heatmap shows Dim-R2 measured with $\hat{y}$ or $\bar{y}$ against $y$ under different $\mathcal{A}$ and $\mathcal{A}_{norm}$ settings. Dashed lines separate brain regions (left to right): DCN (orange), M1 (green), Striatum (red), Thalamus (purple). (a) \& (b) $\mathcal{A}$=(Random seed (RS), Cross validation fold (CV), Trial). (c) \& (d)  $\mathcal{A}$=(RS, CV, Time). Full $\mathcal{A}$ and $\mathcal{A}_{norm}$ combinations are shown in Fig. \ref{fig:dimensional-dc-rnn-full}.}
\label{fig:dimensional-dc-rnn}
\vspace{-10pt}
\end{figure}

\begin{figure}[t]
\centering
\includegraphics[width=0.9\textwidth]{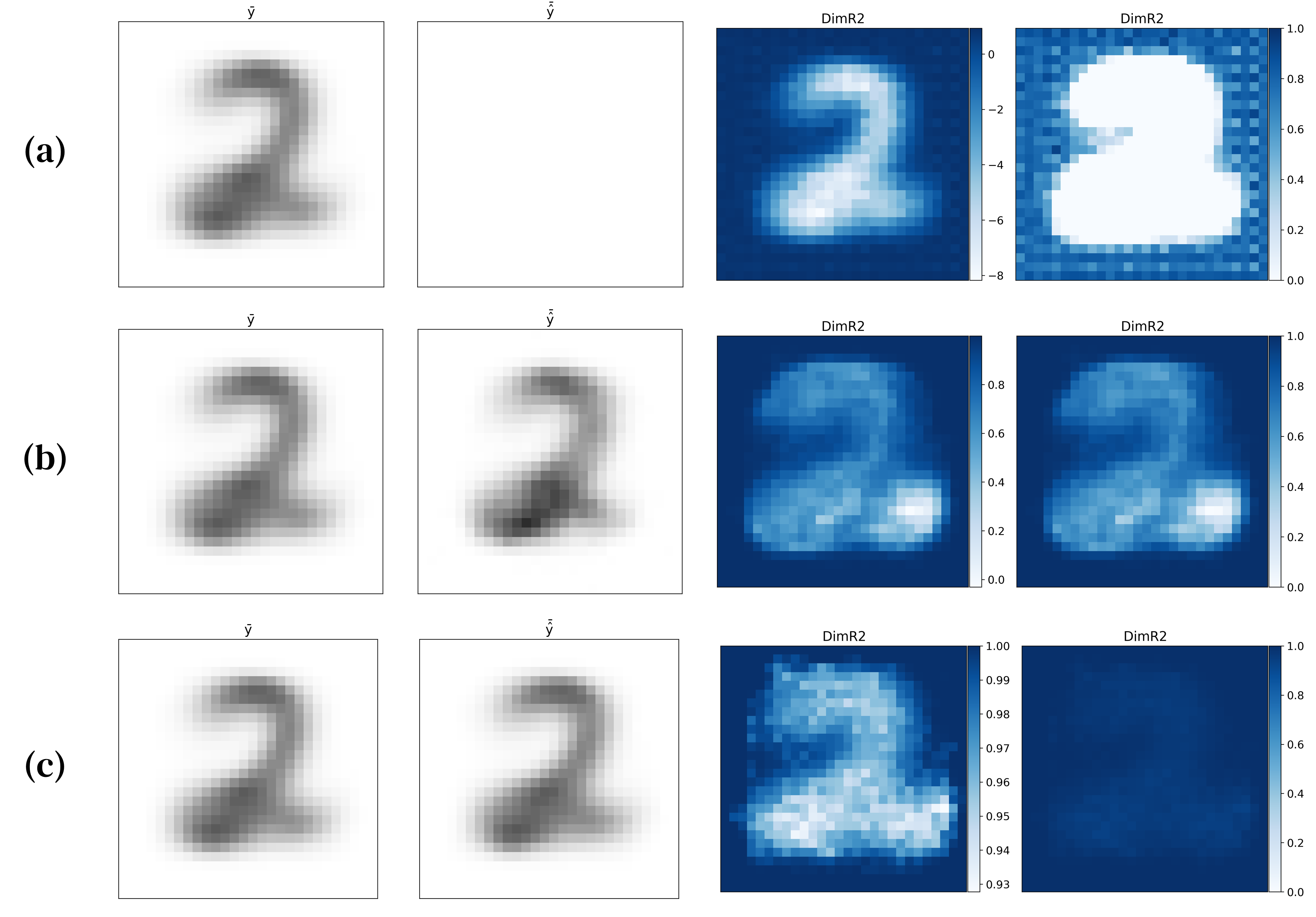}
\caption{Dimensional view of Dim-R2 on MNIST image reconstruction with $\mathcal{A}$=(Data, Channel), $\mathcal{A}_{norm}$=(Width, Height). Columns (left to right): $\bar{y}$, $\bar{\hat{y}}$, Dim-R2 (data min-max scale), Dim-R2 ([0,1] scale). Rows show VAE training iterations: (a) 0 (before training), (b) 600, and (c) 39,840 (after early stopping).}
\label{fig:dimensional-MNIST}
\vspace{-15pt}
\end{figure}

\begin{figure}[t!]
\centering
\includegraphics[width=0.9\textwidth]{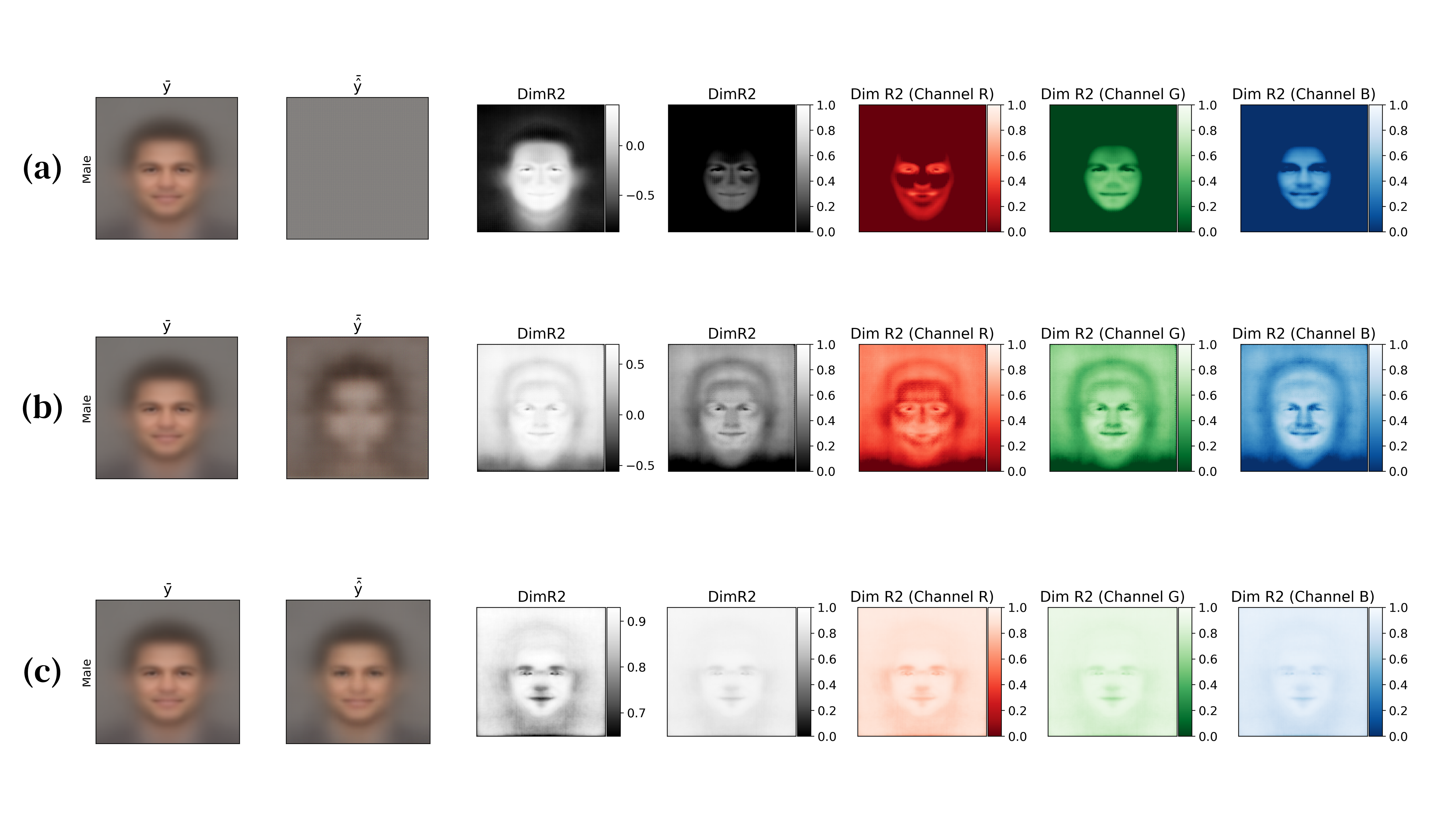}
\caption{Dimensional view of Dim-R2 on CelebA image reconstruction with $\mathcal{A}_{norm}$=(Width, Height). Columns (left to right): $\bar{y}$, $\bar{\hat{y}}$, Dim-R2 with $\mathcal{A}$=(Data, Channel) (data min-max scale), same Dim-R2 ([0,1] scale), per-channel Dim-R2 (Red, Green, Blue) with $\mathcal{A}$=Data ([0,1] scale). Rows show VAE training iterations: (a) 0 (before training), (b) 300, and (c) 39,280 (after early stopping).}
\label{fig:dimensional-celeba}
\vspace{-10pt}
\end{figure}

To demonstrate a use case of Dim-R2, we applied it to DC-RNN neural activity predictions of mouse motor circuit spiketrains for a single representative session (Section \ref{appendix:dc-rnn}, Fig. \ref{fig:sample-rnn_data}), with $y$ and $\hat{y}$ of shape (Random seed (RS), Cross validation folds (CV), Trial, Time, Neuron)=(3,3,78,249,252). Setting $\mathcal{A}$=(RS, CV, Trial) reveals accuracy across Time and Neuron dimensions. The DC-RNN captures trial variability well in DCN throughout most of the trial and in some M1 neurons after 100ms, which is the hand lift onset (Fig. \ref{fig:dimensional-dc-rnn}a). Measuring Dim-R2 with time-averaged $\bar{y}$ and $y$ reveals which time-neuron regions have high trial variability in $y$ (Fig. \ref{fig:dimensional-dc-rnn}b), as time-averaged $\bar{y}$ can only reflect trial variability. Setting $\mathcal{A}$=(RS, CV, Time) with $\mathcal{A}_{norm}$=Time measures accuracy relative to time-variability, across Trial and Neuron dimensions. Neurons that capture time variability well appear as vertical bands of high scores, while noisy trials appear as horizontal bands of low scores (Fig. \ref{fig:dimensional-dc-rnn}c). As expected, $\bar{y}$ yields R2=0 when $\mathcal{A}_{norm}=\mathcal{A}$=Time (Fig. \ref{fig:dimensional-dc-rnn}d).

To demonstrate additional use cases, VAEs were trained to reconstruct MNIST and CelebA images (Section \ref{appendix:vae}). For MNIST handwritten digit images of shape (Data, Channel, Width, Height)=(10000, 1, 28, 28), Dim-R2 with $\mathcal{A}$=(Data, Channel) and $\mathcal{A}_{norm}$=(Width, Height) yields a 2D score map measuring spatial variability (Fig. \ref{fig:dimensional-MNIST}). Measuring Dim-R2 across training iterations reveals which spatial features the model learns first. Before training, the initial prediction reflects data structure rather than yielding R2=0 since $\mathcal{A}\neq \mathcal{A}_{norm}$ (Fig. \ref{fig:dimensional-MNIST}a). After 600 gradient updates, the VAE largely reconstructs the digit "2" but the lower-right tail remains poorly captured, suggesting the model first learns spatial features shared across digits (Fig. \ref{fig:dimensional-MNIST}b). The fully trained VAE reconstructs the whole image at high scores (Fig. \ref{fig:dimensional-MNIST}c).

For CelebA images of shape (Data, Channel, Width, Height)=(19962, 3, 128, 128), two $\mathcal{A}$ settings were used across VAE training iterations: $\mathcal{A}$=(Data, Channel) for a 2D spatial score map, and $\mathcal{A}$=Data for per-channel (RGB) score maps, both with $\mathcal{A}_{norm}$=(Width, Height) to capture spatial variance (Fig. \ref{fig:dimensional-celeba}). Before training, the initial prediction reflects data structure rather than yielding R2=0 since $\mathcal{A}\neq \mathcal{A}_{norm}$ (Fig. \ref{fig:dimensional-celeba}a). After 300 gradient updates, the VAE reconstructs the background and hair but not the pupils, with the red channel showing lower scores around the skin area (Fig. \ref{fig:dimensional-celeba}b). After full training, the VAE achieves high scores across the image (Fig. \ref{fig:dimensional-celeba}c). Full visualizations of Dim-R2 across training for both MNIST and CelebA are provided in the Supplemental File (\ref{file:dim-R2-image-reconstruction}). These dimensional views reveal both the structure of $y$ and the prediction patterns of models, guiding the modeling process.


\subsection{Dim-R2 better reflects high-accuracy channels than Mean R2 in the presence of noise channels}
\label{results:resilience}

\begin{figure}[t]
\centering
\subfloat[]{\includegraphics[width=0.45\textwidth]{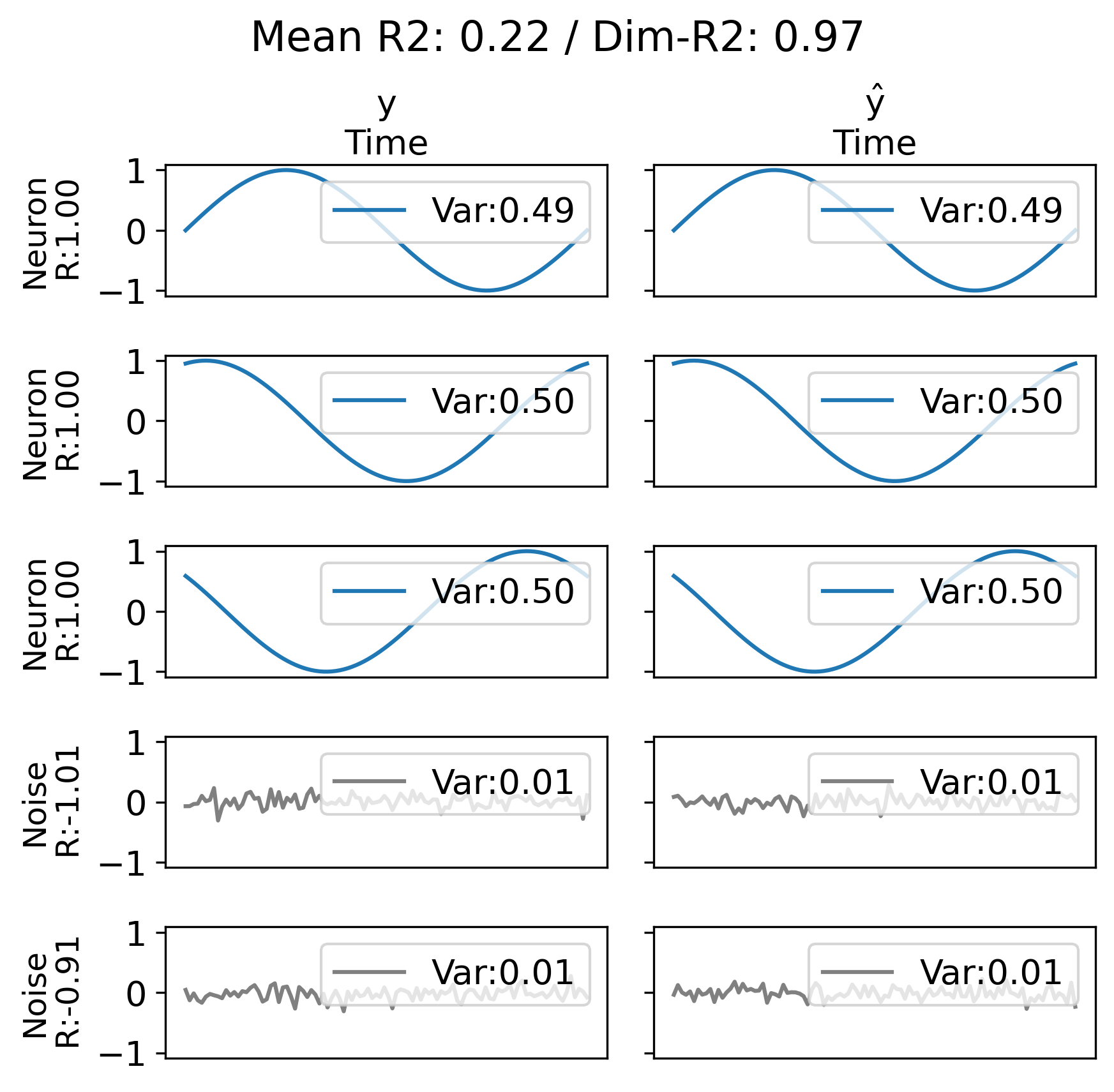}}  
\subfloat[]{\includegraphics[width=0.45\textwidth]{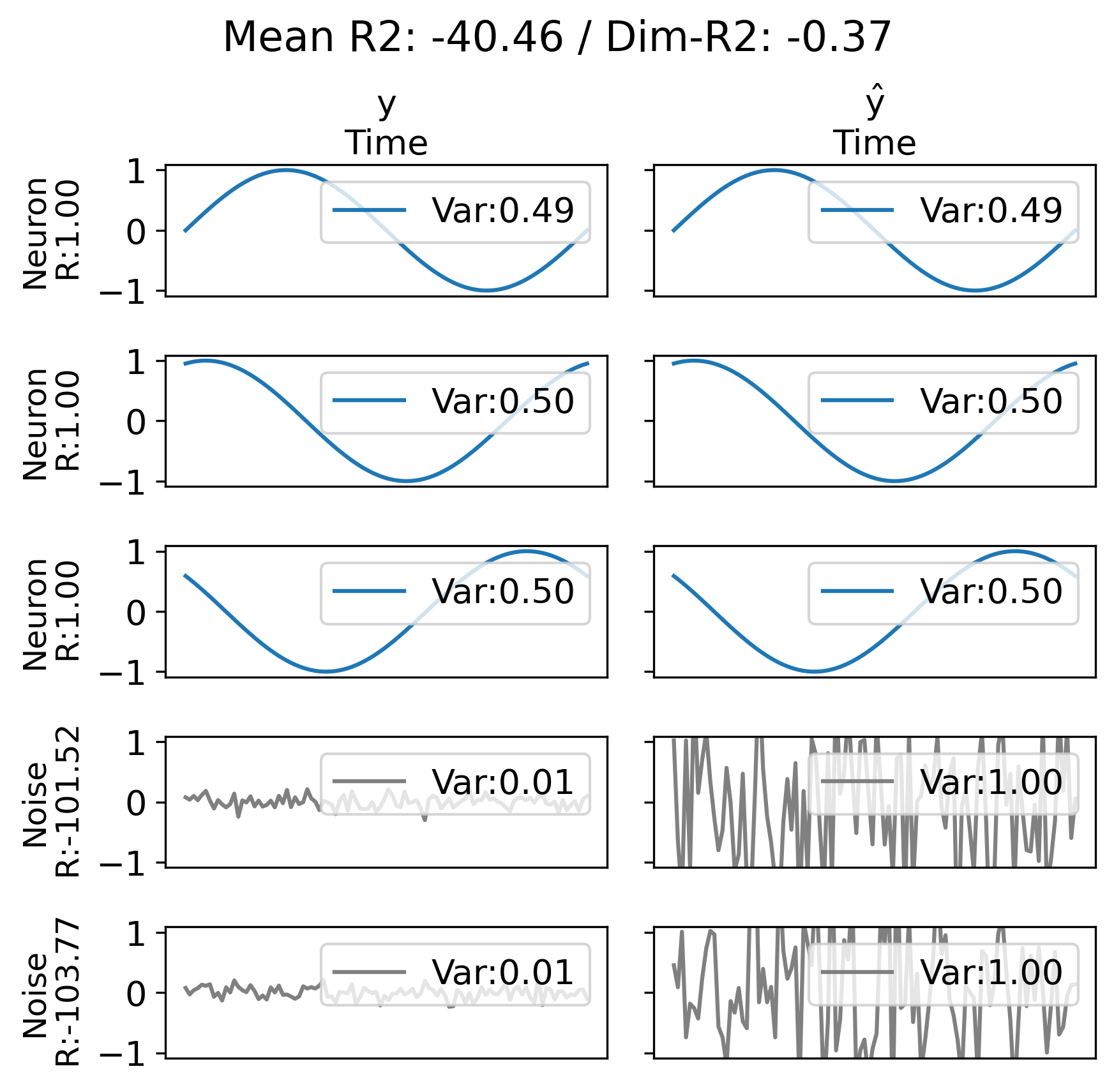}}
\caption{Synthetic sinusoidal dataset $y$ and $\hat{y}$ with noise channels, with corresponding mean R2 and Dim-R2 scores. (a) Noise channel variance: $y$=0.01, $\hat{y}$=0.01, (b) $y$=0.01, $\hat{y}$=1.00. Full combinations of noise channel variances in Fig. \ref{fig:r2-sample-full}.}
\label{fig:r2-sample}
\vspace{-10pt}
\end{figure}

\begin{figure} [t]
    \centering
    \includegraphics[width=0.9\textwidth]{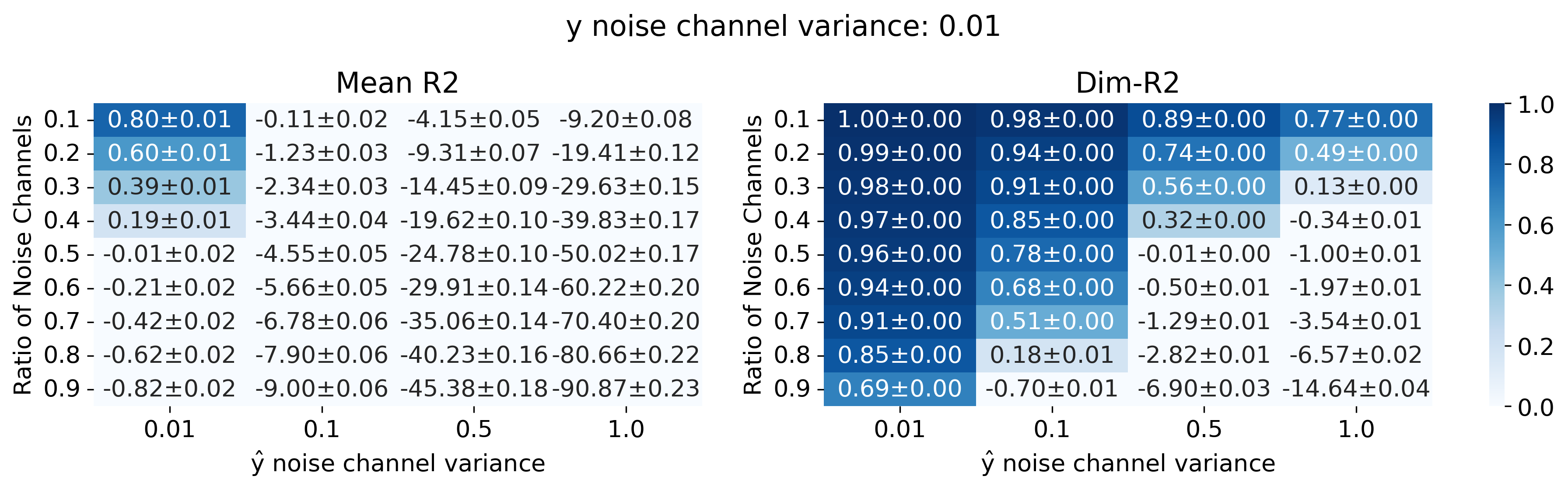}
    \caption{Dim-R2 yields higher scores than mean R2 in the presence of low-variance noise channels. Scores measured on synthetic sinusoidal dataset with noise channels (Fig. \ref{fig:r2-sample}) across $y$ and $\hat{y}$ noise channel variances. Each entry shows mean$\pm$std across 100 random repetitions.}
    \label{fig:noise_resilience}
\end{figure}

To demonstrate Dim-R2's noise resilience, we applied it to the synthetic sinusoidal dataset with noise channels (Section \ref{datasets:synthetic}, Fig.  \ref{fig:r2-sample}, \ref{fig:r2-sample-full}). When $y$ noise channel variance is low (0.01) relative to signal channels (0.5), Dim-R2 yields substantially higher scores than mean R2, better highlighting high-accuracy channels (Fig. \ref{fig:noise_resilience}). This advantage decreases as $y$ noise channel variance increases toward and beyond signal channel variance (Fig. \ref{fig:noise_resilience_full}). Both mean R2 and Dim-R2 decrease as $\hat{y}$ noise channel variance increases. Additional metrics from Table \ref{tab:regression-metric-comparison} are evaluated for reference (Fig. \ref{fig:resilience_d2_ae}-\ref{fig:resilience_corr}).

\begin{figure}[t]
    \centering
    \includegraphics[width=0.8\textwidth]{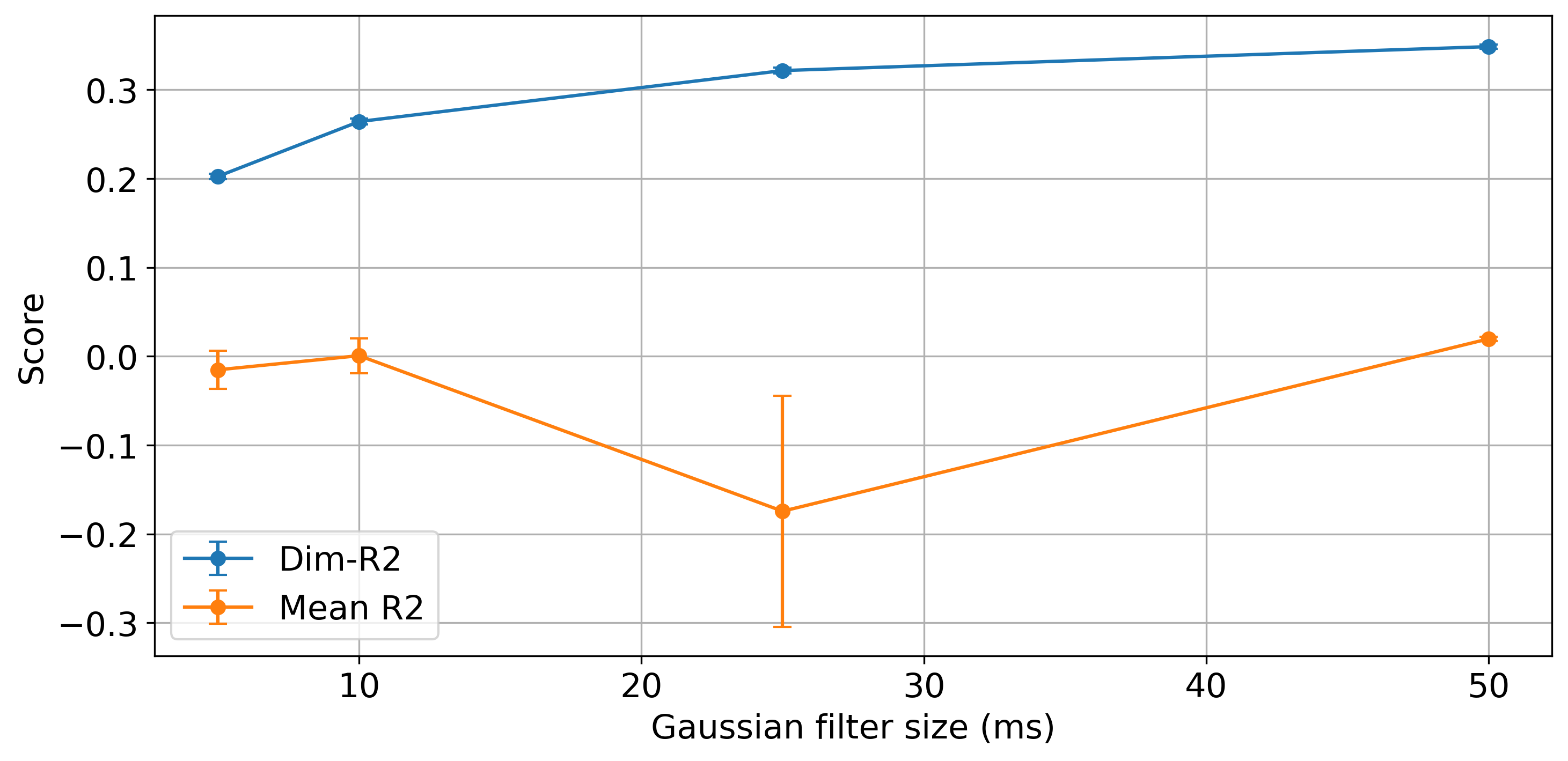}
    \caption{Dim-R2 and mean R2 on DC-RNN neural data predictions across preprocessing Gaussian filter sizes. Dim-R2 better reflects the effect of filter size in the presence of noise. Each point shows mean$\pm$standard error across 22 sessions, 3 random seeds, and 15 cross-validation folds.}
    \label{fig:noise_resilience_dcrnn}

\end{figure}

To demonstrate noise resilience on real data, Dim-R2 and mean R2 were measured on DC-RNN predictions across recording sessions and Gaussian filter sizes used in preprocessing (Fig. \ref{fig:noise_resilience_dcrnn}). Scores increase with larger filter sizes as smoother $y$ and $\hat{y}$ reduce noise. This trend is clearer in Dim-R2 than mean R2, with mean R2 yielding lower scores and larger standard error due to low-variance noise channels. For example, Sessions 3 and 12 at 25ms filter size yield large negative mean R2 --- an outlier effect absent in Dim-R2 (Fig. \ref{fig:noise_resilience_dcrnn_heatmap}). This resilience is also apparent in individual $y$ and $\hat{y}$ pairs for both trial-averaged and single-trial data (Fig. \ref{fig:sample-rnn_data}, Table \ref{tab:sample-resilience}). Dim-R2 thus allows users to assess hyperparameter effects on modeling without confounding influence of noise.


\section{Discussion}
Dim-R2 accepts data of arbitrary dimensionality, enables dimensional evaluation of regression accuracy, and better highlights high-variance signal channels in the presence of noise. It provides a bird's-eye view of regression accuracy and a noise-resilient score for reliable hyperparameter exploration. These benefits make Dim-R2 broadly applicable to multidimensional regression tasks such as image and signal reconstruction, and data-driven computational modeling \citep{huang2024measuring, yoo2024dual, BADRULHISHAM2024470, zador2023catalyzing, yang2020artificial, lin2023temporal, heckel2024deep, ahmed2025comprehensive,10.1093/jrsssb/qkad073}.

When using Dim-R2 to assess accuracy across dimensions, careful consideration of $\mathcal{A}$, $\mathcal{A}_{norm}$, and $\mathcal{A}_{pool}$ is essential, especially when they differ. In such cases, a Dim-R2 score of 0 does not imply mean prediction performance but may instead reflect data structure (Fig. \ref{fig:dimensional-synthetic}d,h; \ref{fig:dimensional-dc-rnn-full}d). For correct interpretation, comparing against domain-specific controls such as time-averaged $\bar{y}$ is important, as is comparing different $\mathcal{A}_{norm}$ settings to measure different dimensions of variability.

When collapsing all dimensions to yield a single score, Mean R2 and Dim-R2 weigh channel variances differently. Mean R2 is sensitive to small-variance channels since all channels are weighted equally, while Dim-R2 is sensitive to high-variance channels since they contribute more to the normalizing term TSS. A metric sensitive to high-variance better detects high-variance signals while reducing the effect of low-variance noise, and vice versa. In natural signals such as images, audio, and neural activity where signal variance typically exceeds noise variance, Dim-R2 is more suitable for detecting high-accuracy channels amidst noise.

The three benefits of Dim-R2 arise from applying a multidimensional generalization to metrics that average per-channel measurements of the form $1-Error/Normalization$. Metrics with this structure --- such as explained variance (not a regression metric) and D2 absolute error (Appendix \ref{appendix:common-regression-metrics}) --- inherit the same three limitations of conventional R2, and could be extended using the same multidimensional generalization by broadcasting the error and normalization terms.

One fundamental limitation of Dim-R2 inherited from conventional R2, is sensitivity to low-variance $y$. When $y$ has small variance, it contributes to a small TSS which amplifies prediction errors into low scores, even when $\hat{y}$ closely matches $y$ in absolute terms. In such cases, the normalizing term should reflect a domain-appropriate expected range rather than the variance of $y$, such as the maximum range of pixel values ($[0,1]$) for image data or the maximum firing rate range ($[0,1kHz]$) for neural activity.

\begin{ack}

\end{ack}


\bibliography{references}
\bibliographystyle{unsrt}


\newpage
\appendix

\section{Common regression metrics}
\label{appendix:common-regression-metrics}

The following are the definitions of several commonly used regression metrics referenced in the Related Works section. While explained variance and correlation are not regression metrics, they are included as they give insights related to regression.

\paragraph{Mean Squared Error (MSE).}
\begin{equation}
\mathrm{MSE} = \frac{1}{N} \sum_{i=1}^{N} (y_i - \hat{y}_i)^2.
\end{equation}
MSE penalizes large errors quadratically but is not normalized to the scale of the data and does not provide a reference baseline such as predicting the mean.

\paragraph{Mean Absolute Error (MAE).}
\begin{equation}
\mathrm{MAE} = \frac{1}{N} \sum_{i=1}^{N} |y_i - \hat{y}_i|.
\end{equation}
MAE is robust to outliers compared to MSE but similarly lacks normalization and does not capture variance.

\paragraph{D2 Absolute Error (D2-MAE).}
\begin{equation}
\mathrm{D2\text{-}MAE}
= 1 - \frac{\sum_{i=1}^{N} |y_i - \hat{y}_i|}
               {\sum_{i=1}^{N} |\,y_i - \bar{y}\,|}.
\end{equation}
D2-MAE normalizes absolute error relative to the deviation of $y$ from its mean, producing a score in $(-\infty, 1]$ with interpretable anchors (1 = perfect prediction; 0 = equivalent to predicting $\bar{y}$). However, because it uses absolute error, it does not capture variance.

\paragraph{Explained Variance (EV).}
\begin{equation}
\mathrm{EV} 
= 1 - \frac{\mathrm{Var}(y - \hat{y})}{\mathrm{Var}(y)}.
\end{equation}
Explained variance measures how well the fluctuations of $\hat{y}$ match those of $y$, but does not penalize additive bias (e.g., $\hat{y} = y + b$ yields $\mathrm{EV} = 1$).

\paragraph{Correlation (Corr).}
\begin{equation}
\mathrm{Corr}
= \frac{\mathrm{Cov}(y, \hat{y})}{\sigma_y \sigma_{\hat{y}}}.
\end{equation}
Pearson correlation quantifies linear co-variation between $y$ and $\hat{y}$ and is scale-invariant, but does not measure prediction error nor bias.

\section{Properties of Dim-R2}
\label{appendix:properties-of-dim-r2}
Dim-R2 inherits the same bounds and invariances from conventional R2.

\subsection{Bounds of Dim-R2}
By definition,
\begin{equation}
    Dim\text{-}R2=1-\frac{RSS}{TSS}
\end{equation}

where
\begin{equation}
    RSS = \sum_{k \in \mathcal{A}} (y_{k} - \hat{y}_{k} )^{2} \ge0,
    \qquad
    RSS \in \mathbb{R}^{\mathcal{D} \setminus \mathcal{A} }
\end{equation}

For each entry of the resulting tensor in $\mathbb{R}^{\mathcal{D} \setminus \mathcal{A}}$:

- $RSS=0$ if and only if $\hat{y}=y$ along the axes in $\mathcal{A}$ yielding Dim-R2=1.

- $RSS=TSS$ when $\hat{y}$ along the axes in $\mathcal{A}$ equals $\bar{y}$ along the axes in $\mathcal{A}_{norm}$, yielding Dim-R2=0.

- $RSS>TSS$ yields negative values of Dim-R2.

Thus, for every entry,
\begin{equation}
Dim\text{-}R2 \in (-\infty, 1]
\end{equation}

\subsection{Invariance}
Dim-R2 is invariant to affine transformations of both $y$ and $\hat{y}$. Let $a,b \in \mathbb{R}$, $a \neq0$ and define

\begin{equation}
y' = a y + b, 
\qquad 
\hat{y}' = a \hat{y} + b.
\end{equation}
RSS and TSS transform as 
\begin{equation}
RSS(y',\hat{y}') 
= a^{2}\,RSS(y,\hat{y}), 
\qquad
TSS(y') 
= a^{2}\,TSS(y).
\end{equation}
Since the scaling factor $a^{2}$ cancels in the ratio, we obtain
\begin{equation}
Dim\text{-}R^{2}(y',\hat{y}')
= 1 - \frac{RSS(y',\hat{y}')}{TSS(y')}
= 1 - \frac{a^{2}RSS(y,\hat{y})}{a^{2}TSS(y)}
= Dim\text{-}R^{2}(y,\hat{y}).
\end{equation}

For the degenerate case of $a=0$, the transformed target becomes constant,
\begin{equation}
y' = \hat{y}' = b,
\end{equation}
so both RSS and TSS vanish:
\begin{equation}
\qquad RSS(y',\hat{y}') = TSS(y') = 0.
\end{equation}
Following the standard convention for R2 when $TSS=0$, the score is defined as $1$ whenever the prediction equals the target. Therefore, 
\begin{equation}
Dim\text{-}R^{2}(y',\hat{y}')
= 1.
\end{equation}

Thus, Dim-R$^{2}$ is invariant under any non-zero affine transformation.

\subsection{Reduction to conventional R2}

Dim-R2 can reduce to conventional R2 under specific argument choices.

\begin{enumerate}
    \item \textbf{1D case} 
    
    Let $y,\hat{y} \in \mathbb{R}^N$ with observation indices $i=1, ...,N$. Let the full set of axes be $\mathcal{D}=\{1\}$.
    
    \begin{equation}
    Mean\ R2 = 1-\frac{\sum_{i \in  }{(y_{i}-\hat{y}_{i})^2}}{\sum_{i}{(y_{i}-\overline{y})^2}}
    \label{eq:mean-r2-1d}
    \end{equation}
    
    For Dim-R2, choose $\mathcal{A}=\mathcal{A}_{norm}=\mathcal{A}_{pool}=\{1\}$.
    
    \begin{equation}
    Dim\text{-}R2 = 1-\frac{\sum_{i}{(y_{i}-\hat{y}_{i})^2}}{\sum_{i}{(y_{i}-\overline{y})^2}}
    \label{eq:dim-r2-1d}
    \end{equation}

    \item \textbf{2D case reducing Dim-R2 to a single score} 

    Let $y,\hat{y} \in \mathbb{R}^{N \times C }$, with observation indices $i=1,...,N$, and channel indices $c=1,...,C$. Let the full set of axes be $\mathcal{D}=\{1,2\}$, where axis 1 is the observation dimension and axis 2 is the channel dimension. 
    
    \begin{equation}
    Var_c=\sum_{i}{(y_{ic}-\overline{y}_{c}})^2
    \end{equation}
    
    \begin{align}
    Weighted\ Mean\ R2 &=\frac{1}{\sum_{c}{Var_{c}} }\sum_{c} Var_{c} * R2_{c} \\
    &= \frac{1}{\sum_{c}{Var_{c}} }\sum_{c} Var_{c} [1-\frac{\sum_{i}{(y_{i,c}-\hat{y}_{i,c})^2}}{Var_c}]
    \label{eq:mean-r2-2d}\\
    &= \frac{1}{\sum_{c}{Var_{c}}}[{\sum_{c}{Var_{c}}}-\sum_{c}{\sum_{i}{(y_{i,c}-\hat{y}_{i,c})^2}}]\\
    &=1-\frac{\sum_{c}{\sum_{i}{(y_{i,c}-\hat{y}_{i,c})^2}}}{\sum_{c}{Var_{c}}}
    \end{align}

    For Dim-R2, choose $\mathcal{A}=\{1,2\}$ and $ \mathcal{A}_{norm}=\mathcal{A}_{pool}=\{1\}$ so that the reference variance (TSS) is computed along the observation axis.

    \begin{equation}
    \bar{y}_c=\frac{1}{N}
    \sum_{i} y_{i,c} \qquad \bar{y} \in \mathbb{R}^{C}
    \end{equation}
    
    \begin{align}
    Dim\text{-}R2 &= 1- \frac{RSS}{TSS} \\
    &=1 - \frac{\sum_{k \in \mathcal{A}}{(y_k-\hat{y}_k )^2} }{\frac{1}{|\mathcal{A}_{pool} \setminus \mathcal{A}|}
    \sum_{j \in \mathcal{A}_{pool} \setminus \mathcal{A}}
    \sum_{k \in \mathcal{A}}
    (y_{k,j} - \bar{y})^{2}} \\
    &= 1- \frac{\sum_{k \in \mathcal{A}}{(y_k-\hat{y}_k )^2} }{\sum_{k \in \mathcal{A}}
    (y_{k} - \bar{y})^{2}} \\
    &= 1- \frac{\sum_{c}{\sum_{i}{(y_{i,c}-\hat{y}_{i,c} )^2}}}{\sum_{c}\sum_{i}{(y_{i,c} - \bar{y}_c)^2}
    }\\
    &= 1- \frac{\sum_{c}{\sum_{i}{(y_{i,c}-\hat{y}_{i,c} )^2}}}{\sum_{c}Var_c}
    \label{eq:dim-r2-2d}
    \end{align}

    Thus, collapsing all dimensions ($\mathcal{A}=\mathcal{D}$) while measuring per-channel variance across the observation dimension ($\mathcal{A}_{norm}={1}$) reduces Dim-R2 to the conventional variance weighted mean R2. When there are no mean shifts in $y$ or $\hat{y}$ across the channel dimension (i.e. $\bar{y}_c=\mu$ for all $c \in \{1,...,C\}$), choosing $\mathcal{A}= \mathcal{A}_{norm} =\mathcal{D}$ also yields the same values for variance weighted R2 and Dim-R2.
    
    \item \textbf{3D and higher case reducing Dim-R2 to a single score} 
    
    The Conventional R2 score does not naturally extend to $y,\hat{y} \in \mathbb{R}^d$ for $d\ge3$ in a way that preserves its interpretation, and therefore cannot be directly compared to Dim-R2.
    
\end{enumerate}

\newpage

\section{Dim-R2 Figures}
\begin{figure}[h]
    \centering
    \includegraphics[width=0.73\textwidth]{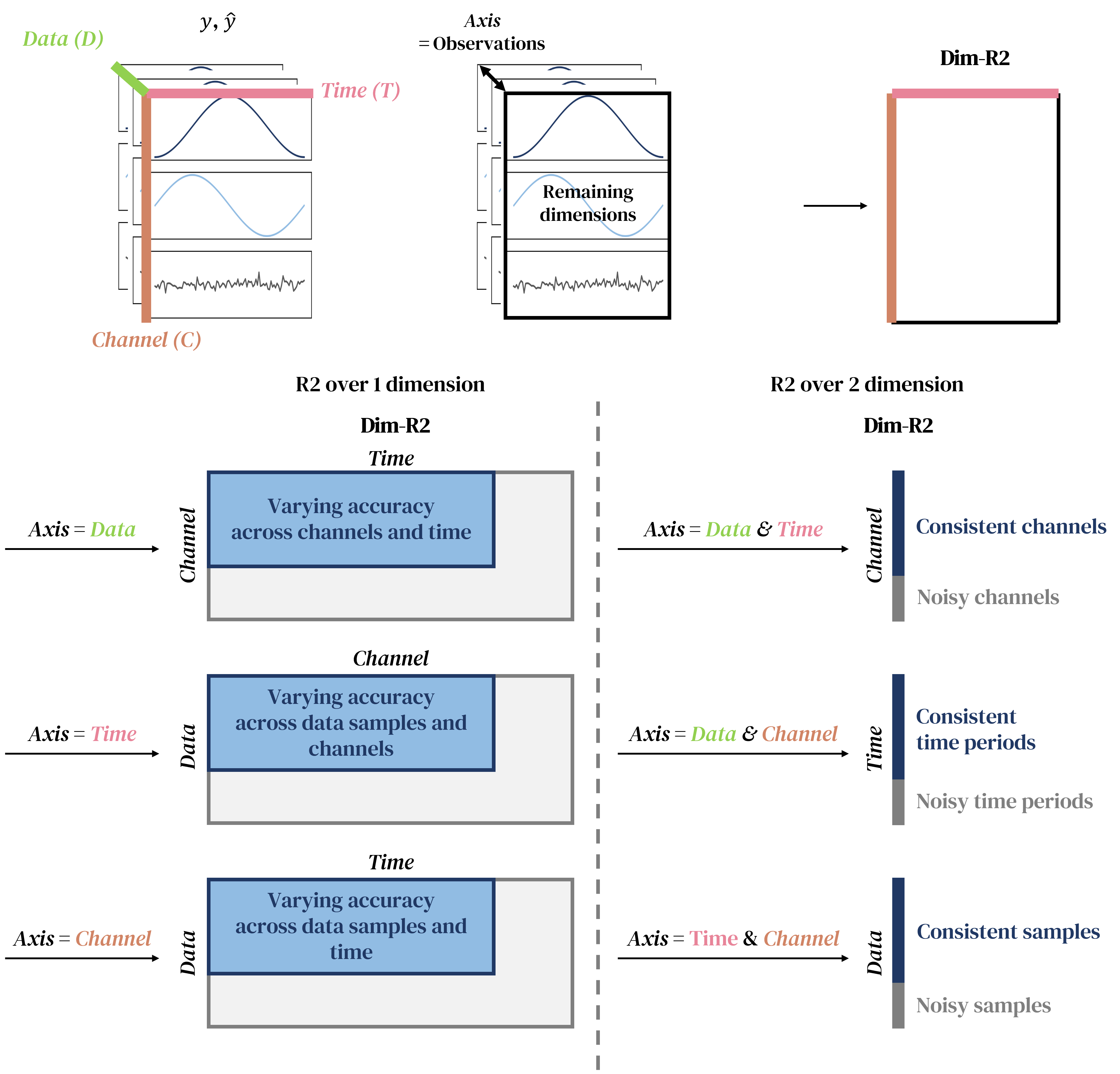}
    \caption{Dim-R2 presents rich patterns of prediction accuracy across designated dimensions (Axis).}
    \label{fig:Dim-R2 dimensional view}
\end{figure}

\begin{figure}[h]
    \centering
    \includegraphics[width=0.75\textwidth]{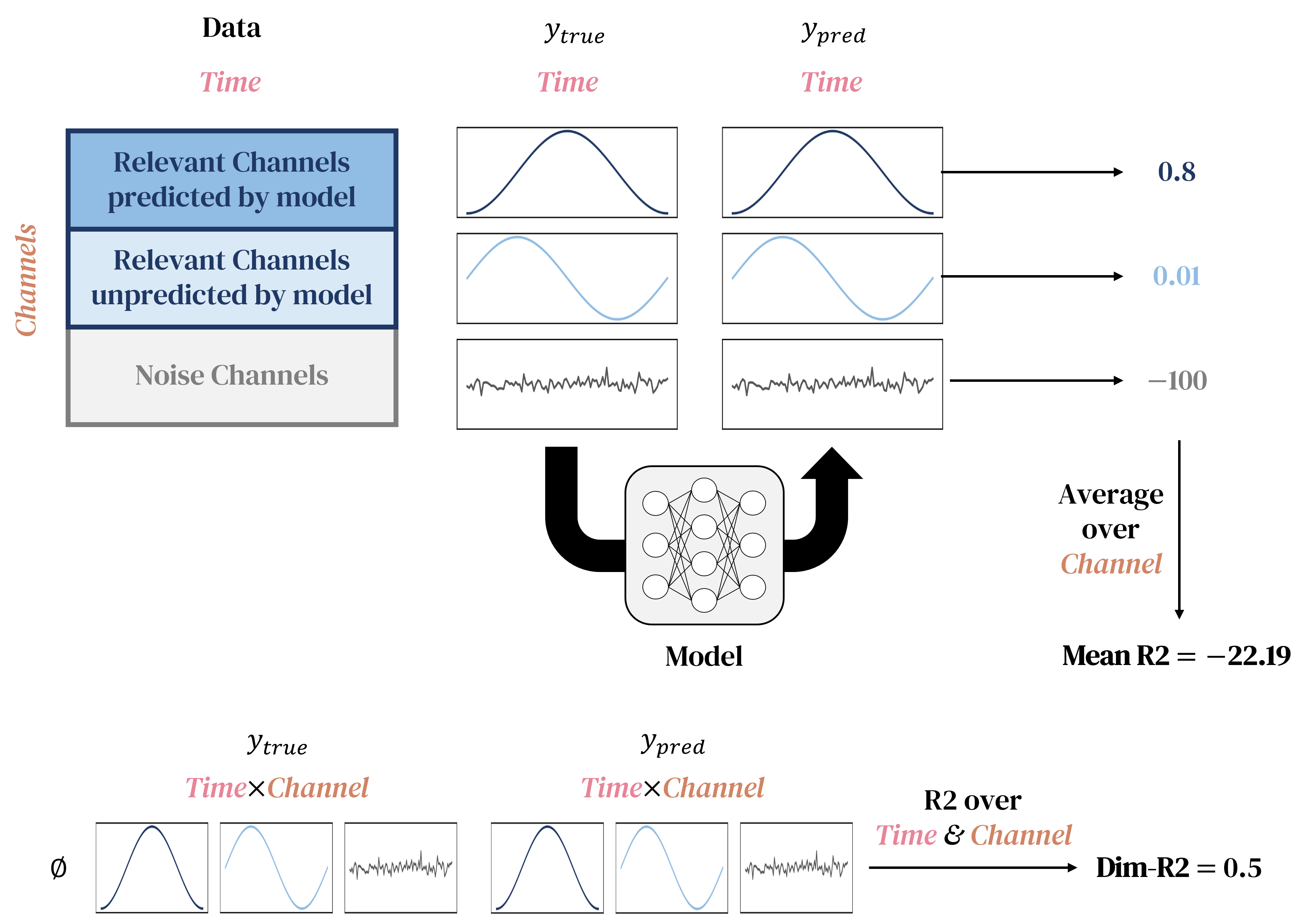}
    \caption{Dim-R2 is more resilient to noise channels than conventional mean R2 because high-variance signal channels dominate the influence of low-variance noise channels. As a result, Dim-R2 presents the existence of high accuracy channels among noise channels.}
    \label{fig:Dim-R2 resilience}
\end{figure}

\FloatBarrier

\begin{figure}[h]
    \centering
    \includegraphics[width=\textwidth]{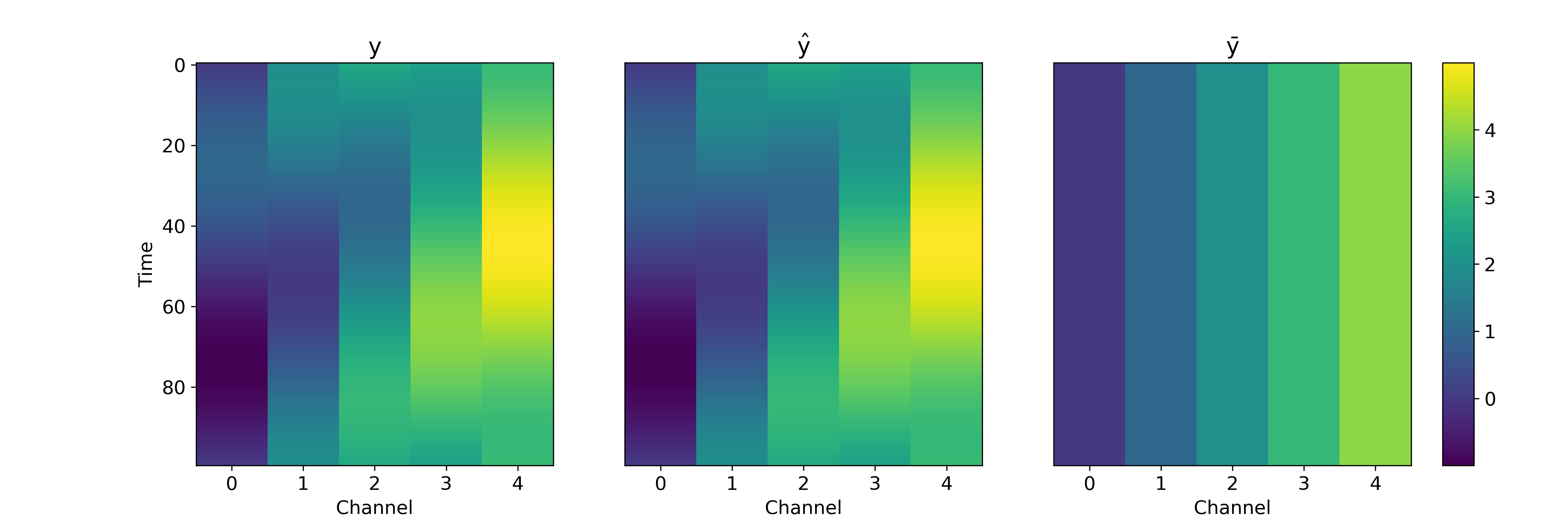}
    \caption{Simple 2D sine wave example to illustrate how $\mathcal{A}_{norm}$ controls which dimensional variability Dim-R2 measures against. Each channel shows $y_{c}=\sin(\frac{2\pi}{100}x+\frac{2\pi}{5}c)+c$, a time-varying sine wave with channel-varying phase and baseline.}
    \label{fig:Dim-R2 example data}
\end{figure}

\begin{table}[h]
    \centering
    \caption{Dim-R2 scores for $\hat{y}$ and $\bar{y}$ under different $\mathcal{A}_{norm}$ settings with $\mathcal{A}$=All dimensions, computed on data from Fig. \ref{fig:Dim-R2 example data}. $\bar{y}$ yields 0 only when $\mathcal{A}_{norm}$ matches the averaging dimension of $\bar{y}$ (Time).}
    \begin{tabular}{cccc}
    \toprule
    Data
    & 
    $\mathcal{A}_{norm}=Time$
    &
    $\mathcal{A}_{norm}=Channel$
    &
    \textbf{$\mathcal{A}_{norm}=Time, Channel$} 
    \\
    \midrule
    $\hat{y}$    & 1 & 1 & 1 \\
    $\bar{y}$    & 0 & 0.8 & 0.8 \\
    \bottomrule
    \end{tabular}
    \label{tab:dim-r2-example}
\end{table}

\FloatBarrier

\section{Datasets}
\label{appendix:datasets}

\subsection{Synthetic sinusoidal dataset}

\begin{figure}[h]
\centering
\includegraphics[width=0.9\textwidth]{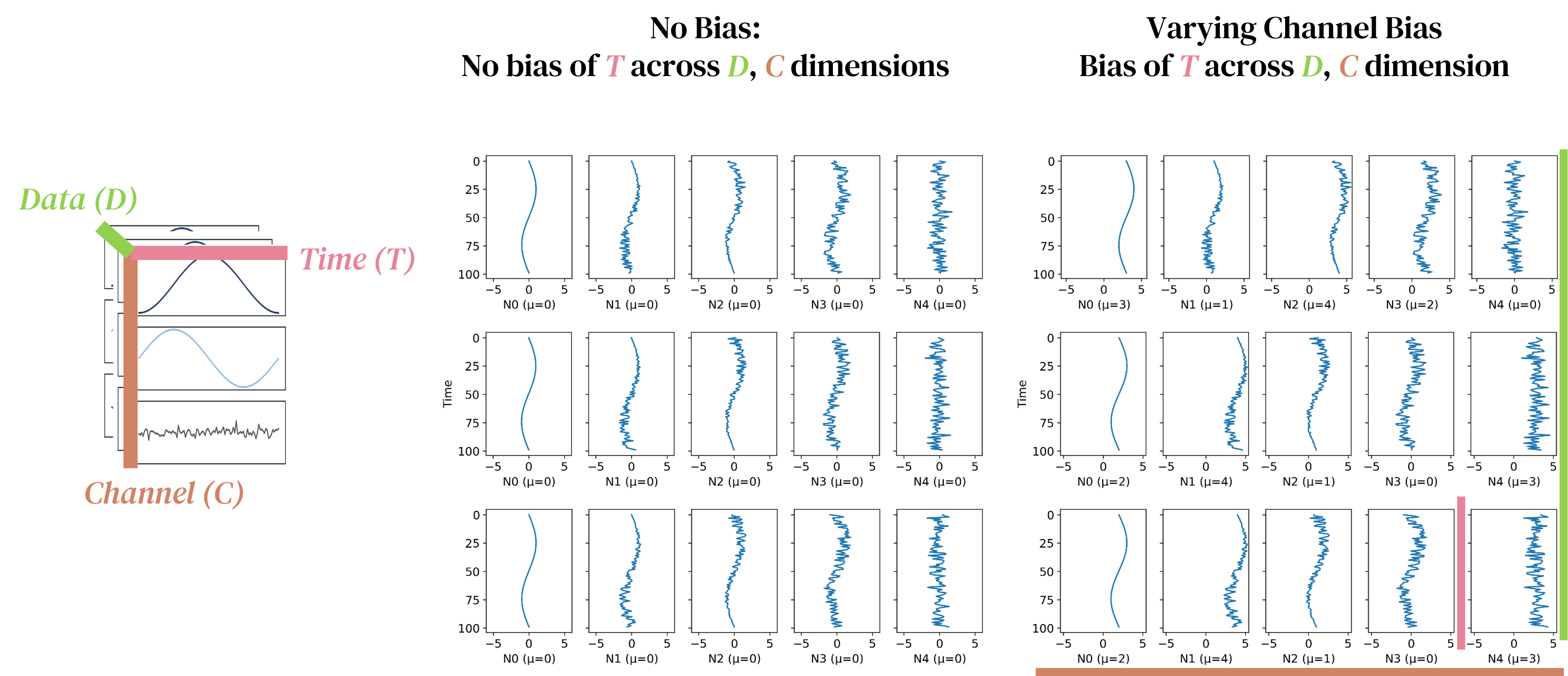}

\caption{Synthetic sinusoidal datasets with time-varying noise ($y,\hat{y} \in \mathbb{R}^{D \times T \times C}$) under No Bias and Varying Channel Bias conditions. Added time biases introduce variability across data and channel dimensions. Color-coded lines in the bottom right indicate dimension types.}
\label{fig:sample-time-varying-full}
\end{figure}

\begin{figure}[h]
\centering
\subfloat[]{\includegraphics[width=0.33\textwidth]{figures/r2_score_0.01_0.01.png}}
\subfloat[]{\includegraphics[width=0.33\textwidth]{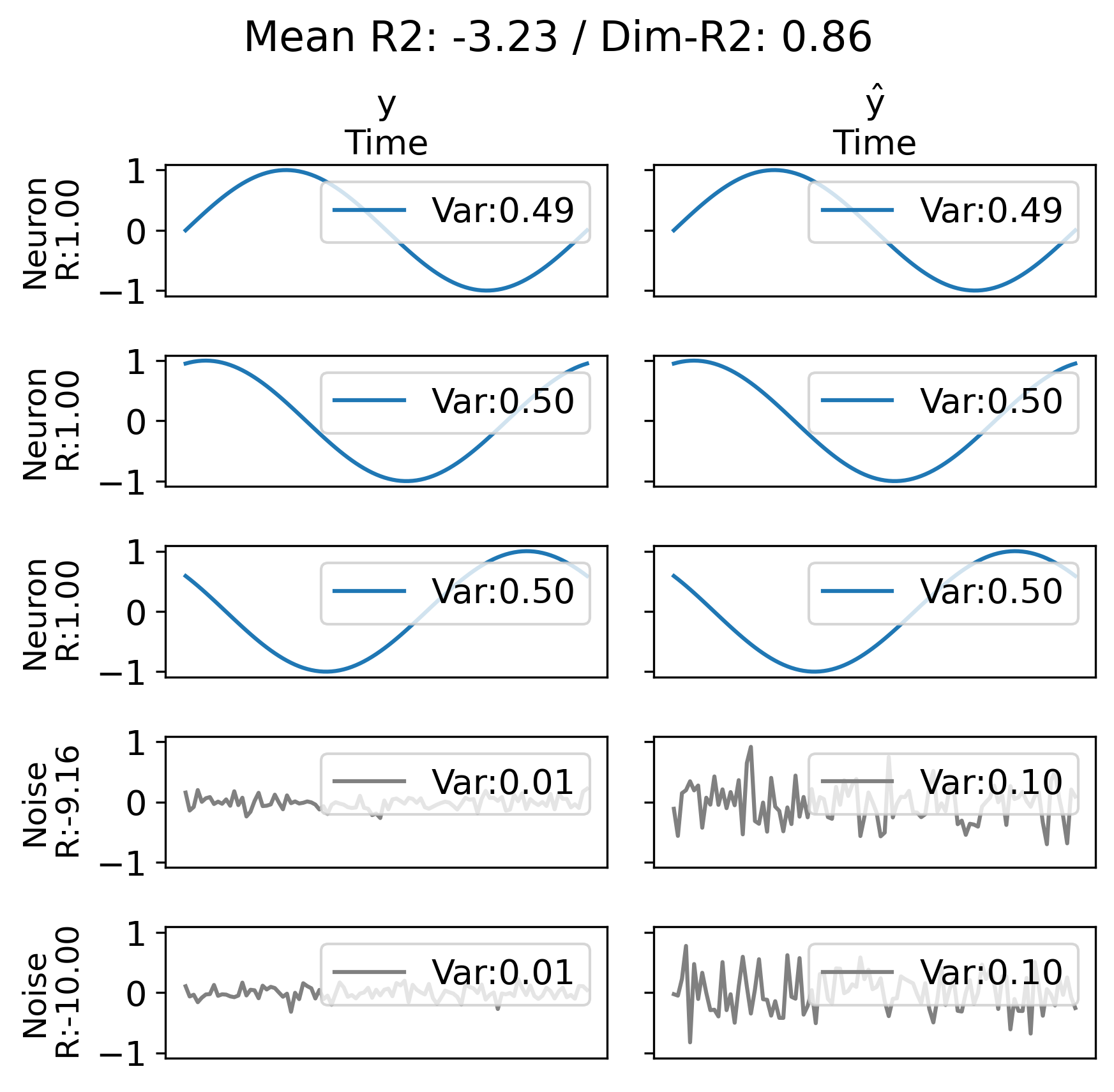}}
\subfloat[]{\includegraphics[width=0.33\textwidth]{figures/r2_score_0.01_1.0.png}}
\\
\subfloat[]{\includegraphics[width=0.33\textwidth]{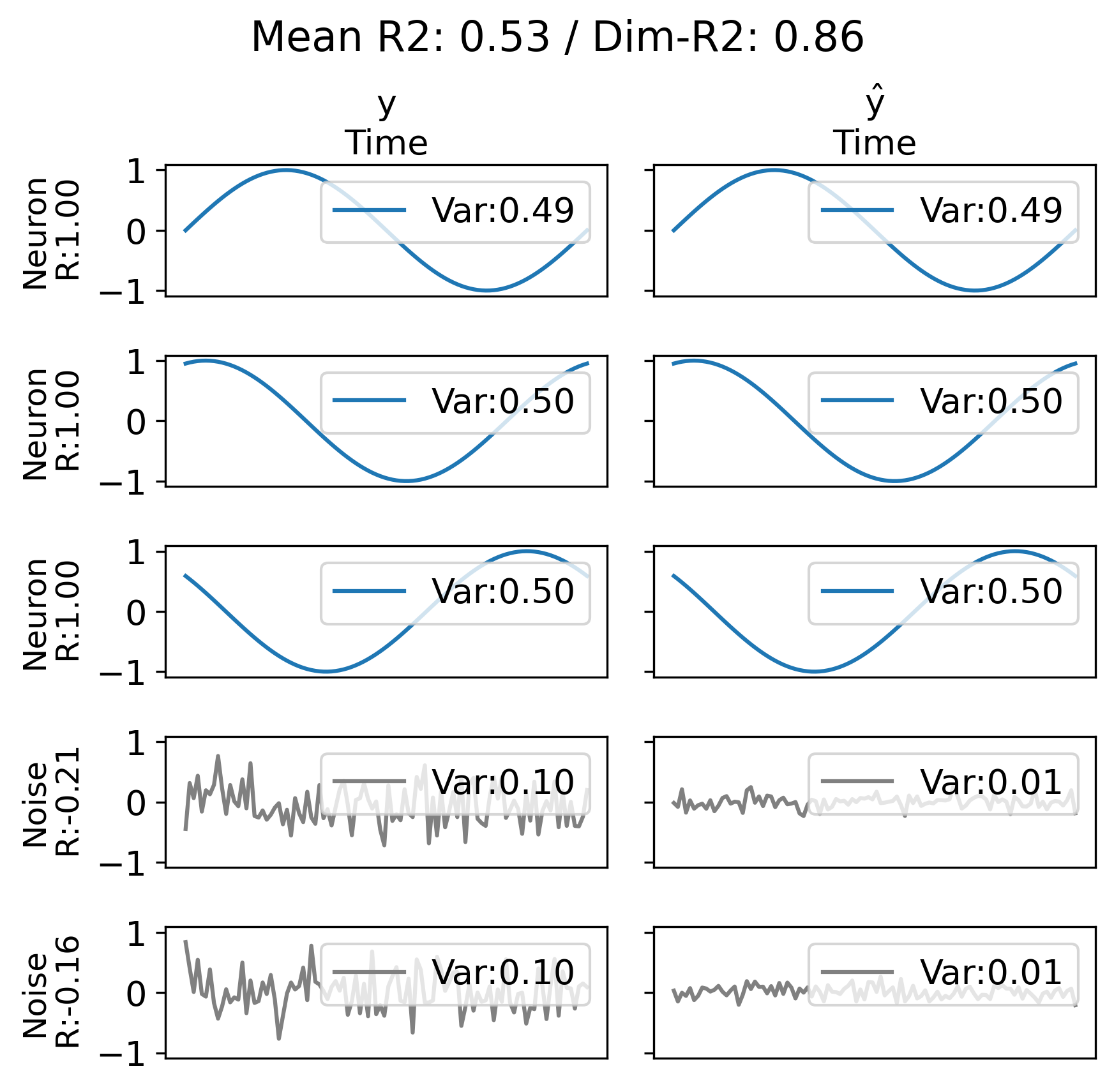}}
\subfloat[]{\includegraphics[width=0.33\textwidth]{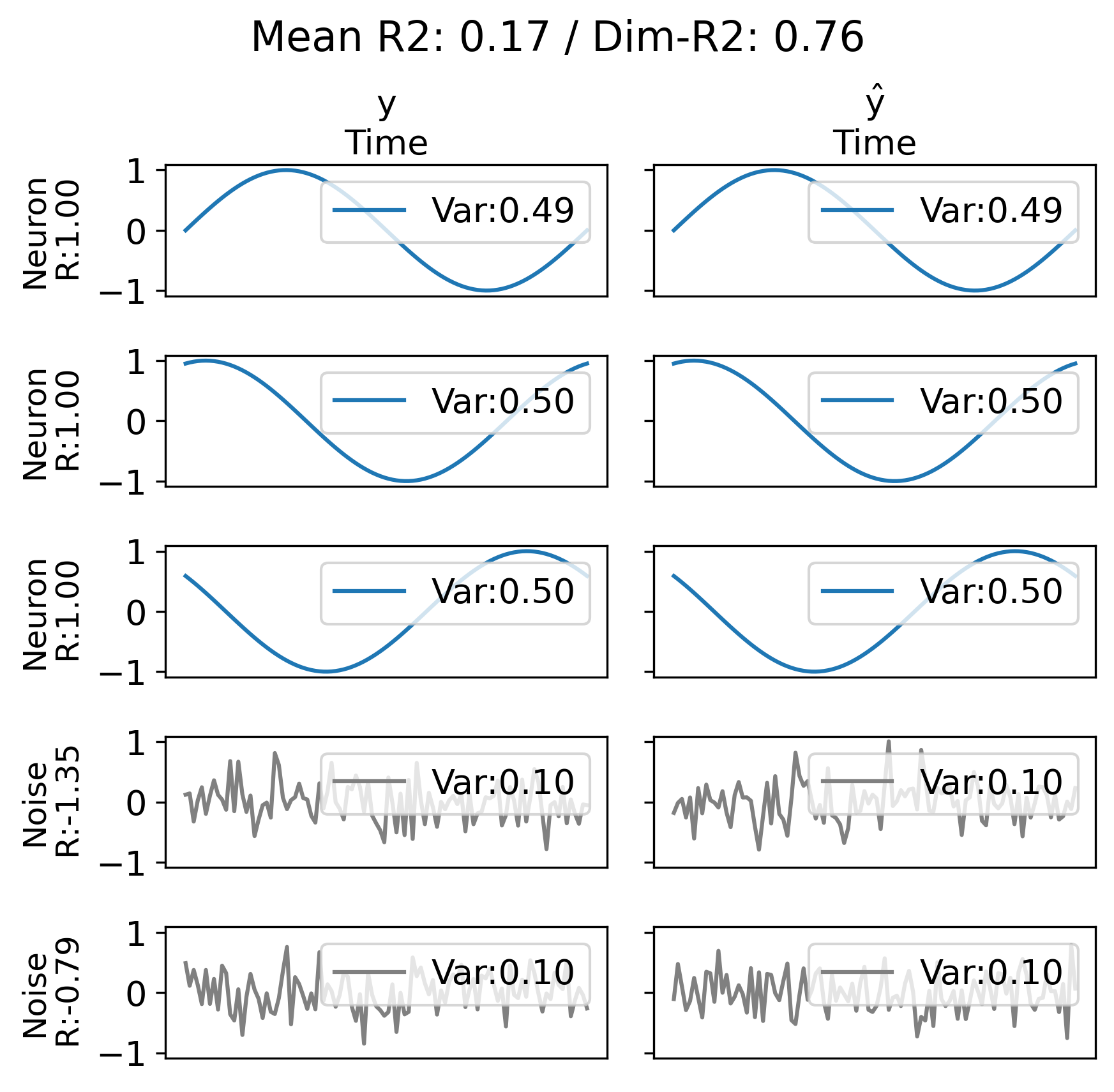}}
\subfloat[]{\includegraphics[width=0.33\textwidth]{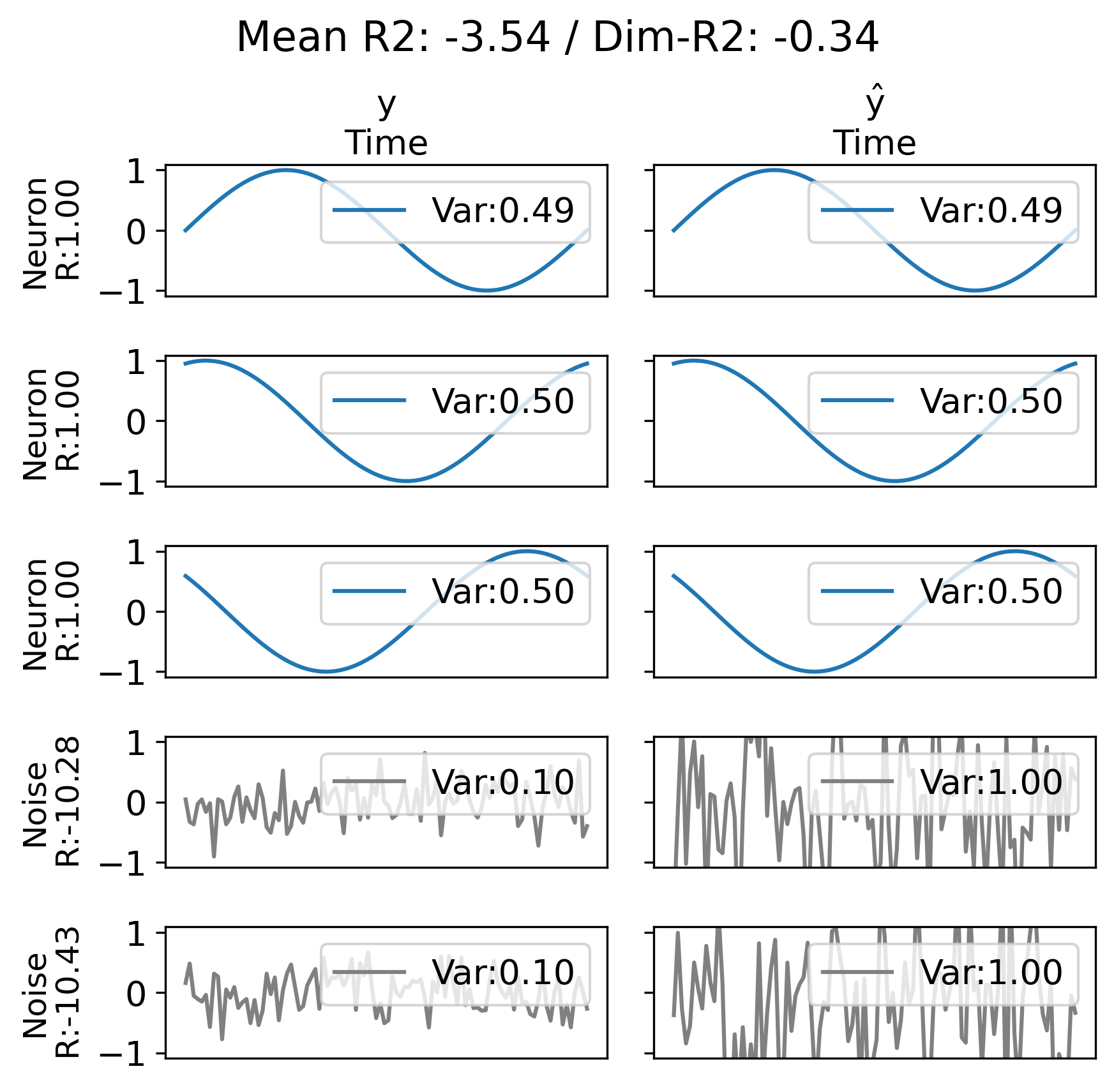}}
\\
\subfloat[]{\includegraphics[width=0.33\textwidth]{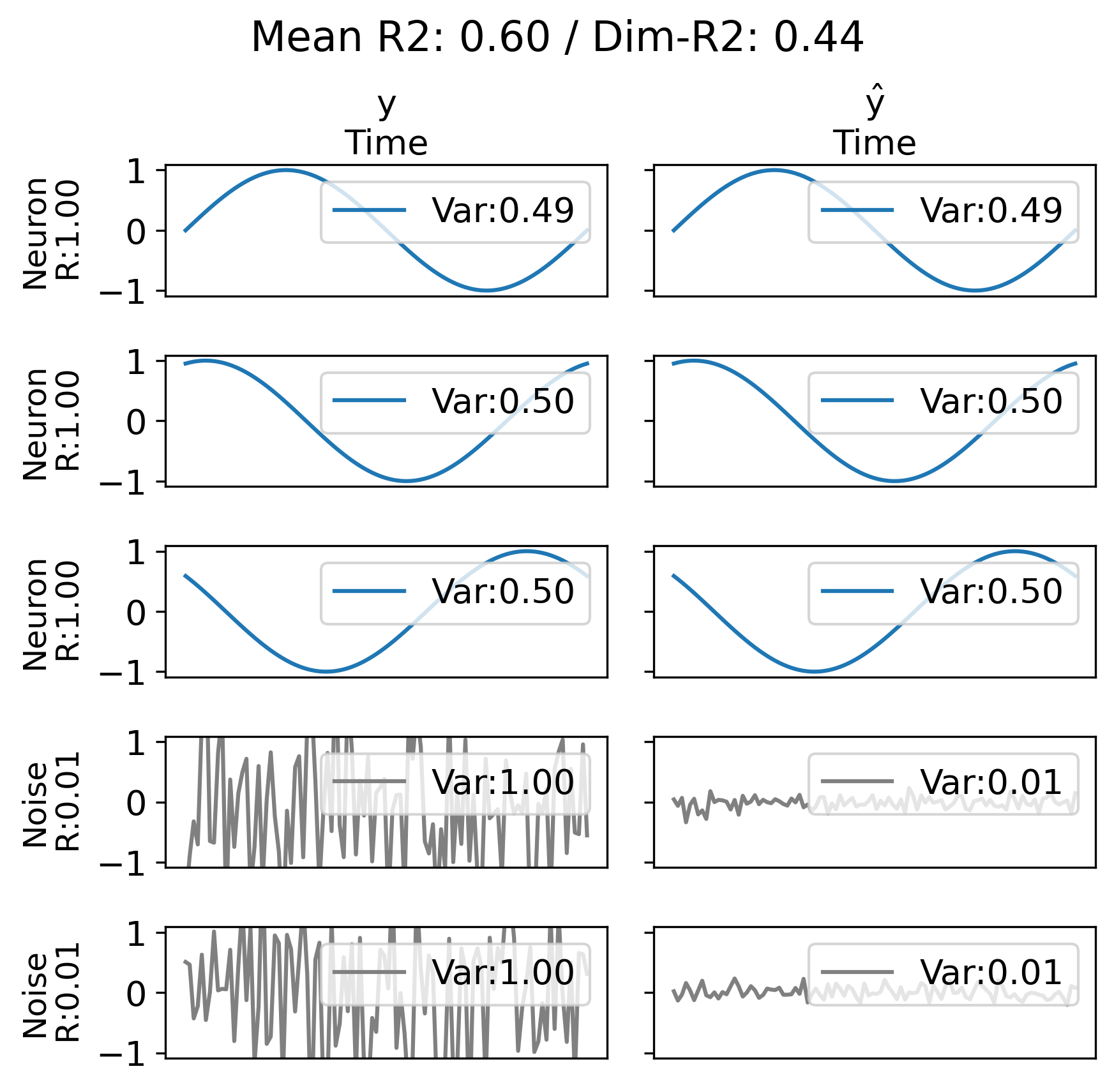}}
\subfloat[]{\includegraphics[width=0.33\textwidth]{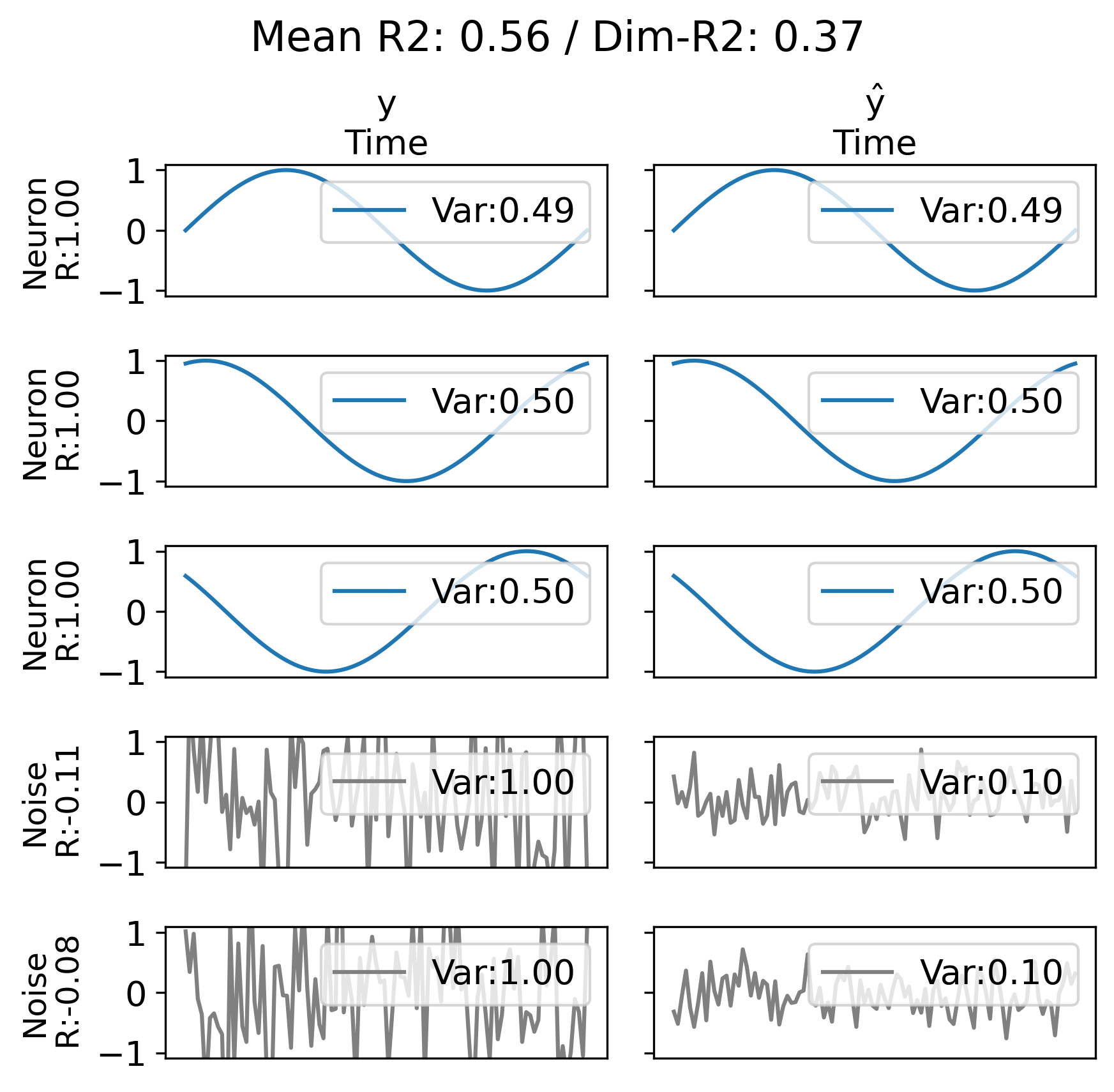}}
\subfloat[]{\includegraphics[width=0.33\textwidth]{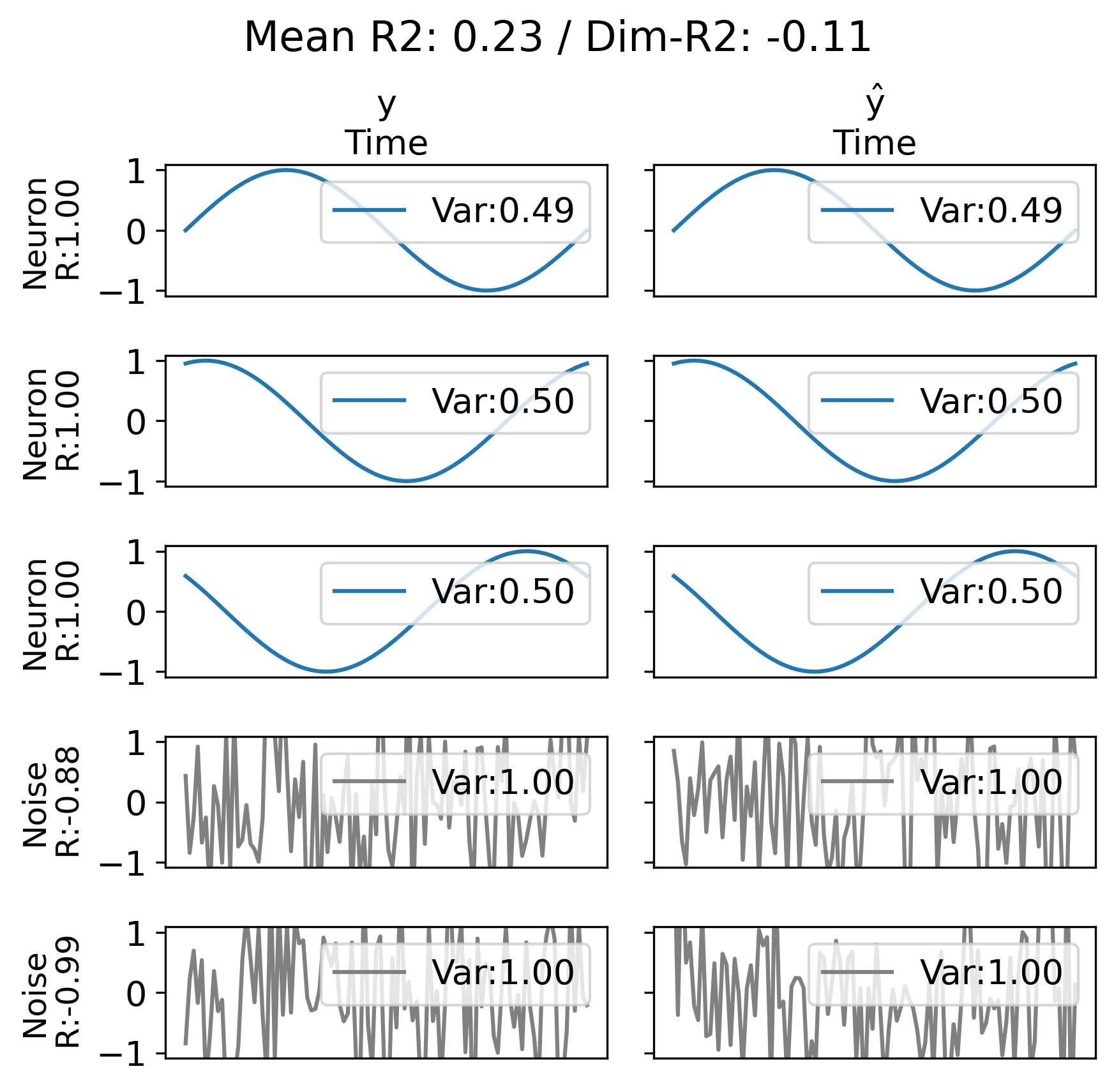}}

\caption{Synthetic sinusoidal datasets with noise channels and corresponding mean R2 and Dim-R2 scores, across noise variance combinations. Each panel shows 5-channel sine waves where 2 channels are replaced with Gaussian noise. Rows share $y$ noise channel variance; columns share $\hat{y}$ noise channel variance.}
\label{fig:r2-sample-full}
\end{figure}

\FloatBarrier

\subsection{Data-constrained recurrent neural network predictions on mouse motor cortex neuropixel recordings}
\label{appendix:dc-rnn}

\begin{figure}[h]
    \centering
    \includegraphics[width=\textwidth]{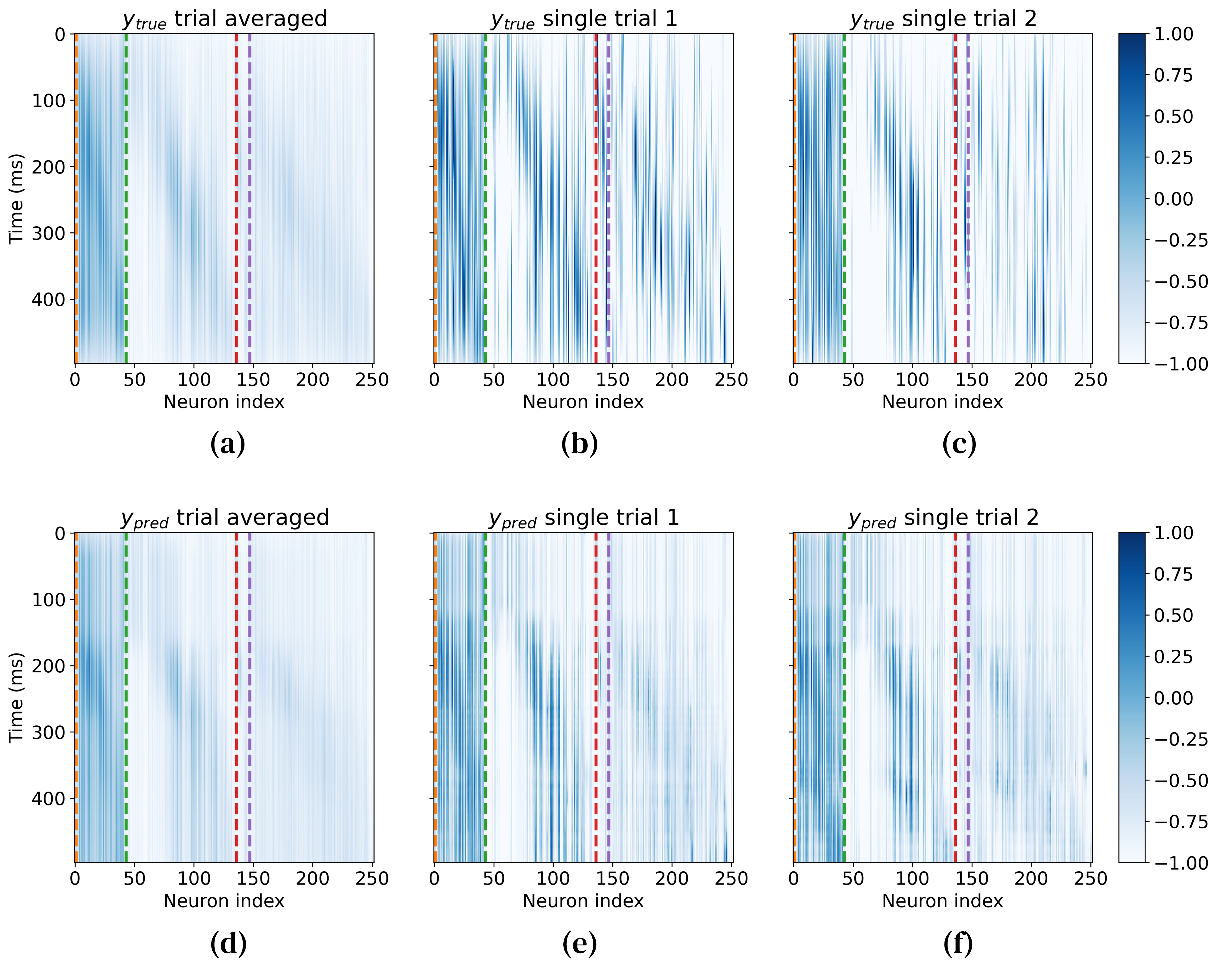}
    \caption{Examples of $y$ and $\hat{y}$ from DC-RNN trained to reproduce neural activity. This single session contains 78 trials, with 42, 93, 11, and 106 neurons from DCN, M1, Striatum, and Thalamus, respectively. This session corresponds to Session index 21 with a 50ms Gaussian filter in Fig. \ref{fig:noise_resilience_dcrnn}. Dashed lines separate neurons by brain region (from left): DCN (orange), M1 (green), Striatum (red), Thalamus (purple). (a) \& (d) Trial averaged activity, (b) \& (e) Single trial example 1, (c) \& (f) Single trial example 2. (a)-(c) $y$, (d)-(f) $\hat{y}$.}
    \label{fig:sample-rnn_data}
\end{figure}

\subsubsection{Data}
A total of 22 sessions were collected from four mice performing a reach-to-grab task \citep{sauerbrei2020cortical, 10.7554/eLife.65906, LEVY2020954}. Neural activity (spiketrains) was recorded simultaneously from the deep cerebellar nuclei (DCN), primary motor cortex (M1), Striatum, Thalamus using Neuropixels probes \citep{doi:10.1126/science.abf4588}. Spiketrains were recorded at 500Hz (2ms bin) where each value indicates the presence (1) or absence (0) of a spike. For each brain region, this results in a binary array of shape (time, neurons).

Hand kinematics were recorded using a camera at sampling rate of 500Hz, synchronized with the Neuropixels. The 3D hand coordinates (Units: mm) were extracted using Animal Part Tracker \citep{APT}. Movement onset was defined as the time point when the hand position exited a predefined square region representing the resting state.

\subsubsection{Preprocessing}
Preprocessing of spiketrains involved 5 steps: slicing it to the region of interest (100ms before to 400ms after movement onset), Gaussian filtering ($\sigma$=5, 10, 25, 50ms), reordering neuron indices based on peak activity for interpretability (does not affect training), normalizing activity range to match the range of DC-RNN activation function, and stacking the neurons across brain regions. The train set metadata was used to process the validation and test set for neuron reordering and activity normalization stages.

\subsubsection{Experiment setup}
The DC-RNNs were trained with 5 by 3 fold nested k-fold cross validation with 3 different random seeds, resulting in 5 $\times$ 3 $\times$ 3 = 45 experiments per condition. Each train, validation, test split were stratified-split with respect to the number of reach success trials, to balance the data characteristics. The random seeds affected model weight initialization and cross validation splits. Including 4 Gaussian filter sizes, 22 sessions, and 45 experiments per condition, there were $4\times22\times45=3960$ DC-RNN experiments in total. All DC-RNNs were trained on a SLURM \citep{10.1007/10968987_3} cluster using CPU nodes with 1 CPU and 4GB memory.

The DC-RNN was a vanilla RNN where the number of hidden neurons matched the number of recorded neurons in the spiketrain data. The activation function was tanh. The Adam optimizer \citep{kingma2014adam} was used to train the DC-RNN \citep{werbos1990backpropagation} with learning rate of 1e-3 and batch size of 32. Early stopping stopped training by measuring Dim-R2 every 100 updates with patience value of 100. 

DC-RNN weights were initialized using He initialization, a uniform distribution $\mathcal{U}(-\sqrt{\frac{1}{hidden\_size}},\sqrt{\frac{1}{hidden\_size}}$, which is the default in PyTorch \citep{paszke2017automatic}. The DC-RNN was implemented with PyTorch \citep{paszke2017automatic}. The learning rate and batch size were selected after initial parameter sweep with learning rate of 1e-2, 1e-3, and batch size of 16, 32.

For evaluating the dimensional view of Dim-R2, the experiment condition is the basis of evaluation. Thus, $y$ and $\hat{y}$ were aggregated \citep{YOO2025110239} across random seeds, validation folds, and test folds under the same experiment conditions, yielding shape (Random seed, Cross validation fold, Aggregated test trial, Time, Neuron). For evaluating noise resilience, Dim-R2 and mean R2 were measured per experiment on $y$ and $\hat{y}$ of shape (Trial, Time, Neuron). For mean R2, $y$ and $\hat{y}$ were reshaped to (Trial$\times$Time, Neuron) since mean R2 is undefined for 3D data.

\subsection{Image reconstruction using Variational Autoencoders}
\label{appendix:vae}

\begin{table}[t]
\centering
\footnotesize
\begin{tabular}{ll}
\toprule
\textbf{Module} & 
\begin{tabular}[c]{l} 
\textbf{Architecture Details}
\end{tabular}\\
\midrule

\textbf{ConvBlock} &
\begin{tabular}[c]{l}
Conv2d(in\_channels, out\_channels, kernel\_size, stride, padding)\\
ReLU\\
BatchNorm2d(out\_channels) \\
\end{tabular} \\
\midrule

\textbf{CNN Encoder} &
\begin{tabular}[c]{l}
Input: (Channels, H, W) \\
ConvBlock(Channels, $n_f$, k5, s2, p2) \\
ConvBlock($n_f$, $2n_f$, k5, s2, p2) \\
ConvBlock($2n_f$, $4n_f$, k3, s2, p1) \\
ConvBlock($4n_f$, $6n_f$, k3, s2, p1) \\
Flatten \\
Linear($\cdot$, 512)\\
ReLU \\
Linear(512, $2 \times n_{z}$) \\
Outputs: $\mu_z$, $\log \sigma_z$ \\
\end{tabular} \\

\midrule

\textbf{CNN Decoder} (MNIST) &
\begin{tabular}[c]{l}
Linear($n_{z}$, $7 \times 7 \times 6n_f$)\\
ReLU \\
Unflatten to $(6n_f, 7, 7)$ \\
ConvTranspose(6$n_f$ $\rightarrow$ 4$n_f$, k3, s2, p1, op1), ReLU \\
ConvTranspose(4$n_f$ $\rightarrow$ out\_ch, k5, s2, p2, op1) \\
Output: (Channels, 28, 28) \\
\end{tabular} \\

\midrule

\textbf{CNN Decoder} (CelebA) &
\begin{tabular}[c]{l}
Linear($n_{z}$, $4 \times 4 \times 6n_f$)\\
ReLU \\
Unflatten to $(6n_f, 4, 4)$ \\
ConvTranspose(6$n_f$ $\rightarrow$ 4$n_f$, k3, s2, p1, op1), ReLU \\
ConvTranspose(4$n_f$ $\rightarrow$ 2$n_f$, k3, s2, p1, op1), ReLU \\
ConvTranspose(2$n_f$ $\rightarrow$ 2$n_f$, k5, s2, p2, op1), ReLU \\
ConvTranspose(2$n_f$ $\rightarrow$ $n_f$, k5, s2, p2, op1), ReLU \\
ConvTranspose($n_f$ $\rightarrow$ Channels, k5, s2, p2, op1) \\
Output: (Channels, 128, 128) \\
\end{tabular} \\

\bottomrule
\end{tabular}

\caption{VAE encoder and decoder architectures used for MNIST and CelebA. $n_f$ = Number of filters, $n_{z}$ = latent dimension size, k = kernel size, s = stride, p = padding, op = output padding. }
\label{tab:vae_architectures}
\end{table}

\subsubsection{Data}
\textbf{MNIST.}  
The MNIST handwritten digits dataset consists of grayscale images with shape (Channels, Width, Height) = (1, 28, 28). The pixel values were normalized to [0,1]. The training and test sets contain 60,000 and 10,000 images, respectively. Only the test set was used for evaluating Dim-R2. Although MNIST includes 10 digit classes, these labels were not used for training and were used only for class-wise visualization of Dim-R2.

\textbf{CelebA.}  
The CelebA face dataset contains 3-channel RGB images originally of size (178, 218), which were resized to (128, 128) for our experiments, yielding images of shape (Channels, Width, Height) = (3, 128, 128). The pixel values were normalized to [0,1]. The training and test sets contain 162,770 and 19,962 images, respectively, and the test set was used for evaluating Dim-R2. Each image also includes 40 binary attributes in one-hot format. For visualization, we selected the eight attributes that had the most balanced True/False counts: \texttt{Smiling}, \texttt{Attractive}, \texttt{Mouth\_Slightly\_Open}, \texttt{High\_Cheekbones}, \texttt{Wearing\_Lipstick}, \texttt{Heavy\_Makeup}, \texttt{Male}, and \texttt{Wavy\_Hair}. These attributes were not used during VAE training but were used for attribute-wise visualization of Dim-R2.

\subsubsection{Experimental Setup}
The VAEs \citep{NIPS2016_eb86d510, doersch2016tutorial, PinheiroCinelli2021} were trained in a single instance with the provided train and test datasets. Both MNIST and CelebA VAEs were trained on a single desktop with NVIDIA T1000 8GB GPU. The VAE was a Convolutional neural network with the following architecture described in Table \ref{tab:vae_architectures}, with number of filters $n_f=10$ and a latent dimension size $n_z=100$. The loss function was the following equation:

\begin{equation}
    \mathcal{L}_{VAE}=\lambda_{KL}\mathcal{L}_{KL}+\mathcal{L}_{recon}
\end{equation}

\begin{equation}
    \mathcal{L}_{KL}= \frac{1}{2}(\sigma_z +\mu_z - 1 - \log \sigma_z )
\end{equation}

where $\mathcal{L}_{recon}$ was L1 loss for MNIST and Binary Cross Entropy with logits loss for CelebA. The $\lambda_{KL}$ was set to 5e-4. The reparametrization trick was only used during training, and only $\mu_z$ was used during evaluation.

The Adam optimizer \citep{kingma2014adam} was used with learning rate of 1e-3 and batch size of 64. Early stopping stopped training by measuring mean absolute error every 40 updates with patience value of 50. The final $y$ and $\hat{y}$ shapes were (Data, Channels, Width, Height).


\section{Additional dimensional view figures}

\begin{figure}[t]
\centering
\includegraphics[width=\textwidth]{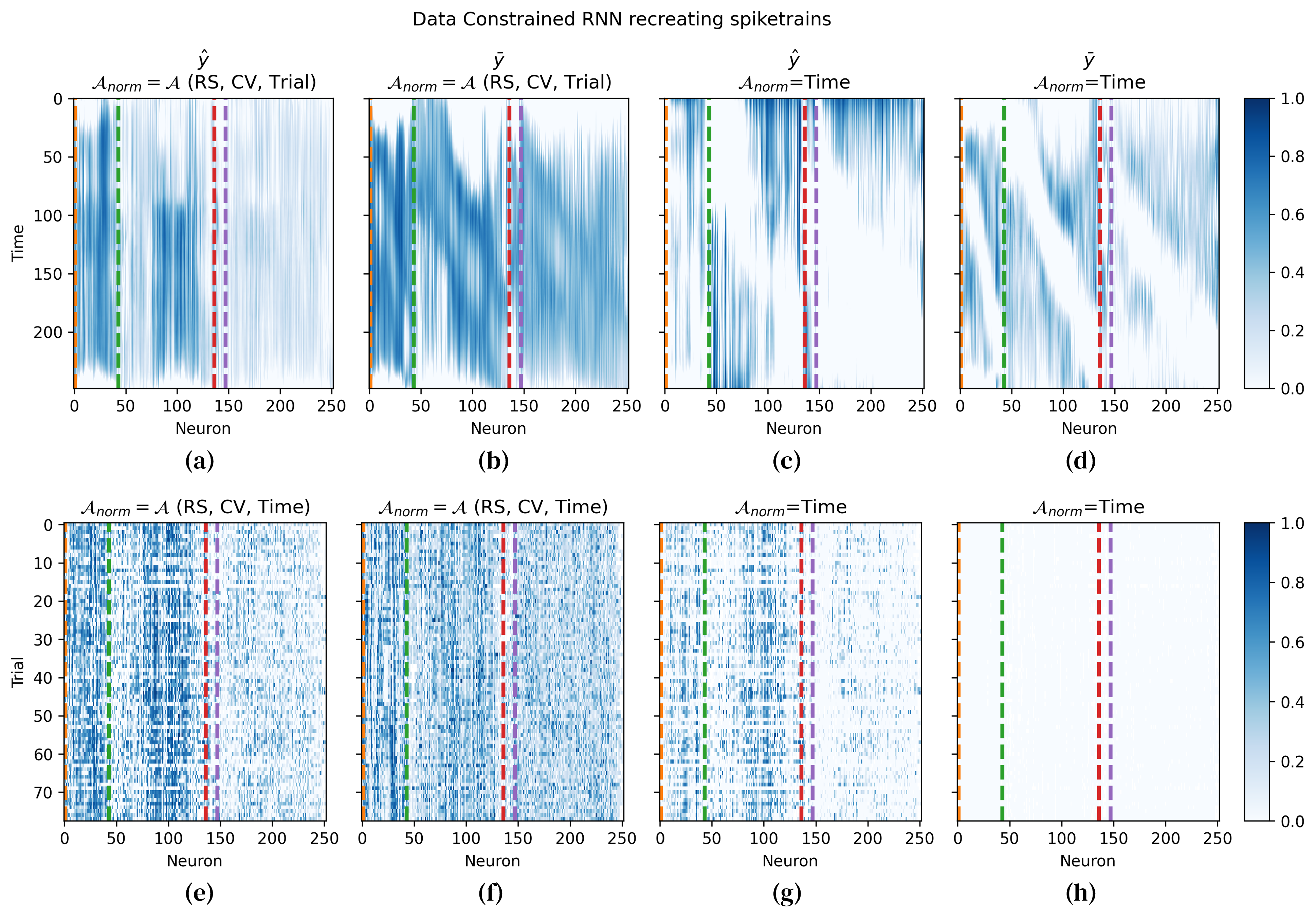}
\caption{Dimensional view of Dim-R2 on DC-RNN neural data. Each heatmap shows Dim-R2 measured with $\hat{y}$ or $\bar{y}$ against $y$ under different $\mathcal{A}$ and $\mathcal{A}_{norm}$ settings. Dashed lines separate brain regions (left to right): DCN (orange), M1 (green), Striatum (red), Thalamus (purple). (a)-(d) $\mathcal{A}$=(Random seed (RS), Cross validation fold (CV), Trial). (e)-(h) $\mathcal{A}$=(RS, CV, Time).}
\label{fig:dimensional-dc-rnn-full}
\vspace{-15pt}
\end{figure}

Fig. \ref{fig:dimensional-dc-rnn-full}a,b,g,h are discussed in the main text (Section \ref{results:dimensional-view}). Setting $\mathcal{A}_{norm}$=Time with $\mathcal{A}$=(RS,CV,Trial) evaluates accuracy relative to time variability, across Time and Neuron dimensions. Some DCN and late-onset M1 neurons are well predicted, appearing as vertical bands of high scores (Fig. \ref{fig:dimensional-dc-rnn-full}c). Fig. \ref{fig:dimensional-dc-rnn-full}d reveals the underlying spiketrain structure, as time-averaged $\bar{y}$ reflects temporal patterns rather than yielding R2=0. Setting $\mathcal{A}$=(RS,CV,Time) with $\mathcal{A}_{norm}$=$\mathcal{A}$ shows neurons with high trial variability as vertical bands of high scores, with $\hat{y}$ capturing trial variability better than $\bar{y}$ (Fig. \ref{fig:dimensional-dc-rnn-full}e,f).

\FloatBarrier

\section{Additional noise resilience figures}

\begin{figure}[h]
    \centering
    \subfloat[]{\includegraphics[width=\textwidth]{figures/r2_comparison_0.01.png}}
    \\
    \subfloat[]{\includegraphics[width=\textwidth]{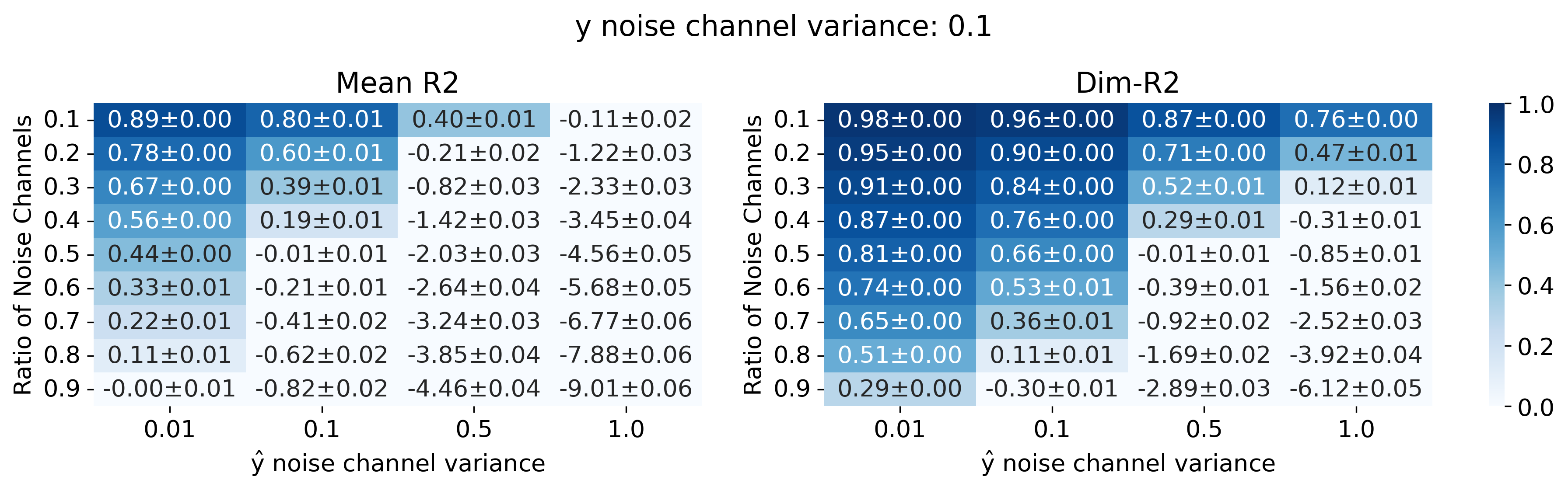}}
    \\
    \subfloat[]{\includegraphics[width=\textwidth]{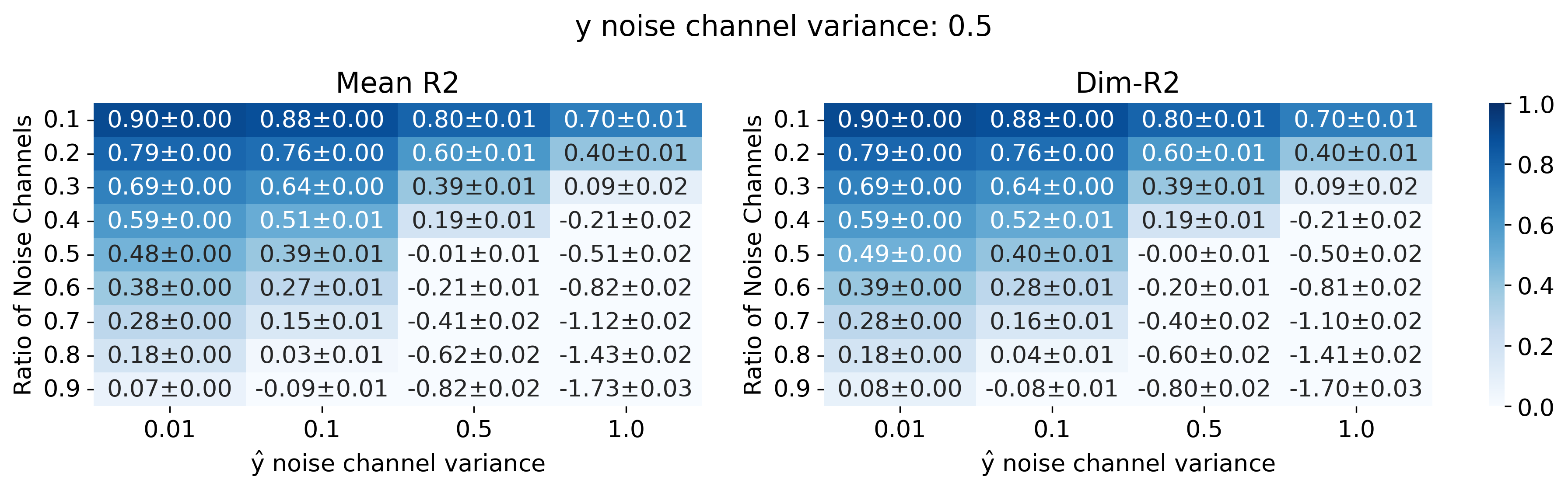}}
    \\
    \subfloat[]{\includegraphics[width=\textwidth]{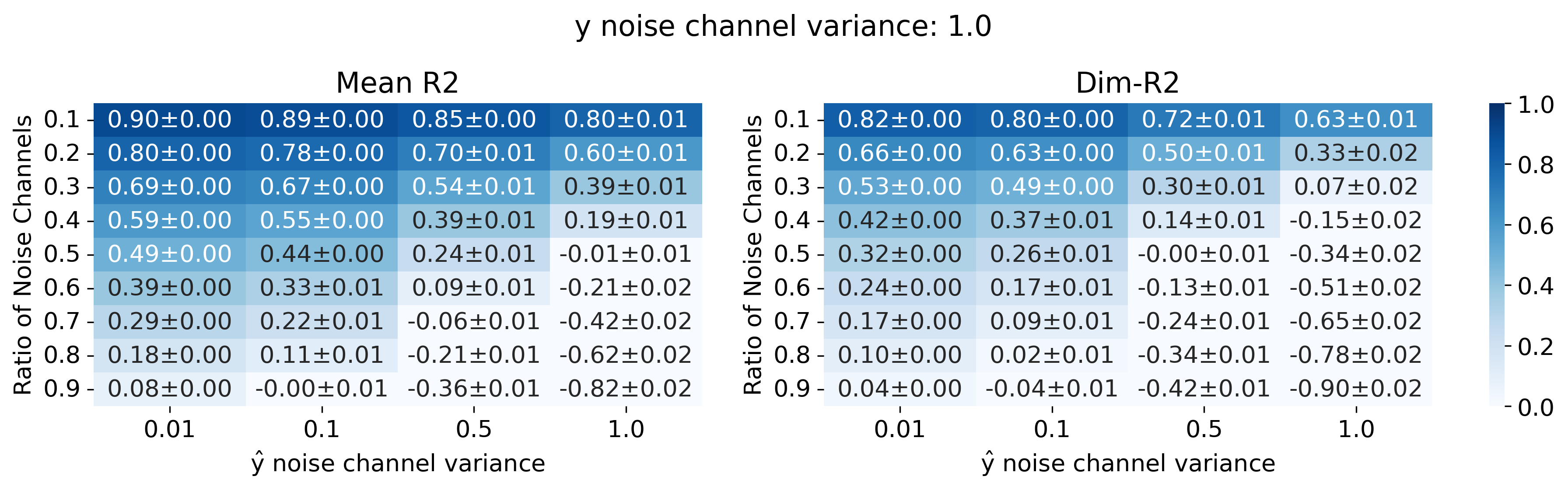}}
    \caption{Dim-R2 highlights the presence of channels with high predictive accuracy in the presence of noisy channels. Scores were measured on simulated sinusoidal data  (Fig. \ref{fig:r2-sample}) across hyperparameter sweeps. Note in (c) that mean R2 and Dim-R2 show similar scores when $y$ noise channel variance equals the signal variance of 0.5 (Fig. \ref{fig:r2-sample-full}). Each entry shows the mean$\pm$standard deviation across 100 random repetitions.}
    \label{fig:noise_resilience_full}
\end{figure}

\begin{figure}[h]
    \vspace{-20pt}
    \centering
    \subfloat[]{\includegraphics[width=0.5\textwidth]{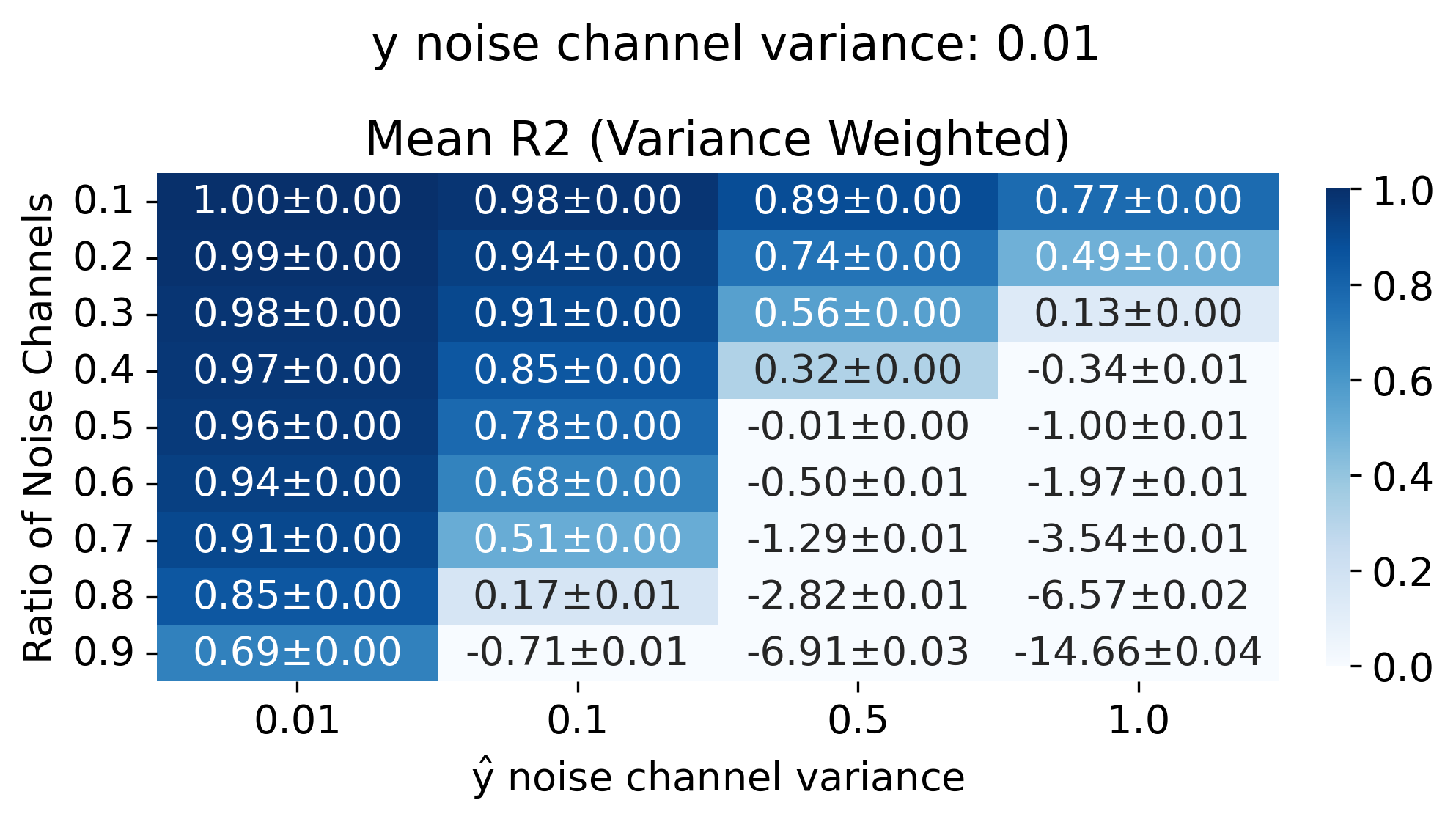}}
    \subfloat[]{\includegraphics[width=0.5\textwidth]{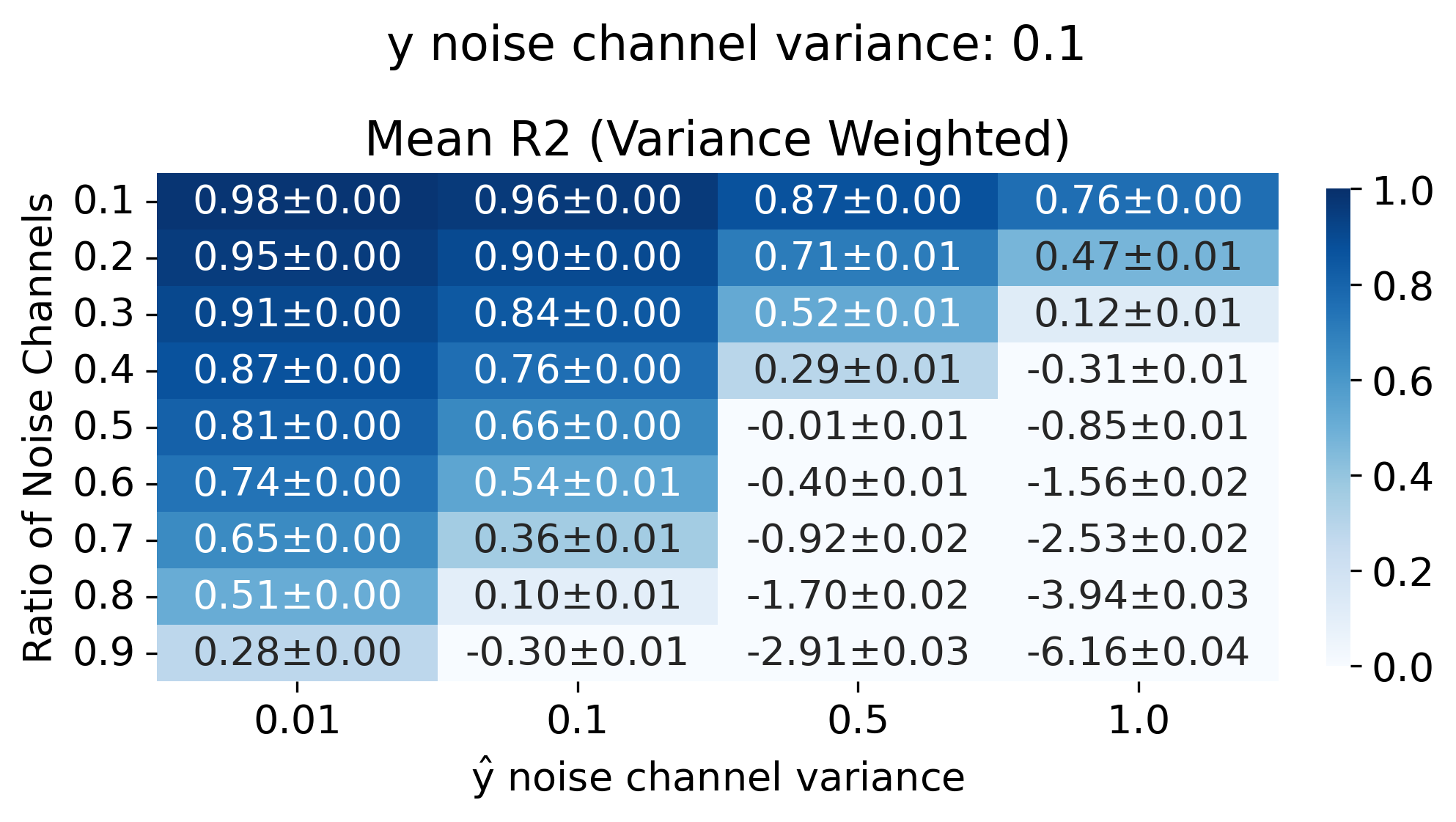}}
    \\
    \subfloat[]{\includegraphics[width=0.5\textwidth]{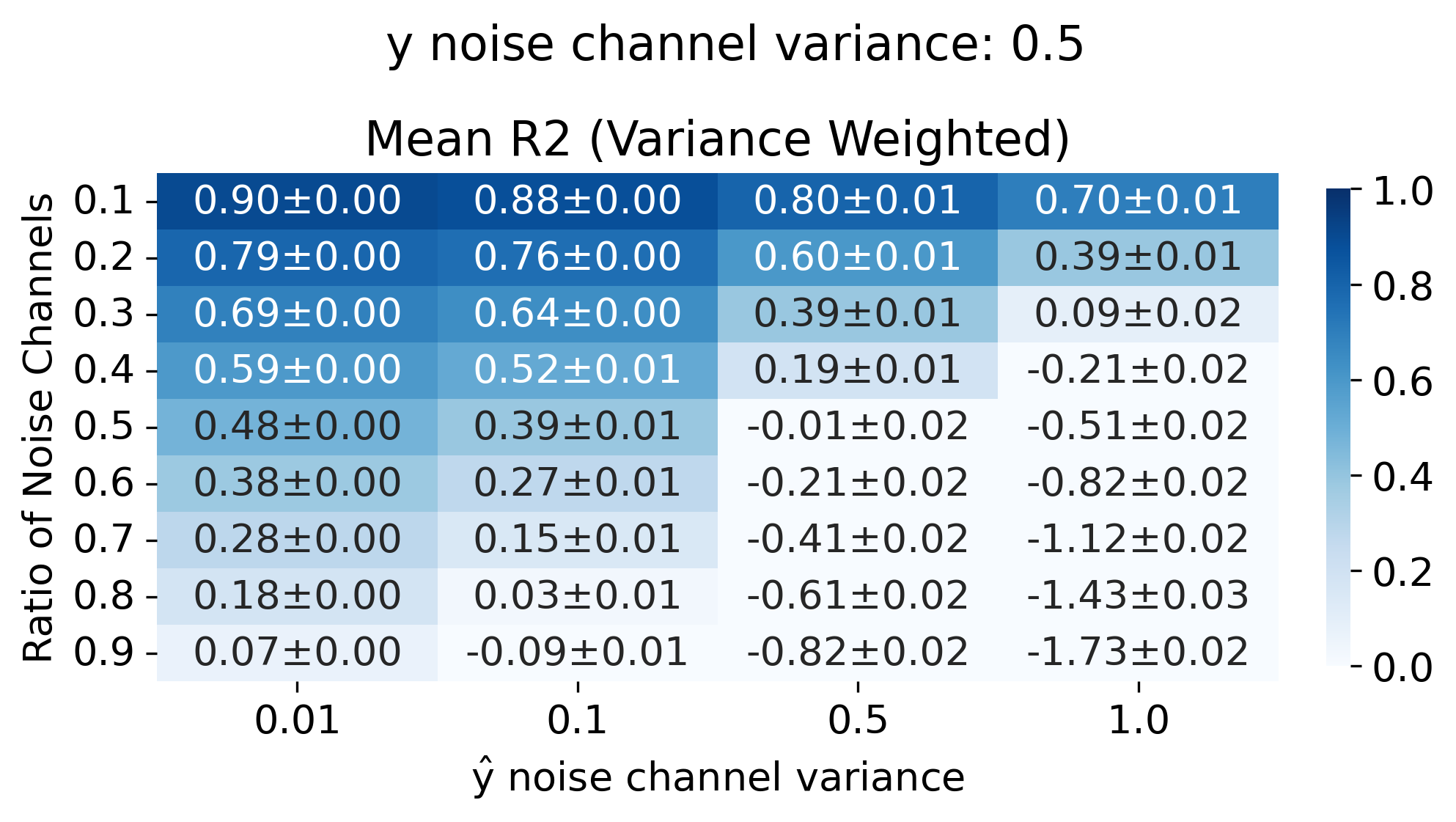}}
    \subfloat[]{\includegraphics[width=0.5\textwidth]{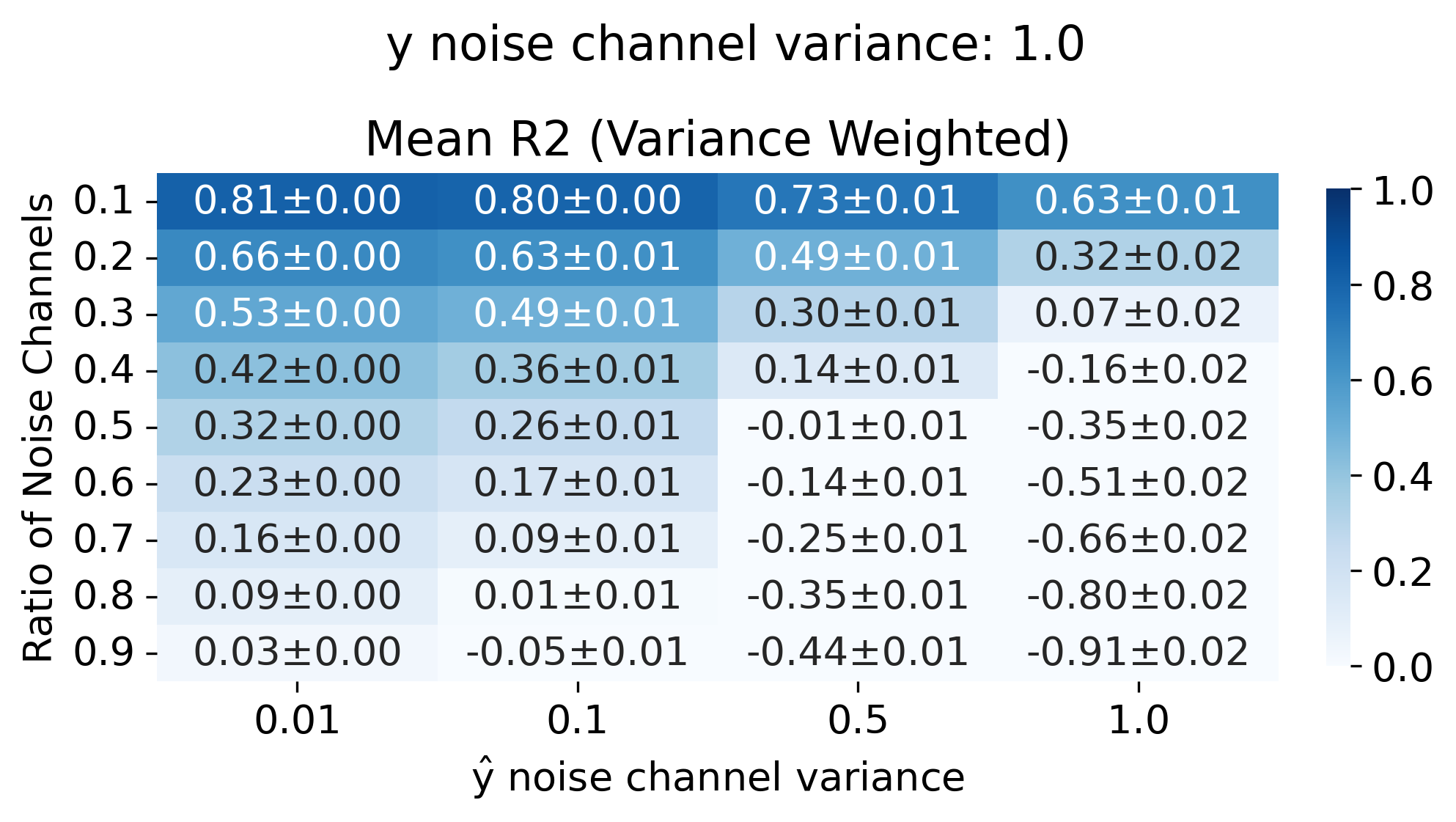}}
    \caption{Variance weighted mean R2 scores measured on simulated sinusoidal data  (Fig. \ref{fig:r2-sample}) across hyperparameter sweeps. They match Dim-R2 scores (Fig. \ref{fig:noise_resilience_full}) because there were no mean shifts in $y,\hat{y}$ across the channel dimension. Each entry shows the mean$\pm$standard deviation across 100 random repetitions.}
    \label{fig:resilience_var_r2}
\end{figure}

\begin{figure}[h]
    \centering
    \subfloat[]{\includegraphics[width=0.5\textwidth]{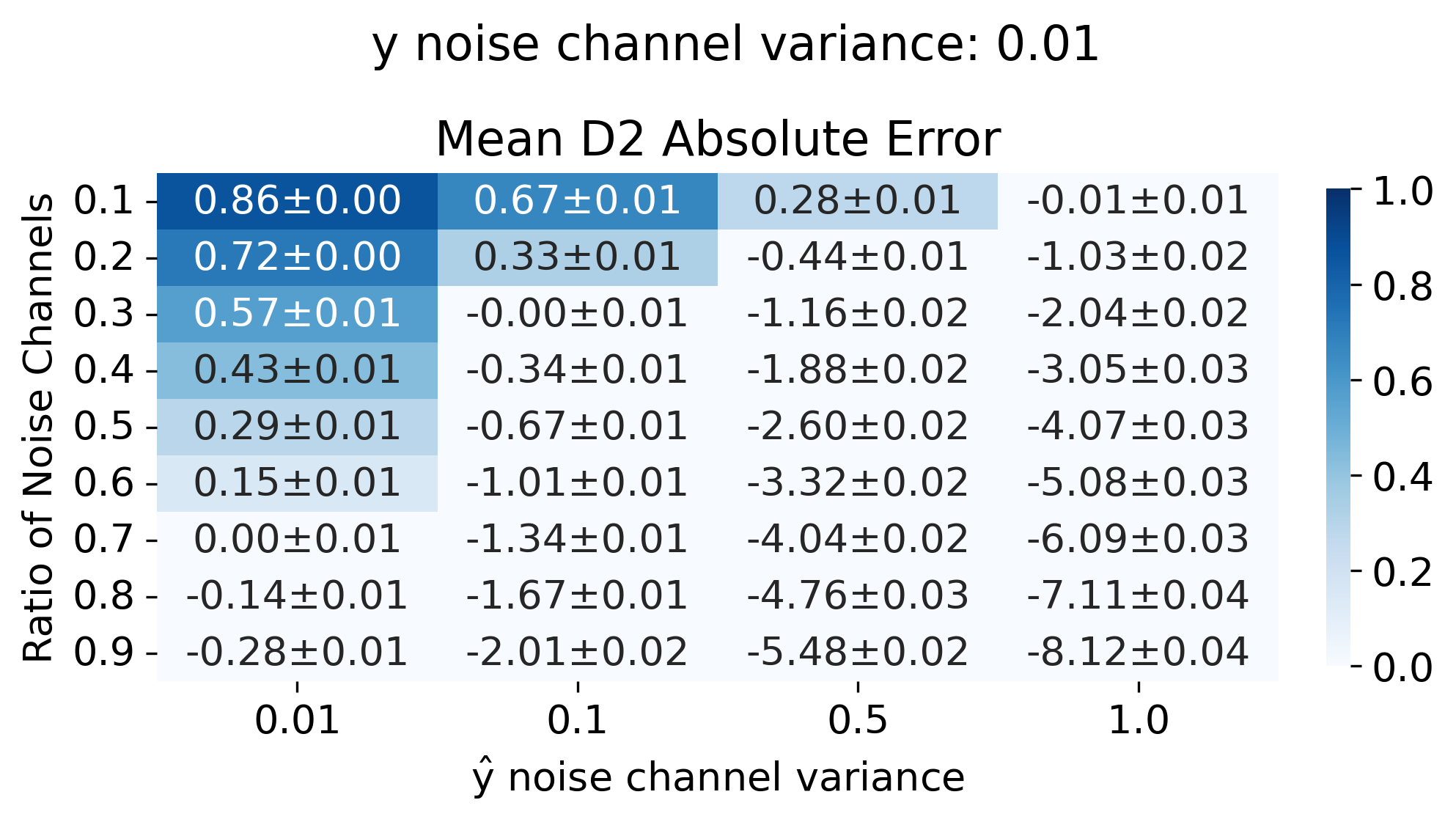}}
    \subfloat[]{\includegraphics[width=0.5\textwidth]{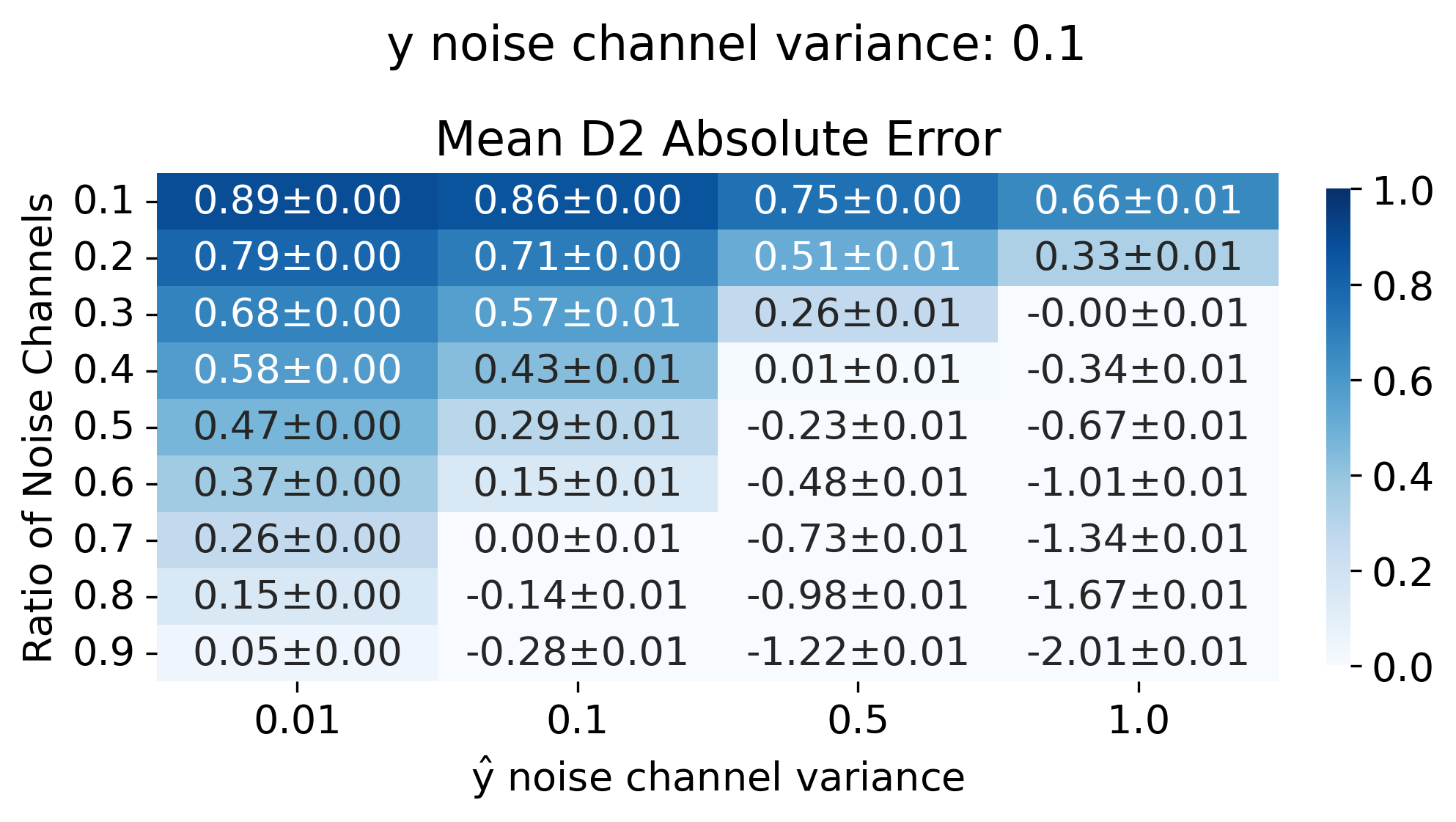}}
    \\
    \subfloat[]{\includegraphics[width=0.5\textwidth]{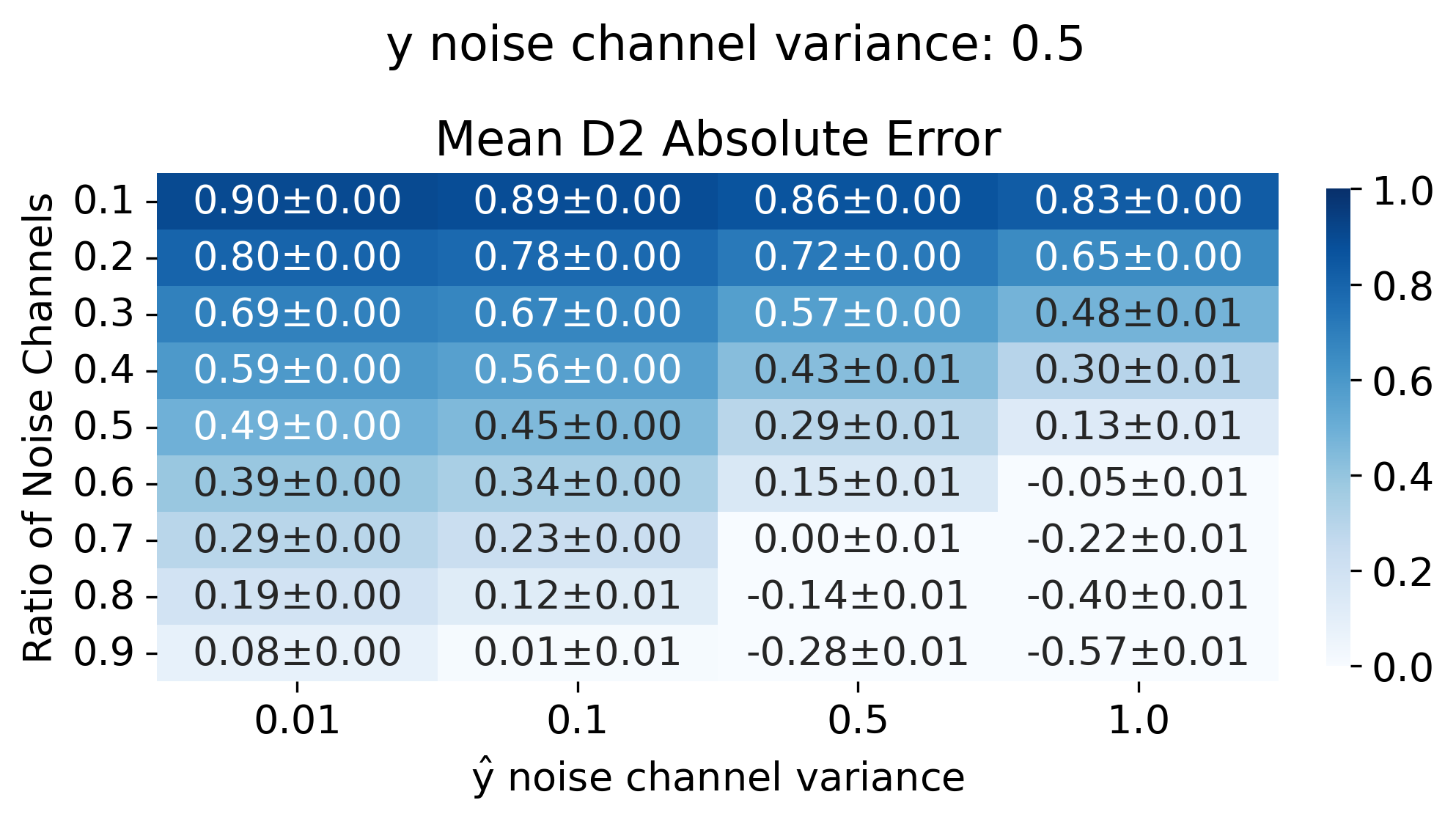}}
    \subfloat[]{\includegraphics[width=0.5\textwidth]{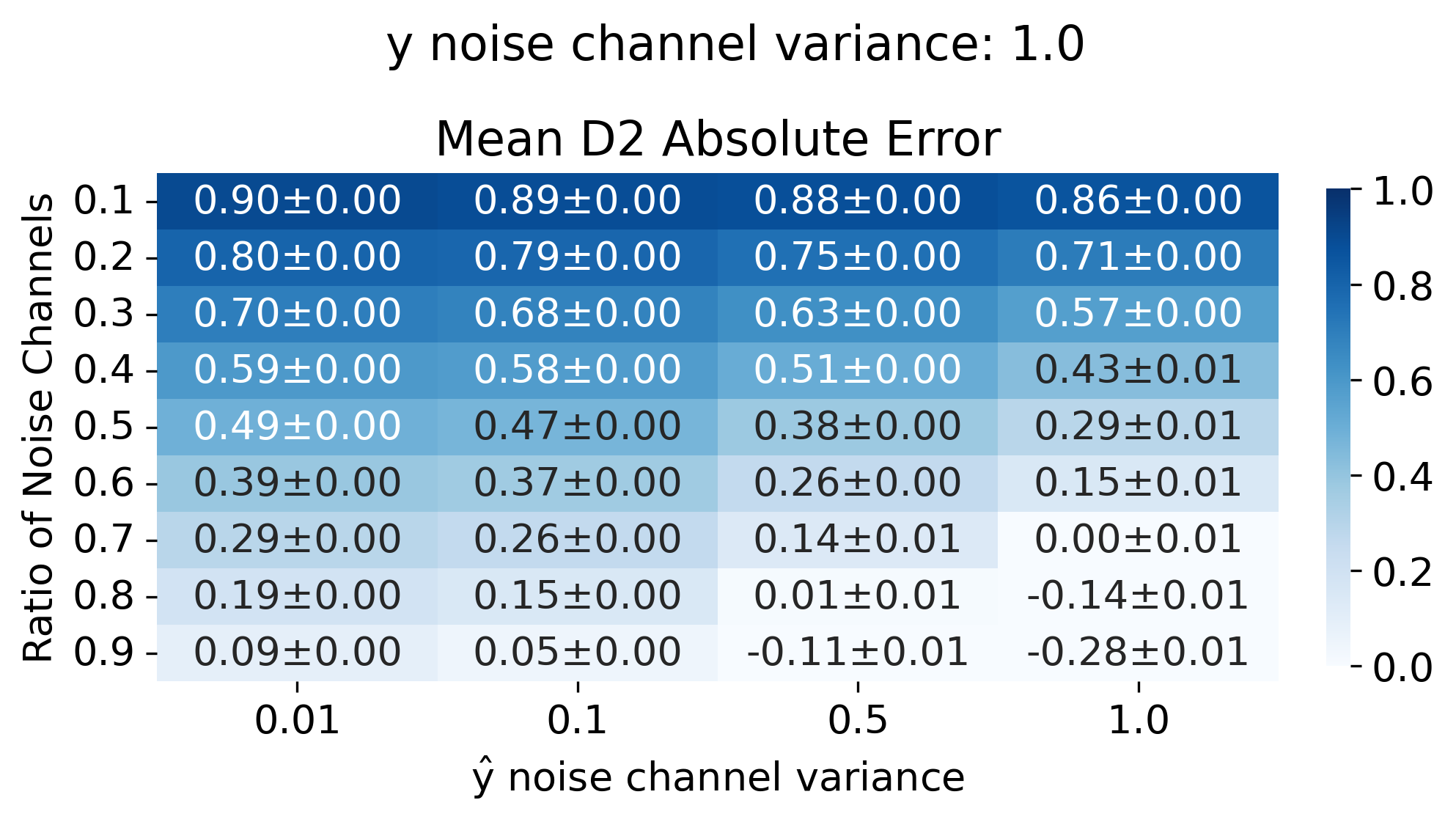}}
    \caption{Mean D2 absolute error scores measured on simulated sinusoidal data  (Fig. \ref{fig:r2-sample}) across hyperparameter sweeps. They exhibit the same noise-sensitivity issues as mean-R2, whereas Dim-R2 remains stable under these conditions (Fig. \ref{fig:noise_resilience_full}). Each entry shows the mean$\pm$standard deviation across 100 random repetitions.}
    \label{fig:resilience_d2_ae}
\end{figure}

\begin{figure}[h]
    \vspace{-30pt}
    \centering
    \subfloat[]{\includegraphics[width=0.5\textwidth]{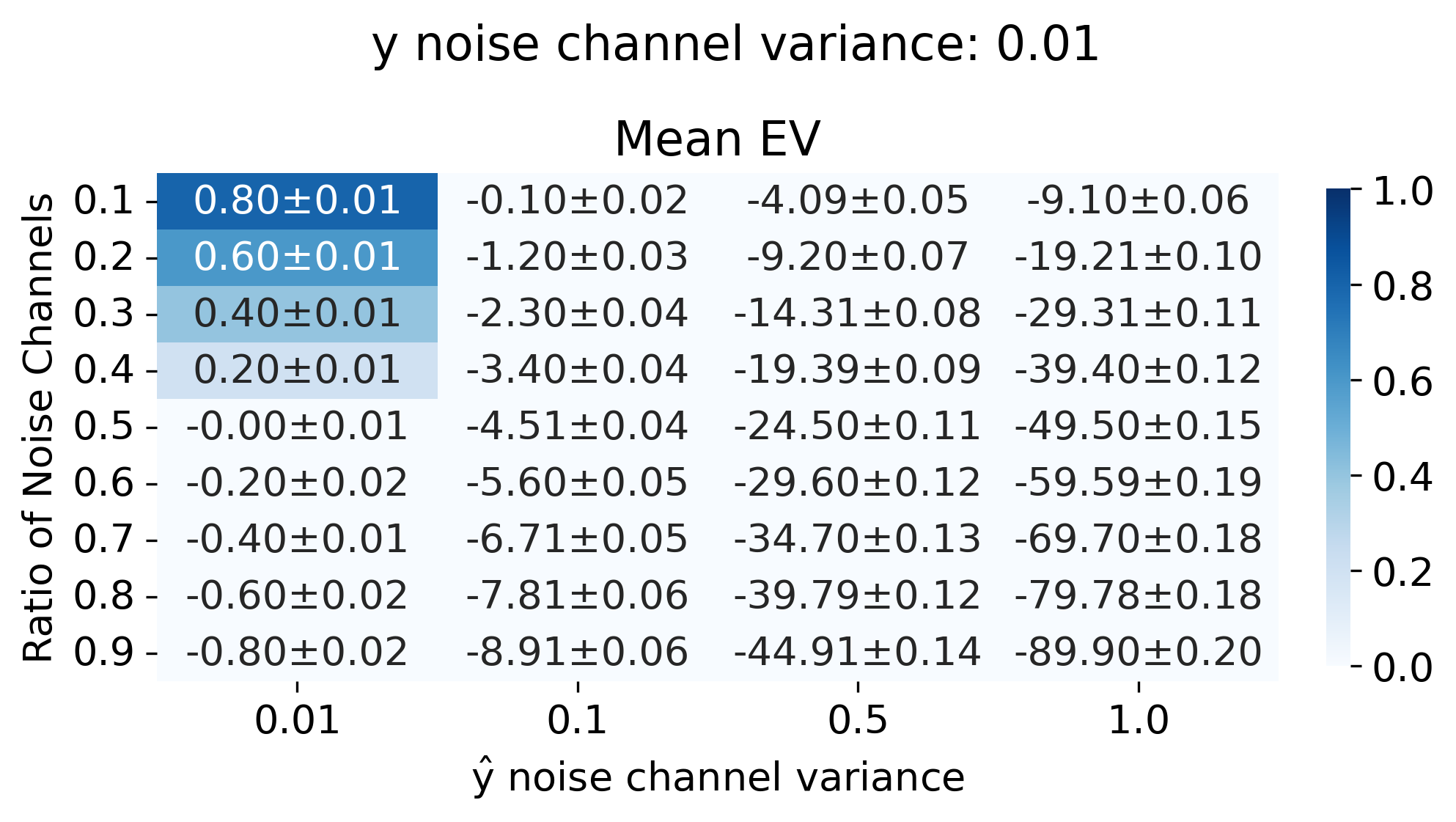}}
    \subfloat[]{\includegraphics[width=0.5\textwidth]{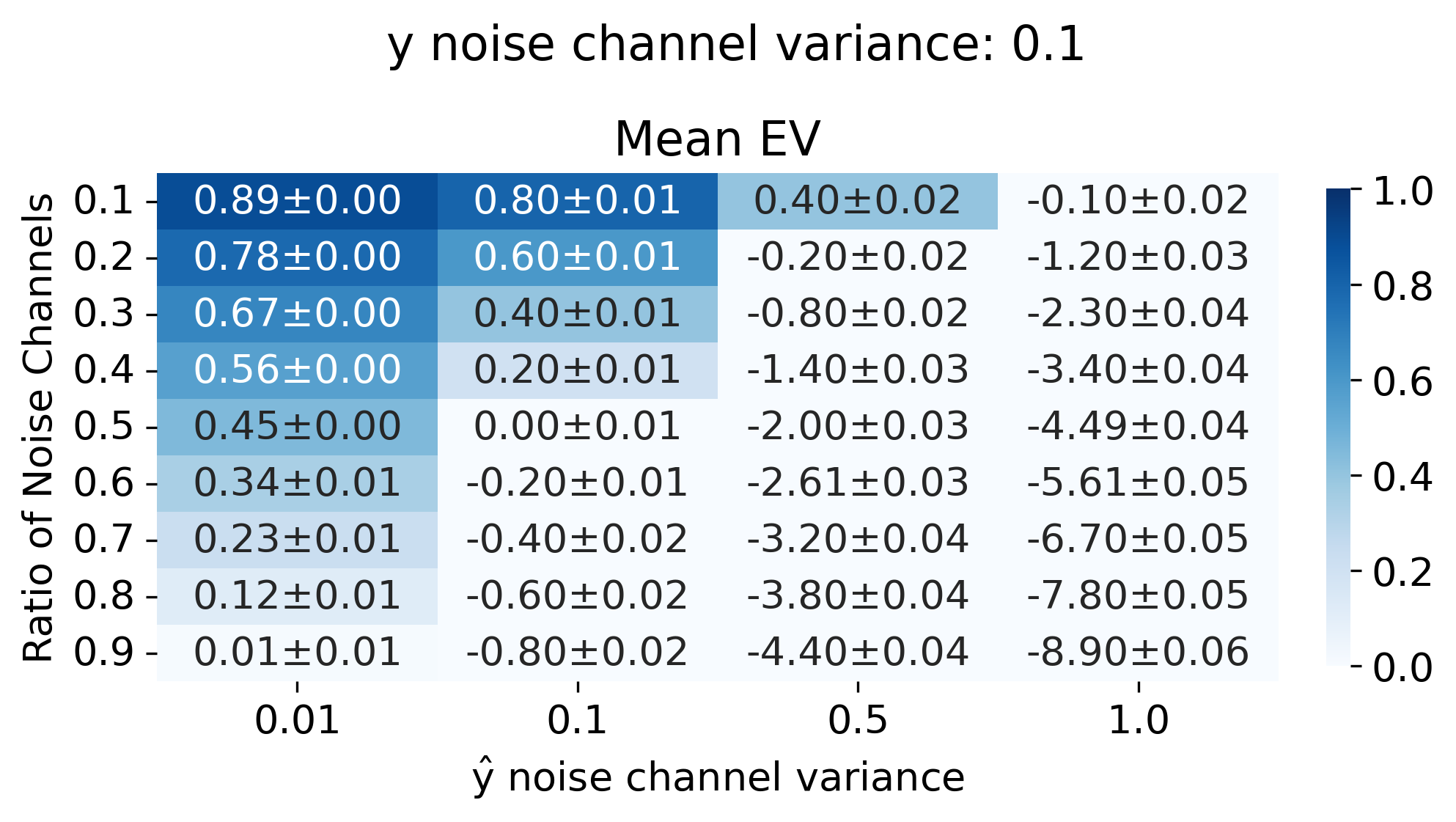}}
    \\
    \subfloat[]{\includegraphics[width=0.5\textwidth]{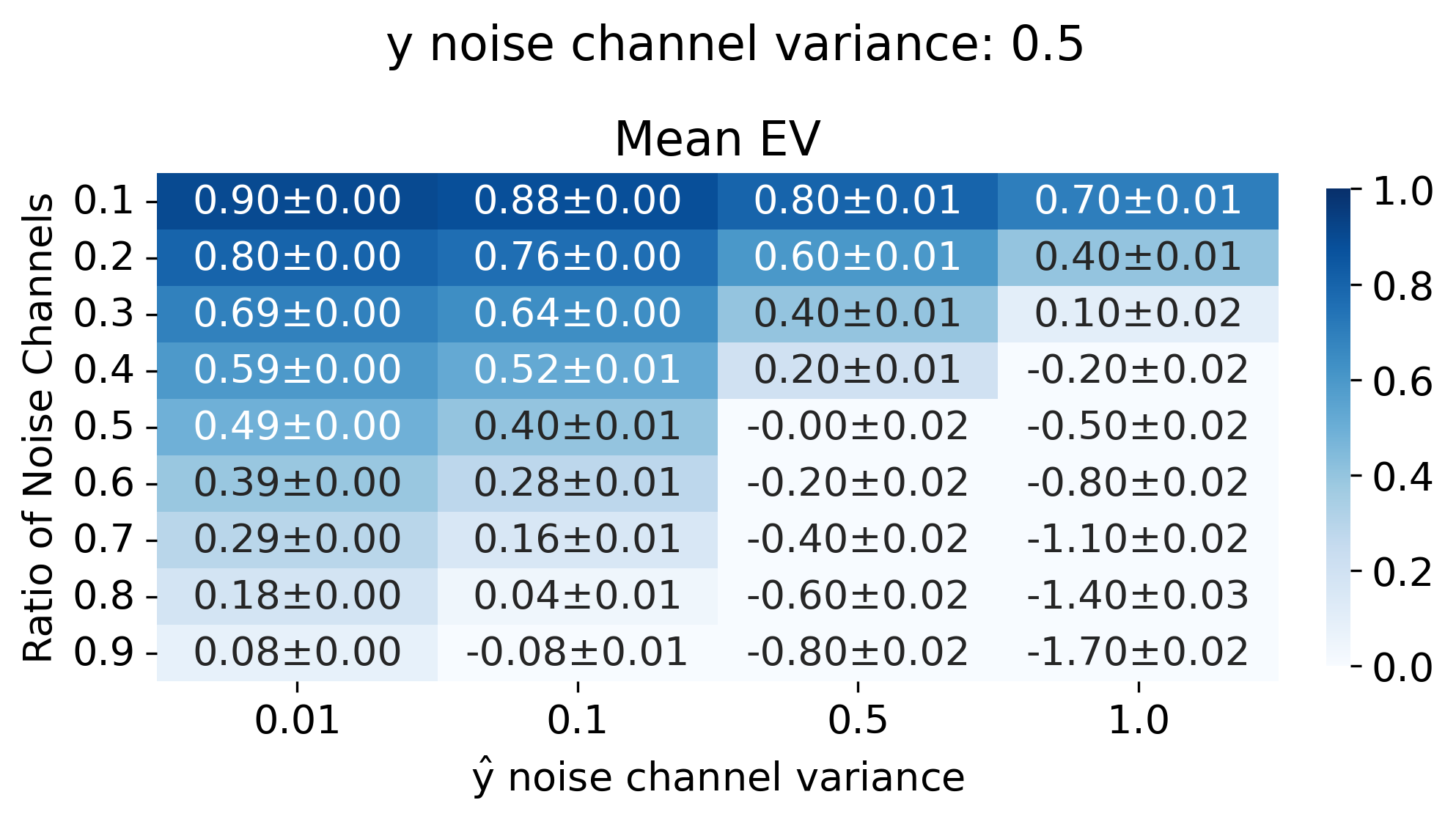}}
    \subfloat[]{\includegraphics[width=0.5\textwidth]{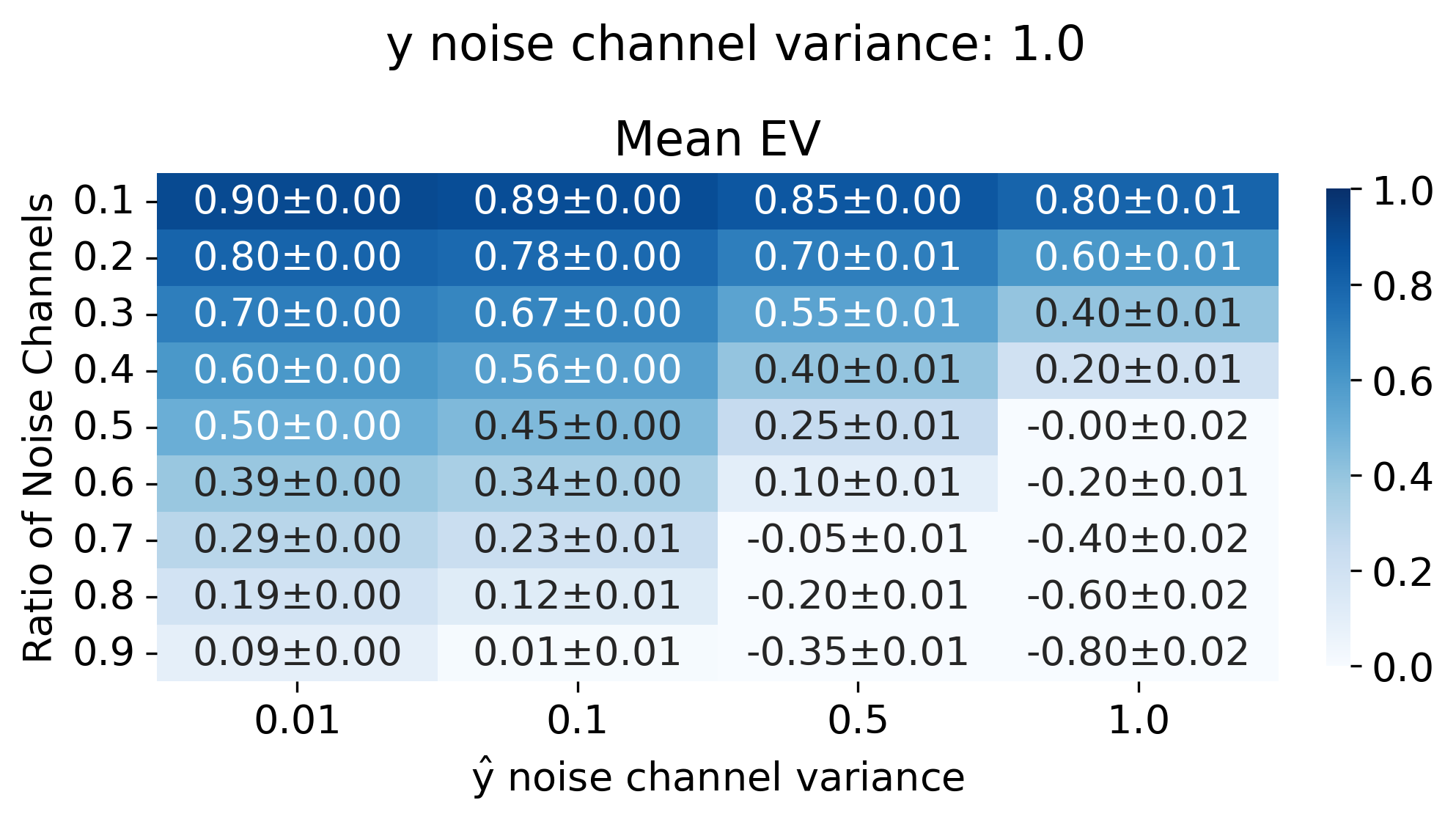}}
    \caption{Mean explained variance scores measured on simulated sinusoidal data (Fig. \ref{fig:r2-sample}) across hyperparameter sweeps. They exhibit the same noise-sensitivity issues as mean-R2, whereas Dim-R2 remains stable under these conditions (Fig. \ref{fig:noise_resilience_full}). Note the scores are identical to mean R2 in Fig. \ref{fig:noise_resilience_full} because $E(y)=E(\hat{y})$ for  $y,\hat{y}$ (Fig. \ref{fig:r2-sample}). Each entry shows the mean$\pm$standard deviation across 100 random repetitions.}
    \label{fig:resilience_EV}
\end{figure}

\begin{figure}[h]
    \centering
    \subfloat[]{\includegraphics[width=0.5\textwidth]{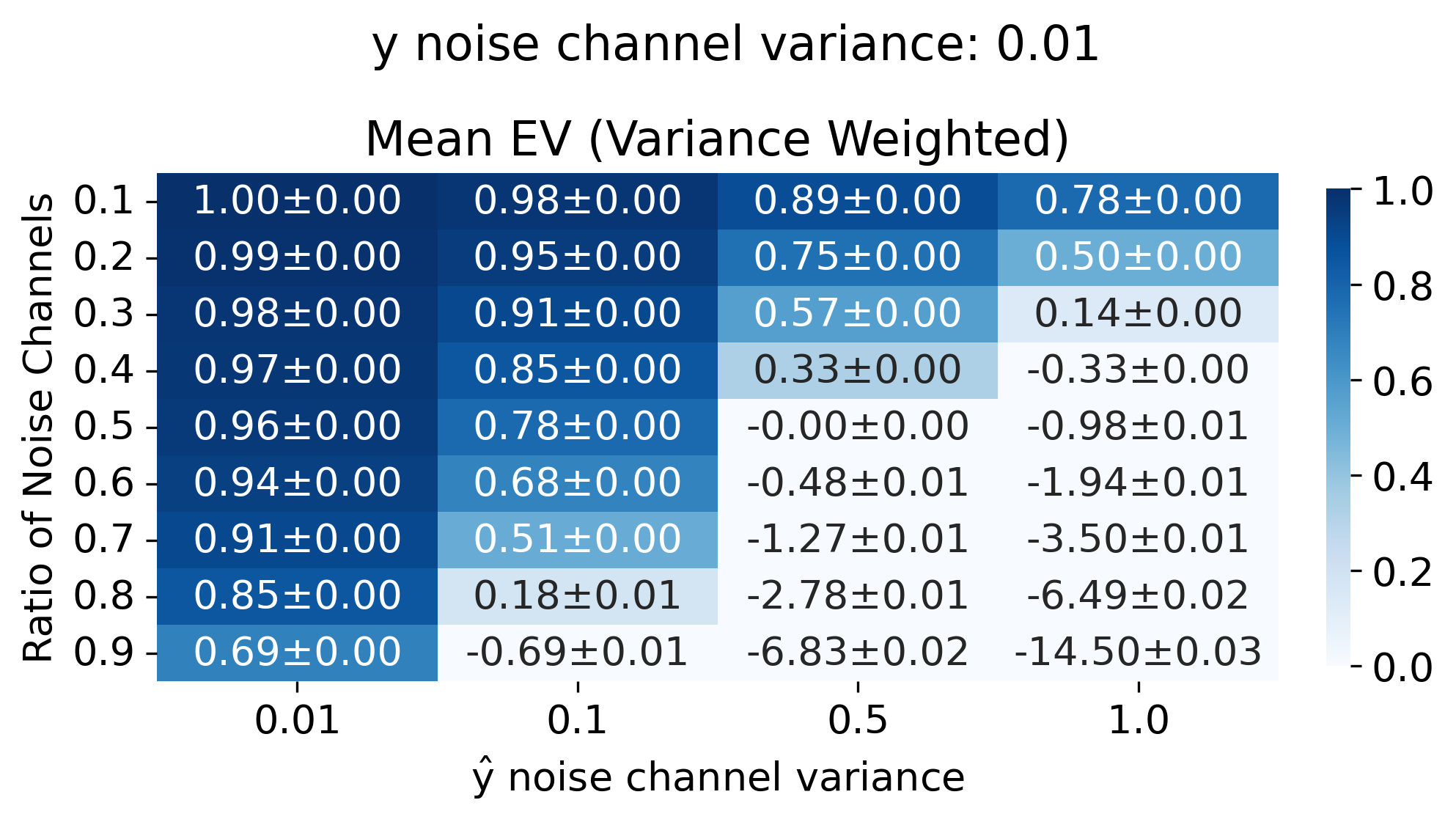}}
    \subfloat[]{\includegraphics[width=0.5\textwidth]{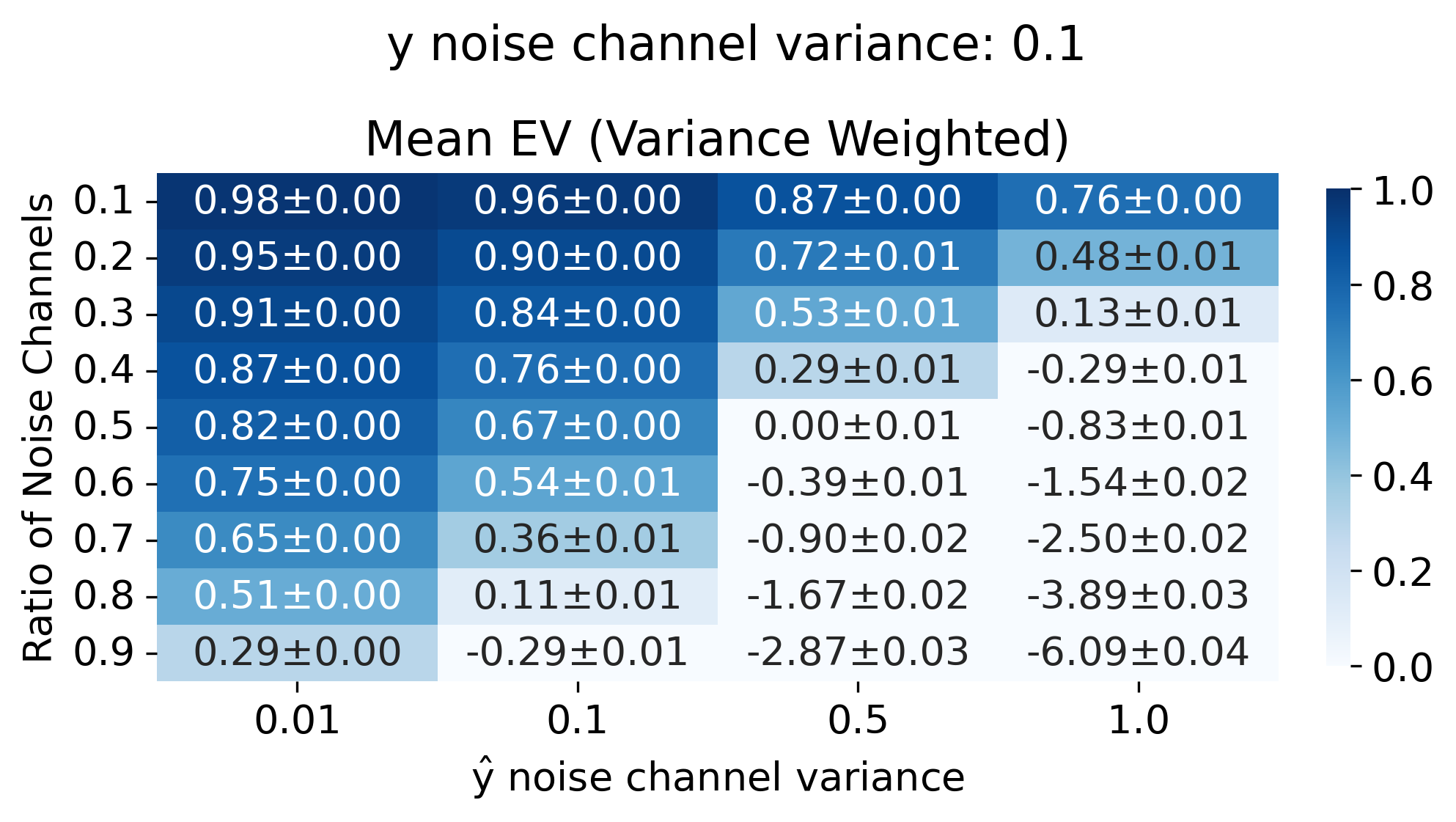}}
    \\
    \subfloat[]{\includegraphics[width=0.5\textwidth]{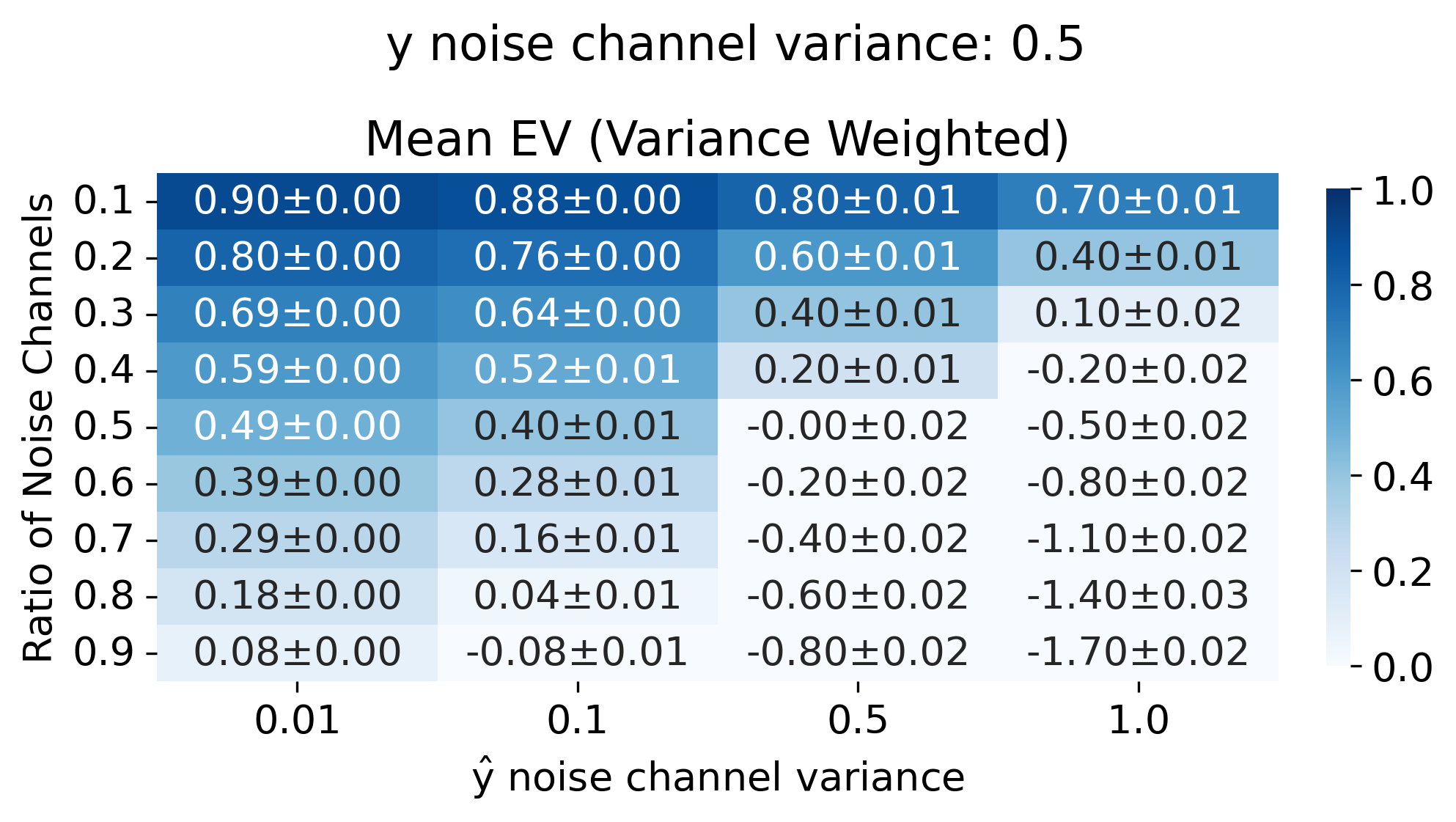}}
    \subfloat[]{\includegraphics[width=0.5\textwidth]{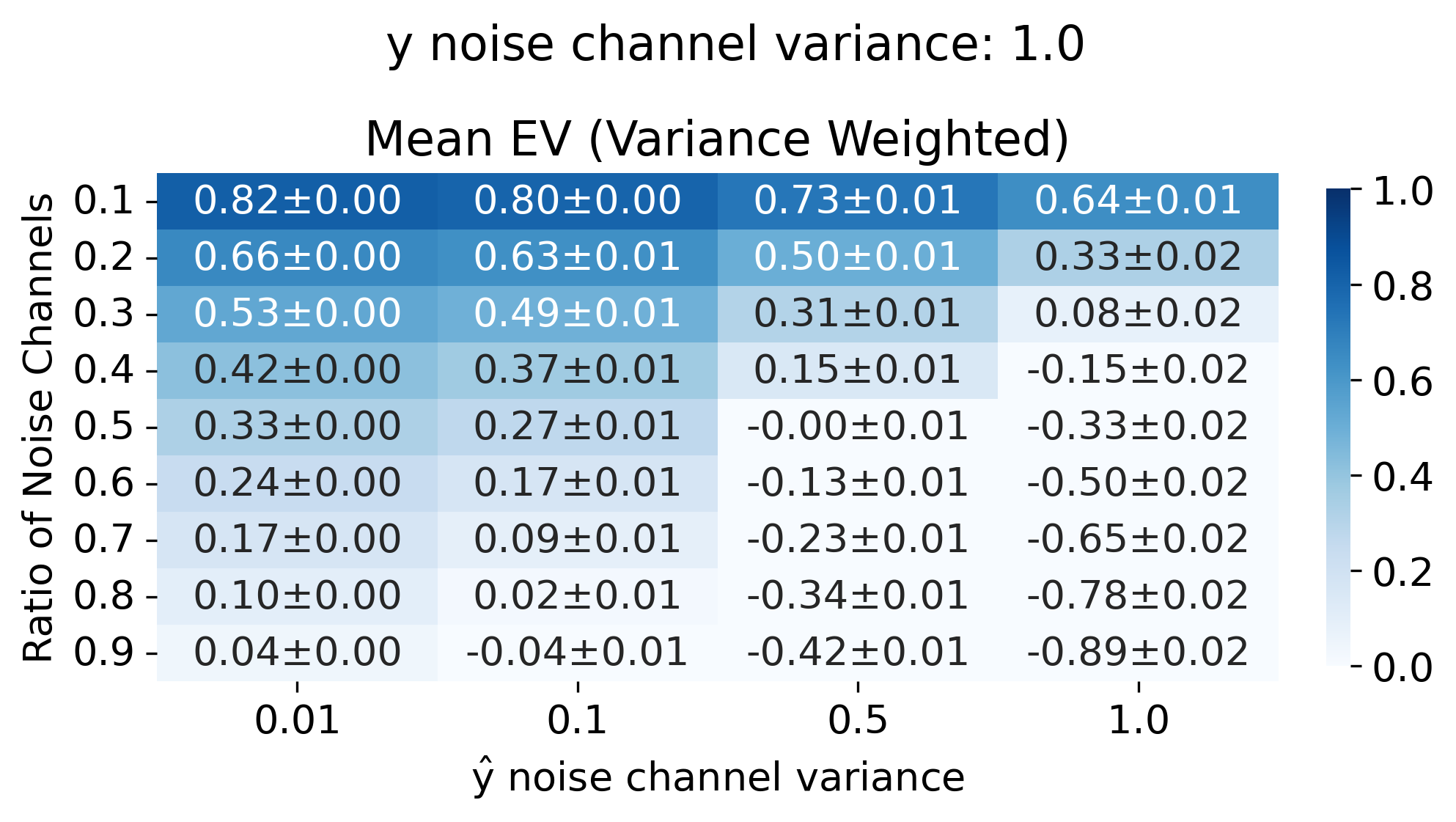}}
    \caption{Mean variance-weighted explained variance scores measured on simulated sinusoidal data (Fig. \ref{fig:r2-sample}) across hyperparameter sweeps. Note the scores are identical to Variance weighted mean R2 in Fig. \ref{fig:resilience_var_r2} because $E(y)=E(\hat{y})$ for  $y,\hat{y}$ (Fig. \ref{fig:r2-sample}). Each entry shows the mean$\pm$standard deviation across 100 random repetitions.}
    \label{fig:resilience_var_EV}
\end{figure}

\begin{figure}[h]
    \centering
    \subfloat[]{\includegraphics[width=0.5\textwidth]{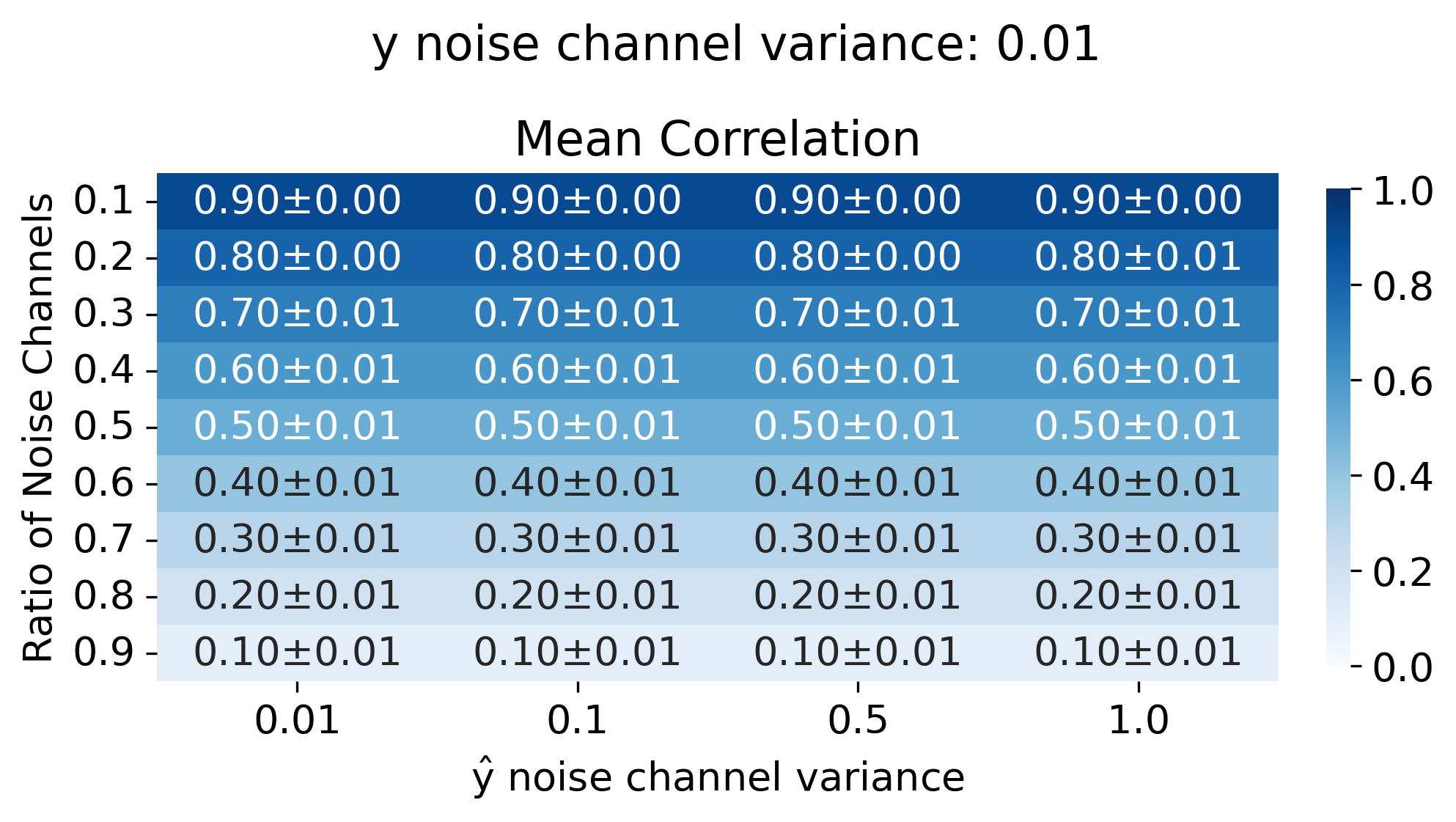}}
    \subfloat[]{\includegraphics[width=0.5\textwidth]{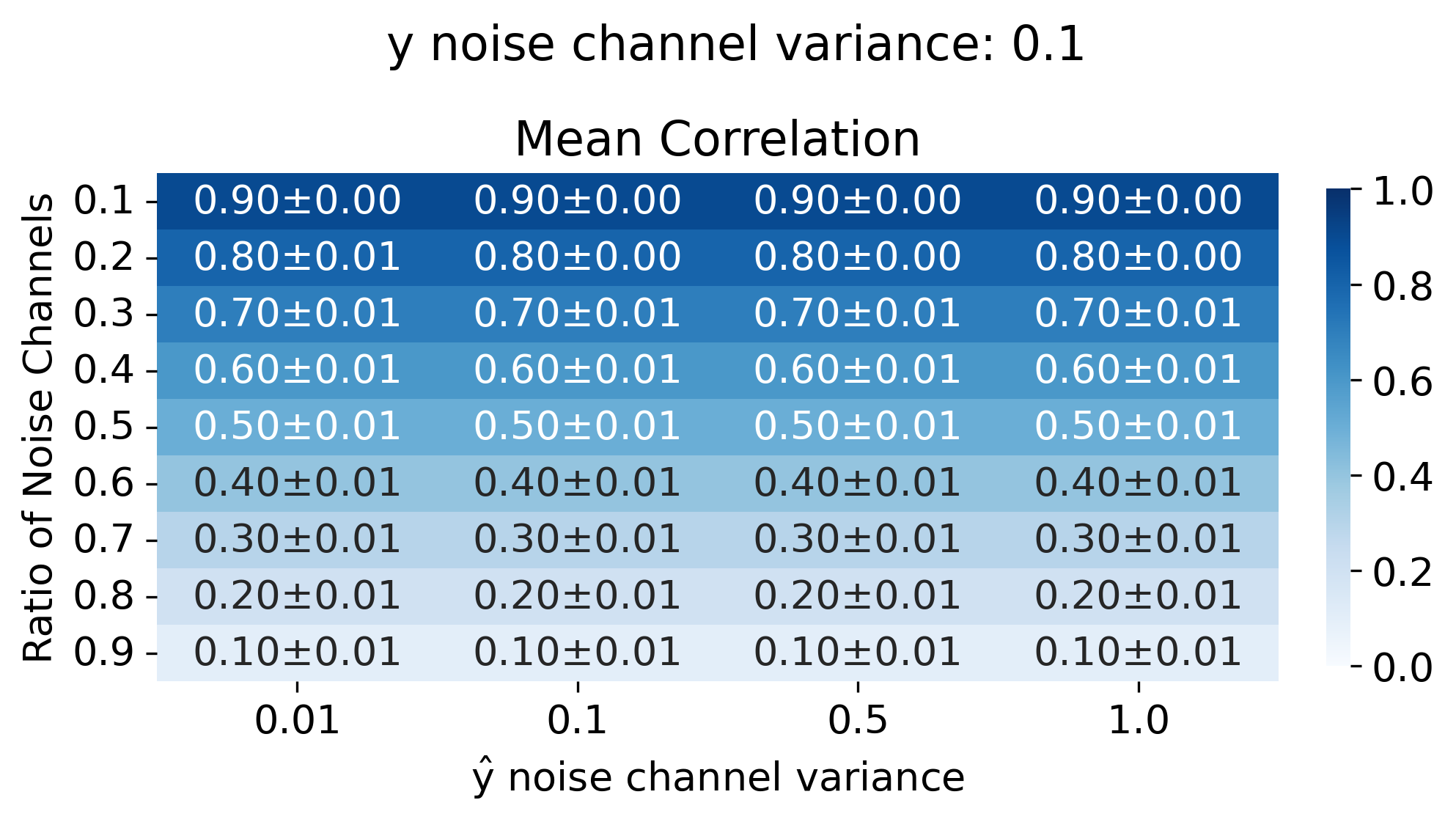}}
    \\
    \subfloat[]{\includegraphics[width=0.5\textwidth]{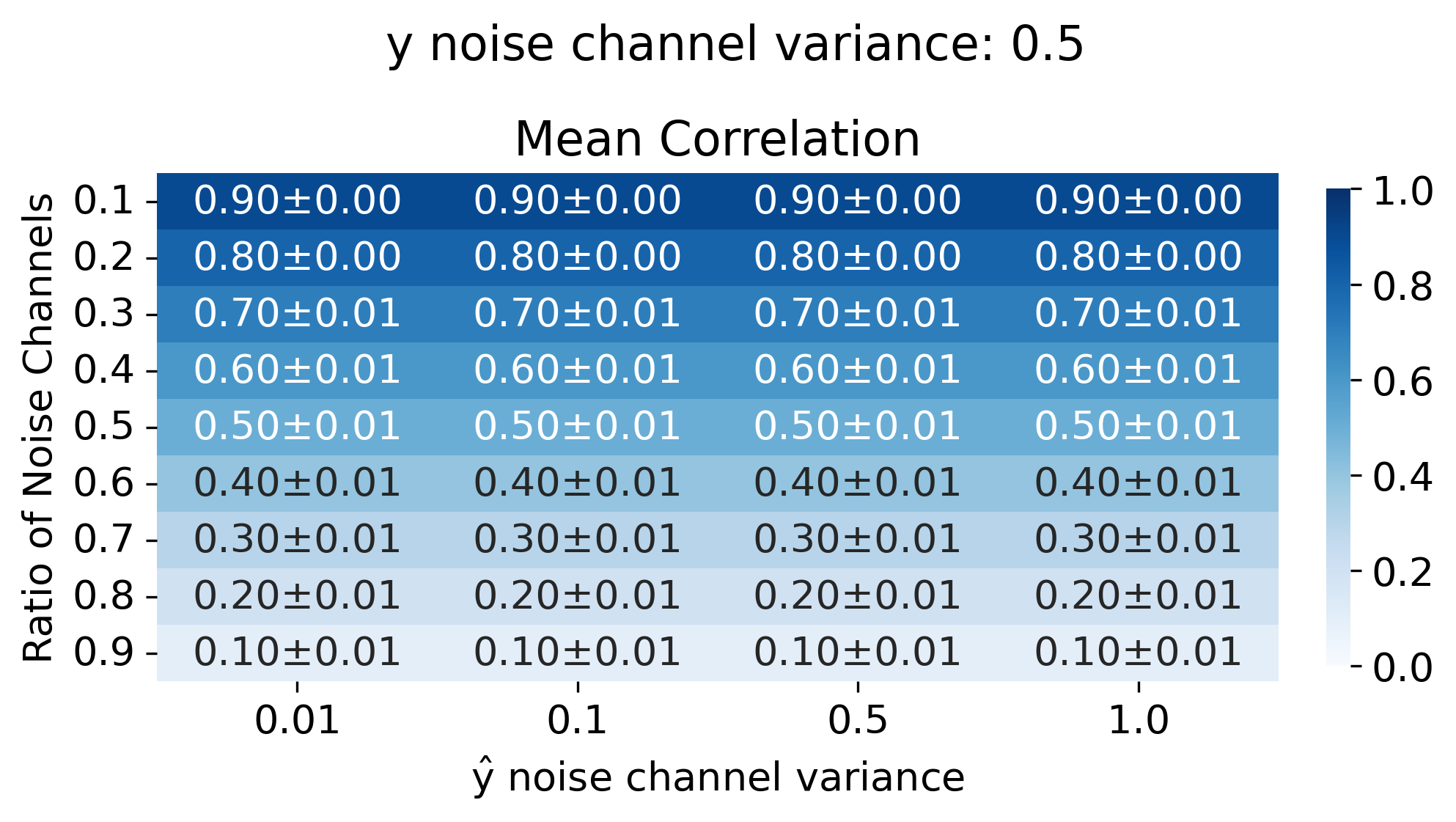}}
    \subfloat[]{\includegraphics[width=0.5\textwidth]{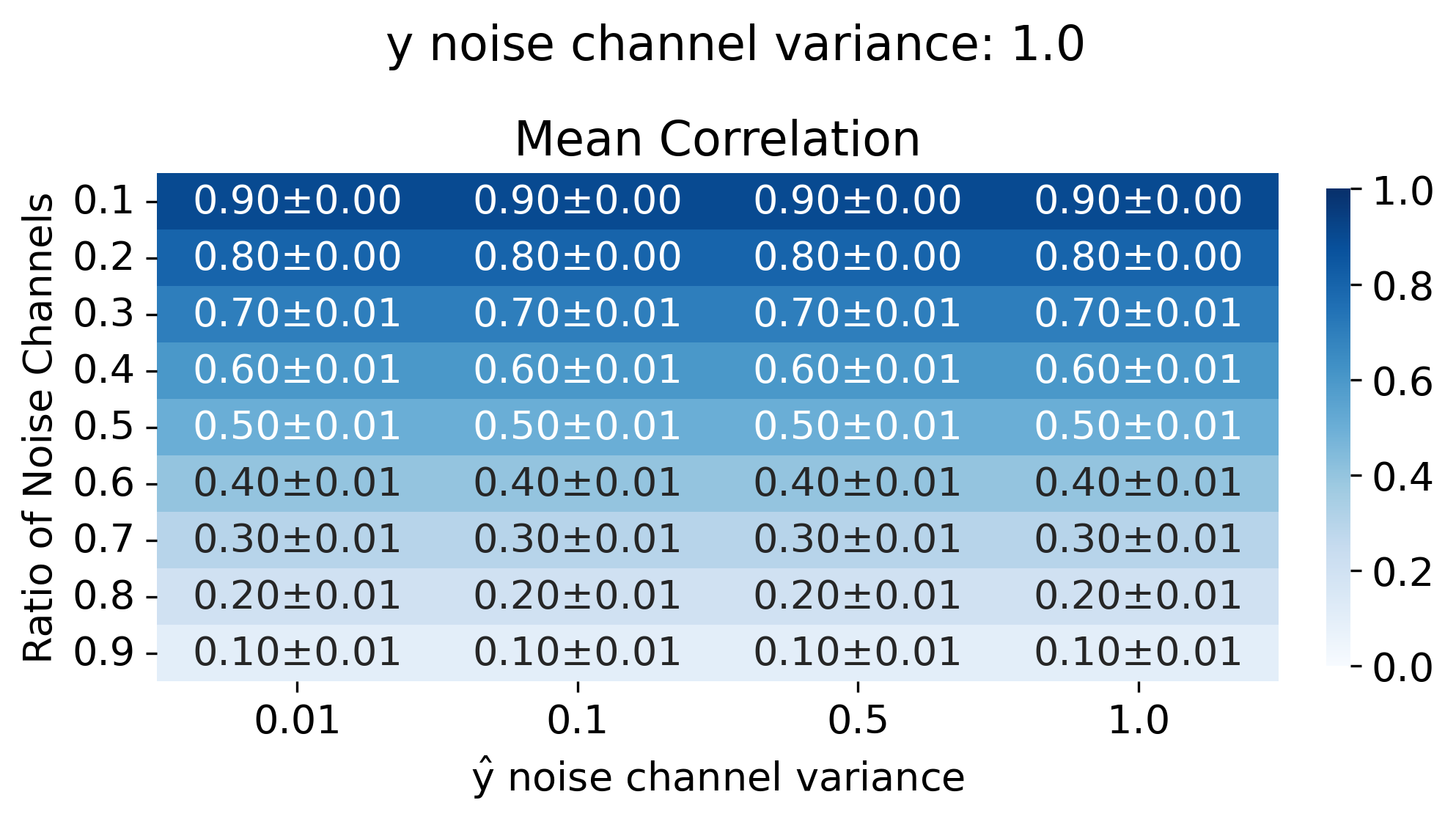}}
    \caption{Mean correlation scores measured on simulated sinusoidal data (Fig. \ref{fig:r2-sample}) across hyperparameter sweeps. Correlation does not account for varying levels of $y$ and $\hat{y}$ noise variance, and therefore is not a regression metric. Each entry shows the mean$\pm$standard deviation across 100 random repetitions.}
    \label{fig:resilience_corr}
\end{figure}

\FloatBarrier

\begin{table}[h]
    \centering
    \begin{tabular}{cccc}
    \toprule
         & Trial averaged (a) \& (d) & Single trial 1 (b) \& (e) & Single trial 2 (c) \& (f) \\
         \midrule
         Dim-R2 & 0.92 & 0.44 & 0.57 \\
         Mean R2& 0.57 & -0.28 & -0.41 \\
         \bottomrule
    \end{tabular}
    \caption{Dim-R2 and Mean R2 measured on $y$ and $\hat{y}$ of Fig. \ref{fig:sample-rnn_data}. Dim-R2 yields higher scores than mean R2 when the $y$ and $\hat{y}$ are heuristically similar, even in the presence of noise.}
    \label{tab:sample-resilience}
\end{table}

\begin{figure}[h]
    \centering
    \includegraphics[width=\textwidth]{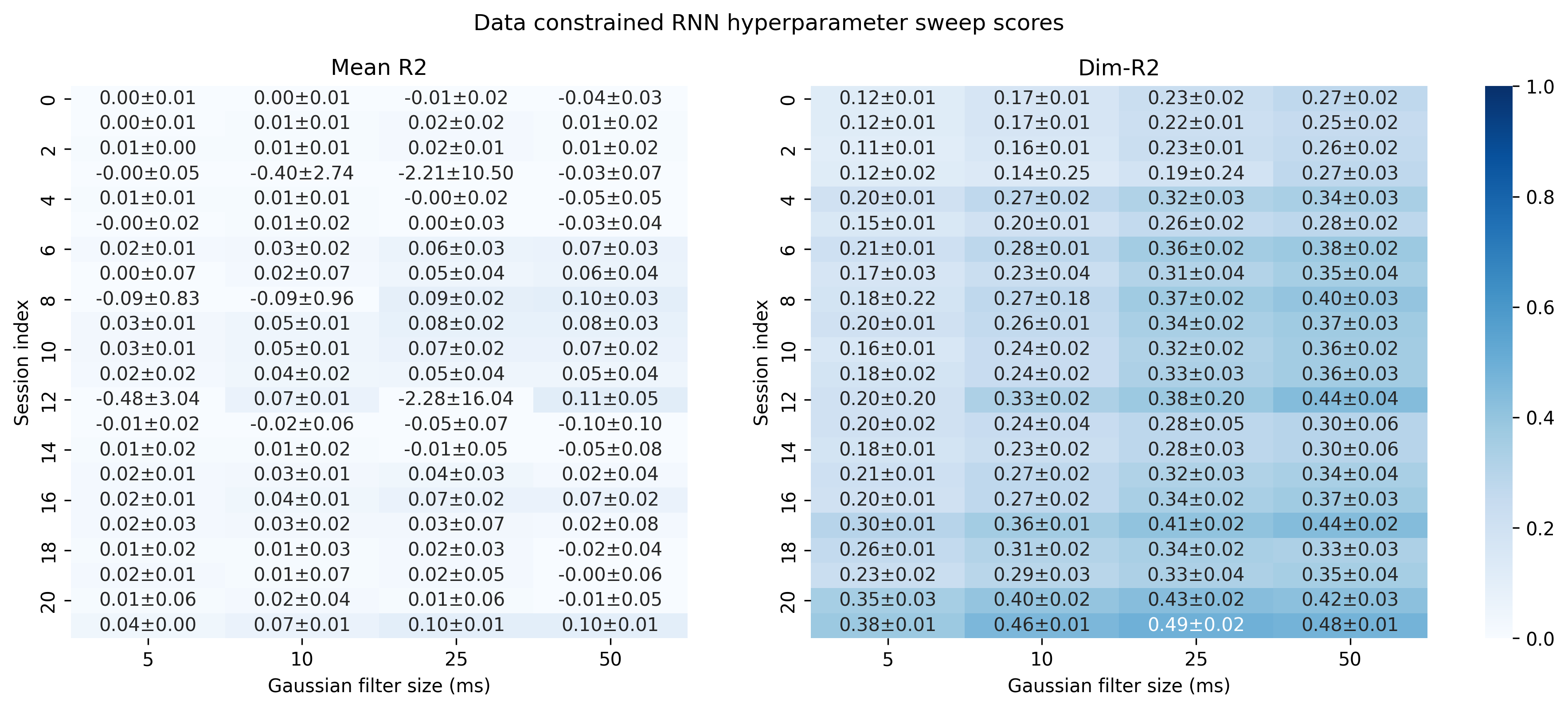}
    \caption{Dim-R2 highlights the presence of channels with high predictive accuracy in the presence of noisy channels. Scores were measured on data-constrained recurrent neural network predictions on neural spiketrains across hyperparameter sweeps. Each entry shows the mean$\pm$std across 3 random seeds and 15 cross-validation folds.}
    \label{fig:noise_resilience_dcrnn_heatmap}
\end{figure}

\FloatBarrier
\section{Supplemental Files}
\subsection{Dim-R2 measured across all training iterations and image classes for MNIST and CelebA}
\label{file:dim-R2-image-reconstruction}
\subsection{Python implementation of Dim-R2}
\label{file:dim-R2}

\FloatBarrier

\end{document}